\pdfoutput=1
\documentclass[12pt]{colt2026}

\usepackage[T1]{fontenc}
\usepackage{times}
\usepackage{booktabs}
\usepackage{multirow}
\usepackage{tikz}
\usetikzlibrary{arrows.meta}
\usepackage{refcount}%

\newcommand{\R}{\mathbb{R}}
\newcommand{\N}{\mathbb{N}}
\newcommand{\E}{\mathbb{E}}
\newcommand{\Hc}{\mathcal{H}}
\newcommand{\X}{\mathcal{X}}

\newcommand{\abst}{\bot}
\newcommand{\Risk}{\mathcal{R}}
\newcommand{\one}{\mathbf{1}}
\newcommand{\eps}{\varepsilon}
\DeclareMathOperator{\VC}{VC}
\DeclareMathOperator{\supp}{supp}

\newcommand{\Z}{\mathbb{Z}}
\newcommand{\bG}{\mathbf{G}}
\newcommand{\val}{\mathrm{val}}

\newcommand{\Q}{\mathbb{Q}}

\newcommand{\Ac}{\mathcal{A}}
\newcommand{\Vc}{\mathcal{V}}
\newcommand{\Wc}{\mathcal{W}}

\newcommand{\Dev}{\mathcal{D}}
\DeclareMathOperator{\Cl}{Cl}
\DeclareMathOperator{\sys}{sys}

\newcommand{\G}{\mathcal{G}}
\DeclareMathOperator{\DIS}{DIS}
\newcommand{\subappendix}{\par\setcounter{subsection}{0}\let\subappendixsaved\thesubsection\renewcommand{\thesubsection}{\thesection.\Alph{subsection}}\ifdefined\theHsubsection\let\subappendixsavedH\theHsubsection\renewcommand{\theHsubsection}{\theHsection.\Alph{subsection}}\fi}
\newcommand{\subappendixend}{\let\thesubsection\subappendixsaved\ifdefined\subappendixsavedH\let\theHsubsection\subappendixsavedH\fi}
\newenvironment{statement}[1]{\par\medskip\noindent\textbf{#1.}\ \itshape}{\normalfont\par\medskip}

\renewenvironment{proof}[1][\proofname]{\par\noindent{\bfseries\upshape #1.}\ }{\jmlrQED}
\providecommand{\qedhere}{}

\newcommand{\mm}{\mathrm{mm}}

\DeclareMathOperator{\conv}{conv}
\newcommand{\Gs}{\mathcal{G}}
\newcommand{\Sc}{\mathcal{S}}

\newcommand{\GF}{\mathrm{GF}(2)}

\usepackage{enumitem}
\setlist[enumerate]{topsep=2pt,itemsep=1pt,parsep=0pt,partopsep=0pt}
\setlist[itemize]{topsep=2pt,itemsep=1pt,parsep=0pt,partopsep=0pt}

\newcommand{\Score}{\mathrm{Sc}}
\newcommand{\Pc}{\mathcal{P}}

\newcommand{\Kc}{\mathcal{K}}

\newtheorem*{informal}{Main Theorem (informal)}
\newcommand{\proofin}[1]{{\unskip\nobreak\hfil\penalty50\hskip1em\null\nobreak\hfil\textup{(proof in #1)}\parfillskip=0pt\finalhyphendemerits=0\par}}
\NewDocumentEnvironment{mainresult}{m m m +b}{%
  \begin{#1}[#2]\label{#3}#4\end{#1}%
  \expandafter\gdef\csname mrkind@#3\endcsname{#1}%
  \expandafter\gdef\csname mrtitle@#3\endcsname{#2}%
  \expandafter\gdef\csname mrbody@#3\endcsname{#4}%
}{}
\newcommand{\mrname}[1]{\csname mrname@\csname mrkind@#1\endcsname\endcsname}
\expandafter\def\csname mrname@theorem\endcsname{Theorem}
\expandafter\def\csname mrname@corollary\endcsname{Corollary}
\expandafter\def\csname mrname@proposition\endcsname{Proposition}
\newcommand{\restatemain}[1]{\begingroup\renewcommand{\proofin}[1]{}%
  \begin{statement}{\mrname{#1}~\ref{#1} (\csname mrtitle@#1\endcsname)}\csname mrbody@#1\endcsname\end{statement}\endgroup}

\makeatletter
\def\ps@jmlrtps{%
  \let\@mkboth\@gobbletwo
  \def\@oddhead{}%
  \let\@evenhead\@oddhead
  \def\@oddfoot{\hfill\small\rmfamily\thepage\hfill}%
  \let\@evenfoot\@oddfoot
}
\makeatother

\hypersetup{pdftitle={Answer or Abstain: The Sample Cost of a Confidence Threshold},pdfauthor={Yilin Jiang}}

\title[Answer or Abstain]{Answer or Abstain: The Sample Cost of a Confidence Threshold}

\coltauthor{%
 \Name{Yilin Jiang} \Email{lien.jiang@ieee.org}\\
 \addr The Hong Kong University of Science and Technology (Guangzhou)%
}

\begin{document}

\maketitle

\begin{abstract}%
Evaluations of language models at a confidence threshold $t$ score a correct answer $1$, an abstention $0$ and a wrong answer $-t/(1-t)$. The shortfall from a perfect score is the risk of realizable learning with Chow's reject option, which interpolates between PAC learning at $t=0$ and reliable learning without errors at $t=1$. For nontrivial classes, the minimax rates at these two ends are set by the VC dimension and the star number, respectively. We determine how the rate depends on $1/(1-t)$ in between, for every binary class. Up to $t=1/2$ abstention gives no gain, and above $1/2$ the minimax risk of a nontrivial class after $n$ examples is $\pi/n$ up to absolute constants, where $\pi$ is the largest value of a panel system on at most $n+1$ points shattered by the class. A panel system consists of two adjacent layers of sets in which every lower set lies in many upper sets and every upper set contains many lower sets. The parameter $\pi$ lies between $\frac12\min\{d,n+1\}$ and $\min\{s,n+1\}$, where $d$ is the VC dimension and $s$ the star number, and its dependence on $n$ is necessary, since no rate $\min\{D,n+1\}/n$ with $D$ independent of $n$ holds for all classes with constants uniform in $t$. As $1/(1-t)$ grows, abstention lowers the risk by a factor of order $1/(1-t)$ for threshold functions and parities, of order $\log(1/(1-t))$ for dictators and monotone conjunctions on many features, and of constant order for intervals, boxes and affine halfspaces in dimension at least two, the last two regimes at large sample sizes. The shattered panel systems of a class, compared with its VC dimension, decide how much abstention removes from the cost $1/(1-t)$ of an error.
\end{abstract}


\section{Introduction}\label{sec:intro}

\citet{KNVZ26} propose evaluations of language models that state the penalty for a wrong answer, so that a model gains by abstaining when it is likely to be wrong. At a confidence threshold $t\in[0,1)$, a correct answer scores $1$, an abstention scores $0$ and a wrong answer scores $-t/(1-t)$, so answering $y$ is at least as good as abstaining exactly when $y$ is correct with probability at least $t$. A model evaluated in this way must learn from data both what to answer and where to abstain. The shortfall $1-\Score_t$ of its expected score from a perfect score is the risk of a loss that charges $1$ for an abstention and $K=1/(1-t)$ for an error, with penalty $\lambda=K-1$ (Section~\ref{sec:setting}).

A learner that never abstains has risk $K$ times its probability of error, so for a nontrivial class of VC dimension $d$ the best such learner has minimax risk of order $K\min\{d,n+1\}/(n+1)$ in the realizable model \citep{HLW94}. Abstention lowers the risk by at most a factor $\max\{1,K/2\}$ (Corollary~\ref{cor:gain}), so at a fixed $t$ the VC dimension gives the rate up to a factor of order $K$, and the part of this factor that a class recovers decides its sample cost at the thresholds $0.75$ and $0.9$ used in evaluations, where $K=4$ and $10$.

Dividing the loss by $K$ gives the loss of \citet{Cho70} with rejection cost $c=1-t$, and for two labels with a known posterior Chow's rule rejects only if $c<1/2$. As $t$ grows from $0$ to $1$, confidence-threshold scoring interpolates between realizable PAC learning, where the VC dimension sets the rate, and reliable and probably useful (RPU) learning \citep{RS88}, or perfect selective classification \citep{EW10}, whose abstention rate is bounded through compression set sizes \citep{WHE15} and the star number \citep{HY15,Han16}. For finite penalties prior results are upper bounds under label noise \citep{HW06,CDM16,BZ21} and results for online protocols \citep{SZB10,ZC16,NZ20}; none of them gives the dependence of the minimax rate on $1/(1-t)$ for binary classes at $1/2<t<1$.

\par\smallskip\noindent\hfil\fbox{\begin{minipage}{0.92\linewidth}
\textbf{Question 1.} How does the number of examples needed to bring the expected score at confidence threshold $t$ within $\varepsilon$ of a perfect score grow with $1/(1-t)$, from PAC learning at $t=0$ to RPU learning at $t=1$, and when does abstaining reduce it?
\end{minipage}}\hfil\par\smallskip

\begin{figure}[t]
\centering
\includegraphics{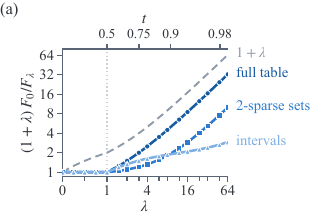}\hfill
\includegraphics{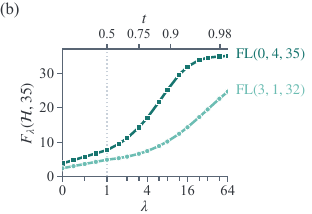}\hfill
\raisebox{2pt}{\begingroup
\input{figures/cube_style}
\begin{tikzpicture}[x=0.53cm,y=0.80cm,font=\sffamily\fontsize{7}{8}\selectfont]
\coordinate (e)    at (0,0);
\coordinate (s1)   at (-2.25,1);  \coordinate (s2) at (-0.75,1);
\coordinate (s3)   at (0.75,1);   \coordinate (s4) at (2.25,1);
\coordinate (s13)  at (-3.125,2); \coordinate (s12) at (-1.875,2); \coordinate (s23) at (-0.625,2);
\coordinate (s34)  at (0.625,2);  \coordinate (s14) at (1.875,2);  \coordinate (s24) at (3.125,2);
\coordinate (s123) at (-2.25,3);  \coordinate (s124) at (-0.75,3);
\coordinate (s134) at (0.75,3);   \coordinate (s234) at (2.25,3);
\coordinate (s1234) at (0,4);
\foreach \a/\b in {e/s1,e/s2,e/s3,e/s4,
  s1/s12,s1/s13,s1/s14,s2/s12,s2/s23,s2/s24,s3/s13,s3/s23,s3/s34,s4/s14,s4/s24,s4/s34,
  s12/s123,s12/s124,s13/s123,s13/s134,s14/s124,s14/s134,s23/s123,s23/s234,s24/s124,s24/s234,
  s34/s134,s34/s234,s123/s1234,s124/s1234,s134/s1234,s234/s1234}
  \draw[psgrid] (\a) -- (\b);
\foreach \v in {s13,s24,s123,s124,s134,s234,s1234} \node[psdot] at (\v) {};
\foreach \v/\t in {s1/1,s2/2,s3/3,s4/4} \node[psA,minimum size=10.8pt] (n\v) at (\v) {\t};
\foreach \v/\t in {s12/12,s14/14,s23/23,s34/34} \node[psW,minimum size=10.8pt] (n\v) at (\v) {\t};
\node[psC,minimum size=10.8pt] (ne) at (e) {$\varnothing$};
\foreach \a/\b in {s1/s12,s2/s12,s1/s14,s4/s14,s2/s23,s3/s23,s3/s34,s4/s34}
  \draw[psarrow] (n\a) -- (n\b);
\node[psInk,right=1.5pt] at (ne.east) {$c$};
\node[psnote,anchor=west] at (-4.6,0.05) {$a=b=2$};
\node[psLowerS,anchor=east] at (-2.75,1) {\normalsize$\mathcal A$};
\node[psUpperS,anchor=east] at (-3.45,2) {\normalsize$\mathcal W$};
\node[pstag] at (-4.9,4.35) {(c)};
\end{tikzpicture}%
\endgroup}
\caption{(a) $(1+\lambda)F_0/F_\lambda$ at $m=128$ for classes of VC dimension $2$, with $N=2$ keys for the full table and $N=12$ for the sparse sets. (b) $F_\lambda(\Hc,35)$ for two classes of VC dimension $4$ and star number $35$. (c) A panel system on four points, directed away from the center.}
\label{fig:hero}
\end{figure}

\noindent Equivalently, we seek the minimax risk $M_n(\Hc,\lambda)$ for every $\lambda\in[0,\infty]$ and compare it with $(1+\lambda)M_n(\Hc,0)$, the minimax risk of learners that never abstain. For nontrivial classes, the leave-one-out argument for one-inclusion graphs \citep{HLW94,ACSZ23,ADDST24} reduces $M_n(\Hc,\lambda)$, up to a constant factor, to $F_\lambda(\Hc,n+1)/(n+1)$, the value of a linear program on the projections of $\Hc$ on at most $n+1$ points (Section~\ref{sec:loss}). In the one-inclusion graphs of the projections, the VC dimension is the largest dimension of a subcube and the star number is the largest degree \citep{Han24}, and we show that above $t=1/2$ the value is determined by a shattered structure that has stars and the middle layers of subcubes as special cases. A panel system consists of a family of $(k-1)$-sets and a family of $k$-sets in which every lower set lies in at least $a$ upper sets and every upper set contains at least $b$ lower sets, and its value is $\min\{a,\lambda b\}$. A star with $s'$ leaves gives $a=s'$ and $b=1$, and for $t\ge1/2$ the two middle layers of a shattered set of size $d$ give a value of at least $d/2$. We write $\pi_\lambda(\Hc,m)$ for the largest value of a panel system on at most $m$ points that $\Hc$ shatters (Section~\ref{sec:panels}, Figure~\ref{fig:hero}(c)).

\begin{informal}
Let $\Hc$ be a nontrivial class with VC dimension $d$ and star number $s$, let $n\ge1$, $t\in[0,1)$ and $\lambda=t/(1-t)$. If $t\le1/2$, then $M_n(\Hc,\lambda)=(1+\lambda)\,M_n(\Hc,0)\asymp(1+\lambda)\min\{d,n+1\}/(n+1)$. If $t\ge1/2$, then
\[
\frac{\pi_\lambda(\Hc,n+1)}{4e(n+1)}\ \le\ M_n(\Hc,\lambda)\ \le\ \frac{4\,\pi_\lambda(\Hc,n+1)}{n+1},
\]
where $\frac12\min\{d,n+1\}\le\pi_\lambda(\Hc,n+1)\le\min\{s,n+1\}$.
\end{informal}

The theorem gives the dependence on $1/(1-t)$ through the abstention gain $B_n(\Hc,\lambda)=(1+\lambda)M_n(\Hc,0)/M_n(\Hc,\lambda)$. The gain equals $1$ up to $t=1/2$, and above $1/2$ it lies between $1$ and $K/2$ and is of order $K\min\{d,n+1\}/\pi_\lambda(\Hc,n+1)$, so abstention removes a factor of order $K=1/(1-t)$ from the risk exactly when the shattered panel systems stay at the scale of $\min\{d,n+1\}$. Abstention pays on a cube, where under an optimal prior the label of a hidden point is a fair coin, and lowers the leave-one-out value of a star with at least $\lambda$ leaves by less than a factor $2$ (Section~\ref{sec:starcube}). On natural classes the gain is of order $K$ for threshold functions, parities and full tables, of order $\log K$ for dictators and monotone conjunctions on many features, and of constant order for intervals, boxes and affine halfspaces in dimension at least two, the last two regimes once $n$ is large compared with $K$ times the dimension (Corollary~\ref{cor:gain}, Figure~\ref{fig:hero}(a)), and group-sparse supports attain every order between a constant and $K$ (Section~\ref{sec:ov-panels}).

Above $t=1/2$, the rate is not determined up to absolute constants by the VC dimension and the star number, nor by any parameter $D$ independent of $n$ in the form $\min\{D,n+1\}/(n+1)$. Two classes with the same VC dimension $d$ and the same star number can have rates that differ by a factor of order $\min\{K,d\}$ (Figure~\ref{fig:hero}(b)). Every such $D$ that gives the rates of all nontrivial classes in this form within constant factors $c\le C$ has $C/c\ge K/(2e)$, whereas the VC dimension at $t\le1/2$ and the star number at $t=1$ do so with $C/c\le4e$ and $C/c\le2e$ (Theorems~\ref{thm:binary} and~\ref{thm:noparam}).

First, we determine the minimax risk of every binary class at every confidence threshold up to absolute constant factors, and with it the dependence of the rate on $1/(1-t)$ between PAC and RPU learning. Above $t=1/2$ it is set for nontrivial classes by the largest value of a shattered panel system, which one bounds from below by exhibiting a single finite set system and from above by excluding large ones. Second, we show that, unlike at the two ends, no parameter independent of the sample size gives the rates of all classes as $\min\{D,n+1\}/(n+1)$ with constants uniform in $t$, through a lemma by which projections of a flag complex on fewer points than its systole are clique complexes of chordal graphs. Third, we derive the gain of natural classes from their panel systems. For dictators and monotone conjunctions the number of patterns limits the shattered panel systems, which gives a gain of order $\log K$ in the ranges of Corollary~\ref{cor:gain}, and for full tables the risk with abstention equals the worst expected missing mass at every $t\ge1/2$. Finally, we determine the minimax quantile at confidence $1-\delta$ up to absolute constants for classes with star number at least $2$ in the range of Proposition~\ref{prop:highprob}(b), and show that on some classes, when errors are expensive, no single learner attains it within a constant factor at all confidence levels.

\subsection{Related work}\label{sec:related}

\paragraph{Reject option.} The reject option goes back to \citet{Cho57,Cho70}, and $\ell_\lambda/K$ is the cost model of \citet[Section~9]{EW10} with rejection cost $1/K$, stated there without a rate. With noisy labels this loss has been studied for plug-in rules \citep{HW06}, convex surrogates \citep{BW08} and rejection functions \citep{CDM16}, with fast agnostic rates for $t>1/2$ \citep{BZ21}. In the realizable case and in our units, the bound of \citet[Theorem~2.1]{BZ21} is of order $K(d\log(n/d)+\log(1/\delta))/(n(t-1/2))$ with probability $1-\delta$ for $1/2<t<1$, where $\log z=\max\{1,\ln z\}$, and that of Proposition~\ref{prop:highprob}(a) is at most $(6.33K/n)(d+\ln(2/\delta))$ for $n\ge2$, without the factor $1/(t-1/2)$. In active learning, abstention gives exponential savings \citep{PZ21}, and the confidence-rated predictor of \citet[Algorithm~3]{ZC14} maximizes coverage by a linear program under bounds on its disagreement with each candidate hypothesis; our linear program prices an error at $K$ and an abstention at $1$ and determines the passive minimax risk up to absolute constants. Monotone conjunctions are not RPU learnable from polynomially many examples and rectangles in $\R^n$, $n\ge2$, not from finitely many \citep{Kiv95}, while at every fixed $t<1$ their rates are polynomial in the dimension (Corollary~\ref{cor:natural}). Abstention has also been studied in online learning \citep{LLW08,SZB10,ZC16,NZ20}, under adversarial injections \citep{GHMS23,EG26} or distribution shift \citep{GKKM20,KK21}, and with class-conditional constraints \citep{KDH26}.

\paragraph{One-inclusion graphs.} One-inclusion graphs \citep{HLW94,Hau95} give optimal multiclass learners \citep{DS14,BCDMY22}, optimal PAC bounds \citep{ACSZ23}, a dimension for realizable regression \citep{AHKKV23} and sample compression \citep{RBR09}. Closest to our approach, \citet{ADDST24} determine the transductive error of realizable multiclass learning under the $0$-$1$ loss exactly through the Hall complexity, a quantity of the one-inclusion graphs indexed by the sample size, which for binary classes equals the maximum subgraph density of \citet{HLW94} \citep[Remark~47]{ADDST24} and gives our $F_\lambda$ for $t\le1/2$ up to the factor $1+\lambda$. Our transfer~\eqref{eq:transfer} follows their comparison of transductive and expected sample complexities for losses with values in $[0,1]$ \citep[Proposition~32]{ADDST24}, with explicit constants for the reject-option loss. Their characterization applies Hall's theorem to orientations, in which a learner selects one endpoint of each edge; with a reject option the learner can also abstain on every edge, and for $t>1/2$ we determine the resulting value through shattered panel systems. \citet[Section~6]{Dug25} suggests Hall-type measures, a direction that Theorem~\ref{thm:binary} instantiates for this loss.

\section{Setting and panel systems}\label{sec:setting}

\subsection{Loss, minimax risk and the leave-one-out value}\label{sec:loss}

\paragraph{Loss and minimax risk.} Let $\X$ be a countable domain and $\Hc\subseteq\{0,1\}^{\X}$ a nonempty class. A rule $g$ assigns to each $x\in\X$ a distribution $g(x)\in\Pc(\{0,1,\abst\})$ over the two labels and the abstention $\abst$, where $\Pc(S)$ is the set of probability vectors on $S$. For a penalty $\lambda\in[0,\infty)$ and $K:=1+\lambda$, the loss of $\nu\in\Pc(\{0,1,\abst\})$ at label $y$ is $\ell_\lambda(\nu,y):=\nu(\abst)+K\,\nu(1-y)$, and the risk of $g$ under a distribution $P$ on $\X$ and a target $h\in\Hc$ is $R_\lambda(g;P,h):=\sum_{x\in\X}P(\{x\})\,\ell_\lambda(g(x),h(x))$. A learner maps each sample $S=((x_i,h(x_i)))_{i\le n}$, with $x_1,\dots,x_n$ drawn i.i.d.\ from $P$, to a rule $g_S$; learners may be randomized and improper. The minimax risk is $M_n(\Hc,\lambda):=\inf_g\sup_{P,\,h\in\Hc}\E_{x_{1:n}\sim P^n}R_\lambda(g_S;P,h)$.
At $\lambda=\infty$, which corresponds to the endpoint $t=1$, errors are forbidden, and $R_\infty(g;P,h)$ is the probability of abstention if the probability of error is $0$, and $+\infty$ otherwise.

\paragraph{Confidence-threshold scoring.} At threshold $t\in[0,1)$ the expected score of $g$ is $\Score_t(g;P,h)=\sum_xP(\{x\})\big[g(x)(h(x))-\frac{t}{1-t}\,g(x)(1-h(x))\big]$. In an evaluation, $x$ is a question, $h(x)$ is its correct answer and $\abst$ is the response that declines to answer. With $\lambda(t):=t/(1-t)$, so that $K=1/(1-t)$ and $t\ge1/2$ if and only if $\lambda\ge1$, we have $1-\Score_t(g;P,h)=R_{\lambda(t)}(g;P,h)$ for every $g$, $P$ and $h$. The class $\Hc$ is nontrivial if there are $h,h'\in\Hc$ and $x,x'\in\X$ with $h(x)=h'(x)$ and $h(x')\ne h'(x')$; a class that is not nontrivial has $M_n(\Hc,\lambda)=0$.

\paragraph{The leave-one-out value.} For a finite set $U$ and a nonempty $V\subseteq\{0,1\}^U$, the one-inclusion graph $G(V)$ has vertex set $V$ and an edge between two patterns that differ in exactly one coordinate. Let $f_K(a,b):=\min\{a+b,Ka,Kb\}$ and
\[
\tau_\lambda(V):=\max_{\mu\in\Pc(V)}\ \sum_{\{u,w\}\in E(V)}f_K(\mu_u,\mu_w),\qquad F_\lambda(\Hc,m):=\max_{U\subseteq\X,\ |U|\le m}\tau_\lambda(\Hc|_U),
\]
where $\Hc|_U=\{h|_U:h\in\Hc\}$, and $\tau_\infty$, $F_\infty$ are the limits as $\lambda\to\infty$. By linear programming duality $\tau_\lambda(V)$ is the value of a leave-one-out game on $G(V)$, and a known transfer argument (\citealp{HLW94}; \citealp[Remark~2.2]{ACSZ23}; \citealp[Proposition~32]{ADDST24}), carried out for this loss with explicit constants, gives for every $\lambda\in[0,\infty)$ and $n\ge1$
\begin{equation}\label{eq:transfer}
\frac{F_\lambda(\Hc,n+1)}{2e(n+1)}\ \le\ M_n(\Hc,\lambda)\ \le\ \frac{F_\lambda(\Hc,n+1)}{n+1},
\end{equation}
where the upper bound holds for every nonempty $\Hc$ and the lower bound for every nontrivial $\Hc$, also at $\lambda=\infty$.

\subsection{Panel systems}\label{sec:panels}

Let $d$ be the VC dimension of $\Hc$, $d_m:=\min\{d,m\}$, $\Delta(V)$ the largest degree in $G(V)$, and $s$ the star number of \citet{HY15}, the largest degree in the one-inclusion graphs of the projections of $\Hc$ \citep{Han24}.

\paragraph{Panel systems.} Let $U$ be a finite set and $k\ge1$. A panel system of rank $k$ on $U$ is a pair $(\Ac,\Wc)$ of nonempty families $\Ac$ of $(k-1)$-subsets and $\Wc$ of $k$-subsets of $U$ such that $a(\Ac,\Wc):=\min_{S\in\Ac}|\{T\in\Wc:S\subset T\}|\ge1$ and $b(\Ac,\Wc):=\min_{T\in\Wc}|\{S\in\Ac:S\subset T\}|\ge1$; its value is $\min\{a(\Ac,\Wc),\lambda\,b(\Ac,\Wc)\}$. The class $\Hc$ shatters $(\Ac,\Wc)$ with center $c\in\{0,1\}^U$ if $c\oplus\one_S\in\Hc|_U$ for every $S\in\Ac\cup\Wc$, where $\oplus$ is coordinatewise addition modulo $2$. Let $\pi_\lambda(\Hc,m)$ be the largest value of a panel system on a set $U\subseteq\X$ with $|U|\le m$ that $\Hc$ shatters with some center, and $\pi_\lambda(V)$ the same quantity for a set of patterns $V\subseteq\{0,1\}^U$, with value $0$ if there is none. A shattered panel system is a pair of adjacent layers of the one-inclusion graph around the center (Figure~\ref{fig:hero}(c)), and as for $d$ and $s$, a lower bound on $\pi_\lambda(\Hc,m)$ is witnessed by one finite projection.

\paragraph{Weighted cores and densities.} A two-sided core of $G(V)$ is a pair $(A,W)$ of disjoint nonempty subsets of $V$ such that $\delta_A$, the smallest number of neighbors in $W$ of a vertex of $A$, and $\delta_W$, the smallest number of neighbors in $A$ of a vertex of $W$, are positive. The $\lambda$-weighted core number $\kappa_\lambda(V)$, a weighted form of the $(\alpha,\beta)$-cores of bipartite graph mining \citep{ABFHMM07,LYLQZZ19}, is the largest value of $\min\{\delta_A,\lambda\delta_W\}$ over two-sided cores, or $0$ when $G(V)$ has no edge, and $\kappa_\lambda(\Hc,m):=\max_{U\subseteq\X,\,|U|\le m}\kappa_\lambda(\Hc|_U)$. For $\lambda>0$ and disjoint $A,W\subseteq V$, not both empty, let $E(A,W)$ be the set of edges between them; the bipartite density $\rho_\lambda(V)$ is the largest value of $|E(A,W)|/(|A|+|W|/\lambda)$, with weight $1$ on $A$ and $1/\lambda$ on $W$.

\paragraph{Example.} Let $N$ keys form $N/L$ groups of $L$ keys each, let $1\le k\le N/L$, and let $\Hc$ be the class of indicators of the sets of keys that meet at most $k$ groups, so that $d=kL$ and $s=N$; for $L=1$ these are the $k$-sparse sets. Every shattered panel system of this class has $a\le kL$ or $b\le k$, and Section~\ref{sec:ov-panels} turns this fact into the rate and the gain of the class.

\section{Main results}\label{sec:results}

\subsection{Binary classes at every threshold}\label{sec:binary}

\begin{mainresult}{theorem}{Binary classes at every threshold}{thm:binary}
Let $\Hc$ be nonempty with star number $s$ and let $n,m\ge1$.
\begin{enumerate}
\item[(a)] \emph{No gain up to $t=1/2$.} If $0\le\lambda\le1$, then $(1+\lambda)d_m/2\le F_\lambda(\Hc,m)\le(1+\lambda)d_m$ and $M_n(\Hc,\lambda)=(1+\lambda)\,M_n(\Hc,0)$.
\item[(b)] \emph{Panel systems.} If $1\le\lambda<\infty$, then $\pi_\lambda(V)\le\kappa_\lambda(V)\le4\pi_\lambda(V)$ for every finite nonempty $V\subseteq\{0,1\}^U$, $\frac K{2\lambda}\pi_\lambda(\Hc,m)\le F_\lambda(\Hc,m)\le4\pi_\lambda(\Hc,m)$ and $\frac K{2\lambda}\kappa_\lambda(\Hc,m)\le F_\lambda(\Hc,m)\le2\kappa_\lambda(\Hc,m)$. If moreover $\Hc$ is nontrivial, then, with $\pi=\pi_\lambda(\Hc,n+1)$ and $\kappa=\kappa_\lambda(\Hc,n+1)$,
\[
\frac{\pi}{4e(n+1)}\le M_n(\Hc,\lambda)\le\frac{4\pi}{n+1}\qquad\text{and}\qquad\frac{\kappa}{4e(n+1)}\le M_n(\Hc,\lambda)\le\frac{2\kappa}{n+1}.
\]
\item[(c)] \emph{Errors forbidden.} $M_n(\Hc,\infty)\le\min\{s,n+1\}/(n+1)$, and if $\Hc$ is nontrivial, $M_n(\Hc,\infty)\ge\min\{s,n+1\}/(2e(n+1))$.
\proofin{Section~\ref{sec:proof} and Appendices~\ref{app:minimax} and~\ref{app:criterion}}
\end{enumerate}
\end{mainresult}

\noindent Part~(a) is the exact minimax counterpart of Chow's rule, which never rejects when the rejection cost $1-t$ is at least $1/2$, and part~(c) is the star-number rate of RPU learning. For integer $\lambda$ the factors $K/(2\lambda)$ and $2$ around $\kappa_\lambda$ are attained or approached, and those of~\eqref{eq:transfer} are asymptotically optimal.

\subsection{Simpler parameters do not determine the rate}\label{sec:beyond}

\paragraph{Equal VC dimension and star number.} For integers $c\ge0$ and $k,N\ge1$, the class $\mathrm{FL}(c,k,N)$ on $c+N$ points consists of all patterns on the first $c$ points combined with at most $k$ ones on the other $N$. The classes $\mathrm{FL}(d-1,1,N)$ and $\mathrm{FL}(0,d,N+d-1)$ both have VC dimension $d$ and star number $N+d-1$, and for $\lambda\ge1$, $K\le N/d$ and $m\ge N+d-1$ the value $F_\lambda$ of the second is at least $dK/(2(d-1+K))\ge\frac14\min\{d,K\}$ times that of the first (Figure~\ref{fig:hero}(b)), since the value of the first is at most $d-1+\min\{N,K\}$ while the $(d-1)$-sets and the $d$-sets of the $N+d-1$ points form a panel system with $a=N$ and $b=d$ that the second class shatters.

\begin{mainresult}{theorem}{No parameter independent of the sample size}{thm:noparam}
Let $1\le\lambda<\infty$ and $K=1+\lambda$. For a family $\mathcal K$ of nontrivial classes and a map $D:\mathcal K\to[0,\infty]$, which may depend on $\lambda$ but not on $n$, call constants $0<c\le C<\infty$ admissible if
\[
c\,\min\{D(\Hc),n+1\}\ \le\ (n+1)\,M_n(\Hc,\lambda)\ \le\ C\,\min\{D(\Hc),n+1\}\qquad\text{for all }\Hc\in\mathcal K\text{ and }n\ge1.
\]
\begin{enumerate}
\item[(a)] For every integer $d\ge2$ and every map $D$ on the nontrivial classes of VC dimension $d$, all admissible constants satisfy $C/c\ge dK/(2e\max\{d,\lambda\})\ge\min\{K,d\}/(2e)$.
\item[(b)] For every map $D$ on the nontrivial classes, all admissible constants satisfy $C/c\ge K/(2e)$.
\item[(c)] For every integer $d\ge1$, the constants $c=1/(8e\min\{\lambda,d\})$ and $C=2$ are admissible for $D=\bar\kappa_\lambda$ on the nontrivial classes of VC dimension $d$, where $\bar\kappa_\lambda(\Hc)=\sup_m\kappa_\lambda(\Hc,m)$, and the constants $c=1/(4e)$ and $C=K$ are admissible for the VC dimension on the nontrivial classes.
\proofin{Appendix~\ref{app:noparameter}}
\end{enumerate}
\end{mainresult}

\noindent By~(c), the VC dimension attains the bound of~(b) up to the factor $8e^2$, while at $t\le1/2$ and at $t=1$ the VC dimension and the star number give all rates in this form with $C/c\le4e$ and $C/c\le2e$ (Theorem~\ref{thm:binary}). For a fixed nontrivial class $\Hc$ of finite VC dimension $d$ and a fixed $\lambda\ge1$, $\kappa_\lambda(\Hc,m)$ is nondecreasing in $m$, at most $2Kd$ and of the form $\min\{a,\lambda b\}$ for integers $a,b$ (Lemma~\ref{lem:kappaeventual}), so it is constant for all $m$ at least some $m_0$. With $\bar\pi_\lambda(\Hc)=\sup_m\pi_\lambda(\Hc,m)$, the minimax risk is then of order $\bar\pi_\lambda(\Hc)/n$ for $n+1\ge m_0$, and the smallest such $m_0$ is not bounded by any function of $d$, $\lambda$ and $\bar\kappa_\lambda(\Hc)$ (Corollary~\ref{cor:C-cutoff}).

\subsection{When abstention pays}\label{sec:gain}

In the natural classes below, boxes include the empty box, dictators $x\mapsto x_i$ the zero function, and $u\asymp v$ means $cv\le u\le Cv$ for absolute constants $c,C>0$. For the rows with $\Psi_\lambda(m)=\min\{m,\Psi\}$, risk $\varepsilon$ needs of order $\Psi/\varepsilon$ examples for every $\varepsilon$ below an absolute constant.

\begin{mainresult}{corollary}{Natural classes}{cor:natural}
Let $\lambda\ge1$, $K=1+\lambda$, $n\ge1$ and $k\ge1$. Each class below is nontrivial, $F_\lambda(\Hc,m)\asymp\Psi_\lambda(m)$ for all $m\ge1$, and $M_n(\Hc,\lambda)\asymp\Psi_\lambda(n+1)/(n+1)$.\proofin{Appendix~\ref{app:applications}}
\begin{center}\footnotesize
\renewcommand{\arraystretch}{0.92}\setlength{\tabcolsep}{3pt}
\begin{tabular}{@{}lc@{\hspace{0.7em}}lc@{}}
\toprule
class & $\Psi_\lambda(m)$ & class & $\Psi_\lambda(m)$\\
\midrule
threshold functions on $\Q$ & $1$ & affine halfspaces in $\Q^d$, $d\ge2$ & $\min\{m,(d+1)K\}$\\
parities on $\{0,1\}^D$, $D\ge1$ & $\min\{m,D\}$ & homogeneous halfspaces in $\Q^d$, $d\ge3$ & $\min\{m,dK\}$\\
full table on $N\ge2$ keys & $\min\{m,N\}$ & boxes in $\Q^d$, $d\ge1$ & $\min\{m,2dK\}$\\
$k$-sparse sets on $N\ge2$ keys & $\min\{m,N,kK\}$ & dictators on $p\ge2$ features & $\min\{m,p,K\log_2p/\log_2K\}$\\
unions of at most $k$ intervals & $\min\{m,kK\}$ & monotone conjunctions, disjunctions & $\min\{m,KD/\log_2K\}$\\
& & \quad on $D\ge2$ attributes, $K\le2^{D/2}$ & \\
\bottomrule
\end{tabular}
\end{center}
\end{mainresult}

\paragraph{Stars and cubes.}\label{sec:starcube} Let $t\ge1/2$. Without abstention the leave-one-out value of a pattern set $V$ is $K\tau_0(V)$. On the cube $Q_d=\{0,1\}^d$, $\tau_0=d/2$ and $\tau_\lambda=d$, a factor $K/2$, since under an optimal prior the label of a hidden point is a fair coin, so abstaining costs $1$ and guessing costs $K/2$ in expectation. On the star $S_m$, a default pattern with $m$ possible single-point exceptions, $\tau_0=m/(m+1)$ and $\tau_\lambda=Km/(m+\lambda)$, a factor $(m+\lambda)/(m+1)<2$ for $m\ge\lambda$, since an optimal prior gives each exception posterior probability $1-t$ and answering the default costs as much as abstaining. A single cube or star does not decide the gain of a class: full tables contain large stars yet gain $K/2$, and $k$-sparse sets contain cubes yet gain less than $2$ when $kK\le\min\{N-1,n+1\}$ (Corollary~\ref{cor:gain}(iii)); the panel systems of the projections decide it.

\begin{mainresult}{corollary}{Abstention gain}{cor:gain}
Let $\Hc$ be nontrivial, $n\ge1$, $K=1+\lambda$ and $B_n(\Hc,\lambda)=(1+\lambda)M_n(\Hc,0)/M_n(\Hc,\lambda)$. Then $1\le B_n(\Hc,\lambda)\le\max\{1,K/2\}$, with $B_n(\Hc,\lambda)=1$ for $\lambda\le1$, and $B_n(\Hc,\lambda)=K/2$ for every $\lambda\ge1$ if $\Hc$ has infinite VC dimension, is a full table on at least two points or is a class of parities. For $\lambda\ge1$, $B_n(\Hc,\lambda)\asymp Kd_{n+1}/\kappa_\lambda(\Hc,n+1)$, and for the classes of Corollary~\ref{cor:natural} this gives three regimes as $K$ grows.
\begin{enumerate}
\item[(i)] \emph{Order $K$.} $B_n\asymp K$ for threshold functions, parities and full tables, for every $n$, and for every class when $n+1\le d$.
\item[(ii)] \emph{Order $\log K$.} $B_n\asymp\log K$ for dictators on $p$ features when $K\log_2p/\log_2K\le\min\{p,n+1\}$, and for monotone conjunctions and disjunctions on $D$ attributes when $K\le2^{D/2}$ and $n+1\ge KD/\log_2K$.
\item[(iii)] \emph{Constant order.} $B_n<2$ for $k$-sparse sets on $N$ keys when $kK\le\min\{N-1,n+1\}$, and $B_n$ is bounded by an absolute constant for unions of $k$ intervals when $n+1\ge kK$ and for affine halfspaces, homogeneous halfspaces and boxes in $\Q^d$ when $n+1\ge(d+1)K$, $dK$, $2dK$, respectively.
\proofin{Appendices~\ref{app:constants} and~\ref{app:gain}}
\end{enumerate}
\end{mainresult}

\noindent Figure~\ref{fig:hero}(a) plots $(1+\lambda)F_0/F_\lambda$ at $m=128$, which by~\eqref{eq:transfer} determines $B_{127}$ within the factor $2e$ and is nondecreasing in $\lambda$. It equals $\max\{1,K/2\}$ for the full table on two points and is at most $3$ for intervals on the plotted range.

\begin{mainresult}{proposition}{Full tables}{prop:exact}
Let $n\ge1$, $\lambda\in[0,\infty)$, $K=1+\lambda$, $|\X|=N\in\{2,3,\dots\}\cup\{\infty\}$ and $\Hc=\{0,1\}^{\X}$. For a distribution $P$ on $\X$ let $\mm_n(P)=\sum_xP(\{x\})(1-P(\{x\}))^n$ be the expected missing mass of $n$ samples, and let $\mm^*_n(N)=\sup_P\mm_n(P)$. Then $M_n(\Hc,\lambda)=\min\{1,K/2\}\,\mm^*_n(N)$ and $(1+\lambda)M_n(\Hc,0)=\frac K2\mm^*_n(N)$, so $B_n(\Hc,\lambda)=\max\{1,K/2\}$. For each fixed $P$, the smallest worst-case risks over the targets $h\in\Hc$ alone are $\min\{1,K/2\}\,\mm_n(P)$ for learners that may abstain and $\frac K2\,\mm_n(P)$ for learners that never abstain.\proofin{Appendix~\ref{app:exactcosts}}
\end{mainresult}

\noindent For every $t\ge1/2$, Proposition~\ref{prop:exact} makes the minimax risk of a full table with abstention equal to the worst expected missing mass $\mm^*_n(N)$, and $\mm_n(P)$ is the expected fraction of $n+1$ samples from $P$ that are points seen exactly once, the quantity estimated by the Good--Turing estimator \citep{Goo53}. \citet{KV24} relate the hallucination rate of calibrated language models on arbitrary facts to the fraction of facts seen once in training. With the keys of the full table as such facts, Proposition~\ref{prop:exact} is the corresponding exact minimax identity for binary questions with a reject option (Remark~\ref{rem:F-KV}). At the thresholds $0.5$, $0.75$ and $0.9$ of \citet{KNVZ26}, the gain of a full table is exactly $K/2=1$, $2$ and $5$ for every $n$.

\subsection{High probability}\label{sec:remarks}

With probability $1-\delta$ the rate acquires a confidence term that depends on the cost of an error. For a learner $g$ let $Q_\delta(g;P,h)$ be the $(1-\delta)$-quantile of its risk over the sample, let $\eps_n(\Hc,\lambda,\delta)=\inf_g\sup_{P,h\in\Hc}Q_\delta(g;P,h)$, and let $s_n=\min\{s,n\}$.

\begin{mainresult}{proposition}{High probability}{prop:highprob}
Let $\Hc$ be nontrivial with star number $s$, $\lambda\in[0,\infty)$, $K=1+\lambda$ and $n\ge2$.
\begin{enumerate}
\item[(a)] Let $\delta\in(0,1)$. With probability at least $1-\delta$, the average of the learner of the upper bound in~\eqref{eq:transfer} over the prefixes of the sample of lengths $\lfloor n/2\rfloor$ to $n-1$ \citep{ACSZ23}, which does not depend on $\delta$, has risk at most $(6.33/n)\big(F_\lambda(\Hc,n)+K\ln(2/\delta)\big)$. With probability at least $1-\delta$, the learner that abstains exactly on the disagreement region of the version space has risk at most $(6.33/n)\big(s_n+\ln(2/\delta)\big)$.
\item[(b)] If $s\ge2$, $\delta\le1/64$ and $\ln(1/\delta)\le n$, then $\eps_n(\Hc,\lambda,\delta)$ lies between $\frac1{108}$ and $\frac{8e}{e-1}$ times $\big(F_\lambda(\Hc,n)+\min\{s_n,K\ln(1/\delta)\}+\ln(1/\delta)\big)/n$.
\item[(c)] For every $C_0\ge1$ there are a class of finite star number, $\lambda\ge1$, $\delta\le1/64$ and $n_0$ such that for every $n\ge n_0$ both $\eps_n(\Hc,\lambda,1/64)$ and $\eps_n(\Hc,\lambda,\delta)$ are of order $1/n$, $\ln(1/\delta)\le n$, and every learner $g$ has $\sup_{P,h}Q_{\delta'}(g;P,h)>C_0\,\eps_n(\Hc,\lambda,\delta')$ for $\delta'=1/64$ or for $\delta'=\delta$.
\proofin{Appendix~\ref{app:highprob}}
\end{enumerate}
\end{mainresult}

\noindent The term $\min\{s_n,K\ln(1/\delta)\}$ comes from unseen points of small total mass whose labels the sample leaves open. Answering at such a point costs $K$ times its mass when the answer is wrong, and abstaining costs its mass, so a learner pays of order $K\ln(1/\delta)/n$ by answering there or a star-number term of order $s_n/n$ by abstaining on all of them, as the two learners of~(a) do; the second bound sharpens $(21s+16\ln(3/\delta))/n$ of \citet[Theorem~10]{Han16}. For $K\le2$ the prefix-averaged learner, which does not depend on $\delta$, attains $\eps_n$ within an absolute constant at every $\delta$ of~(b), as in PAC learning \citep{ACSZ23}. For large $K$, part~(c) gives classes on which no learner does. A learner near-optimal at constant confidence answers on most of a large star set, and at small $\delta$ a target that differs from these answers costs it order $K\ln(1/\delta)/n$.

The lower bound of~(b) on $F_\lambda(\Hc,n)/n$ uses the two targets at the ends of an edge whose direction is not sampled. They receive the same sample, so the action there costs one of them at least $1/2$ or $K/2$, and charging the edge to that target bounds the quantile at constant confidence, without the factor $K$ lost in converting an expected-risk bound.

\section{Proof of Theorem~\ref{thm:binary} and technical overview}\label{sec:overview}

\subsection{Proof of Theorem~\ref{thm:binary}}\label{sec:proof}

\paragraph{The leave-one-out game and part~(a).} On an edge of $G(V)$ with prior weights $a$ and $b$ at its endpoints, abstaining costs $a+b$ and predicting either label costs $K$ times the weight of the other endpoint, which gives $f_K(a,b)$; by linear programming duality, $\tau_\lambda(V)$ is both the smallest worst-case total leave-one-out cost of a rule (summed over the held-out coordinates) and the largest Bayes cost of a prior. For $\lambda\le1$, $f_K(a,b)=K\min\{a,b\}$, so $\tau_\lambda=K\tau_0$, the largest edge density of a subgraph times $K$, which is at most $Kd_V$, with $d_V$ the VC dimension of $V$ \citep{HLW94}, and equals $K|U|/2$ for $V=\{0,1\}^U$. Moving the abstention mass of a rule to the two labels in equal parts changes its loss by $\nu(\abst)(K/2-1)\le0$, so forbidding abstention does not change the minimax risk, and part~(a) follows. Part~(c) follows from the transfer at $\lambda=\infty$, $\tau_\infty=\Delta$ and $\max_{|U|\le m}\Delta(\Hc|_U)=\min\{s,m\}$.

For part~(b) fix $1\le\lambda<\infty$ and a finite nonempty $V\subseteq\{0,1\}^U$, with the bipartite density $\rho_\lambda(V)$ of Section~\ref{sec:panels}. If $G(V)$ has no edge, then $\tau_\lambda(V)$, $\rho_\lambda(V)$, $\kappa_\lambda(V)$ and $\pi_\lambda(V)$ are all $0$, so assume that it has one; then $\rho_\lambda(V)>0$.

\paragraph{Step 1: a layer cake with two levels.} For $a\ge b$ the edge cost is $f_K(a,b)=b+\min\{a,\lambda b\}$. Given a prior $\mu$ on $V$ and $u>0$, let $A_u=\{v:\mu_v\ge u\}$ and $B_u=\{v:\mu_v\ge u/\lambda\}$, and let $e(A_u,B_u)$ count the ordered pairs of adjacent patterns in $A_u\times B_u$. An edge with weights $a\ge b$ is counted from its heavier endpoint when $u\le\min\{a,\lambda b\}$ and from its lighter endpoint when $u\le b$, so the objective of $\mu$ equals $\int_0^\infty e(A_u,B_u)\,du$. Every edge joins a pattern with an even number of ones to one with an odd number, so the pairs counted by $e(A_u,B_u)$ are the edges between $A_u\cap V^{\mathrm{ev}}$ and $B_u\cap V^{\mathrm{odd}}$ and those between $A_u\cap V^{\mathrm{odd}}$ and $B_u\cap V^{\mathrm{ev}}$, and $e(A_u,B_u)\le\rho_\lambda(V)(|A_u|+|B_u|/\lambda)$. Since $\int_0^\infty|A_u|\,du=1$ and $\int_0^\infty|B_u|\,du=\lambda$, this gives $\tau_\lambda(V)\le2\rho_\lambda(V)$. Conversely, the prior with mass $\theta$ on each vertex of $A$ and $\theta/\lambda$ on each vertex of $W$, $\theta=1/(|A|+|W|/\lambda)$, earns $f_K(\theta,\theta/\lambda)=K\theta/\lambda$ on each edge between them, so $\tau_\lambda(V)\ge(K/\lambda)\rho_\lambda(V)$, and such two-level priors are optimal up to the factor $2\lambda/K<2$.

\paragraph{Step 2: peeling to a weighted core.} Let $(A,W)$ attain $\rho=\rho_\lambda(V)>0$ and repeatedly delete a vertex of $A$ with fewer than $\rho$ neighbors in $W$ or a vertex of $W$ with fewer than $\rho/\lambda$ neighbors in $A$, as in the peeling algorithm of \citet{Cha00}. Each deletion removes fewer edges than $\rho$ times its weight ($1$ in $A$, $1/\lambda$ in $W$), while the edges number exactly $\rho$ times the total weight at the start, so the process stops at a two-sided core with $\delta_A\ge\rho$ and $\lambda\delta_W\ge\rho$. Hence $\rho_\lambda\le\kappa_\lambda$. Conversely, a two-sided core of value $\kappa$ has $|E(A,W)|\ge\kappa\max\{|A|,|W|/\lambda\}$, so $\kappa_\lambda\le2\rho_\lambda$.

\paragraph{Step 3: from dense pairs to panel systems.} Let $(A,W)$ attain $\rho=\rho_\lambda(V)>0$. Each edge between $A$ and $W$ points upward, with its $W$-endpoint farther from the center in Hamming distance, for exactly one of the two centers $0$ and $\one_U$, since $|v\oplus\one_U|=|U|-|v|$; fix the center $c\in\{0,\one_U\}$ for which at least half of the edges point upward, and group these edges by the distance $j$ of their $W$-endpoint from $c$. The patterns $A_{j-1}\subseteq A$ at distance $j-1$ and $W_j\subseteq W$ at distance $j$ are disjoint for different $j$, so some $j$ with at least one such edge keeps at least $\frac\rho2(|A_{j-1}|+|W_j|/\lambda)$ edges between them. Peeling as in Step~2 leaves nonempty $A'\subseteq A_{j-1}$ and $W'\subseteq W_j$ with minimum degrees at least $\rho/2$ and $\rho/(2\lambda)$ between them. Since $u\in A'$ and $w\in W'$ are adjacent exactly when $\supp(u\oplus c)\subset\supp(w\oplus c)$, these supports form a panel system of rank $j$ that $V$ shatters with center $c$, with $a\ge\rho/2$ and $b\ge\rho/(2\lambda)$ (as in Figure~\ref{fig:hero}(c)), so $\rho_\lambda\le2\pi_\lambda$.

\paragraph{Conclusion.} A shattered panel system is a two-sided core, so $\pi_\lambda\le\kappa_\lambda$, and Steps~1 to~3 give
\[
\tfrac K{2\lambda}\,\pi_\lambda(V)\ \le\ \tfrac K{2\lambda}\,\kappa_\lambda(V)\ \le\ \tfrac K\lambda\,\rho_\lambda(V)\ \le\ \tau_\lambda(V)\ \le\ 2\rho_\lambda(V)\ \le\ \min\{2\kappa_\lambda(V),\,4\pi_\lambda(V)\}
\]
and $\kappa_\lambda\le2\rho_\lambda\le4\pi_\lambda$. Maximizing over $|U|\le m$ gives the bounds on $F_\lambda(\Hc,m)$, and with $F_\lambda\ge\frac12\pi_\lambda$ and $F_\lambda\ge\frac12\kappa_\lambda$ the transfer~\eqref{eq:transfer} gives those on $M_n(\Hc,\lambda)$. \jmlrQED

\subsection{Panel systems on natural classes}\label{sec:ov-panels}

Panel systems certify the rate in both directions. A shattered panel system carries the two-level prior of Step~1, which gives $F_\lambda\ge\frac K{2\lambda}\min\{a,\lambda b\}$, and an upper bound follows once the class shatters no panel system in which $a$ and $\lambda b$ are both large. For the class as a whole, with $F_\lambda(\Hc)=\sup_mF_\lambda(\Hc,m)$, $1\le d<\infty$, $t\ge1/2$ and $g\ge1$, if $F_\lambda(\Hc)\ge Kd/g$ then $\Hc$ shatters a panel system with $a\ge Kd/(4g)$ and $b\ge d/(4g)$, and a shattered panel system with $a\ge Kd/g$ and $b\ge d/g$ gives $F_\lambda(\Hc)\ge Kd/(2g)$ (Corollary~\ref{cor:B-criterion}).

\paragraph{Affine halfspaces.} Let $d\ge2$, $j=\lfloor d/2\rfloor$, and let $U$ be a set of $m\ge j$ points $\gamma(s)=(s,s^2,\dots,s^d)$ of the moment curve in $\Q^d$. For a set $A$ of at most $j$ of these points, the polynomial $-\prod_{\gamma(s)\in A}(x-s)^2$ has degree at most $2j\le d$, so it is an affine function of $\gamma(x)$ that vanishes on $A$ and is negative at the other points of $U$. Hence affine halfspaces cut out every set of at most $j$ points of $U$, which is the $j$-neighborliness of cyclic polytopes \citep{Gal63}, and the $(j-1)$-subsets and the $j$-subsets of $U$ form a panel system with $a=m-j+1$ and $b=j$ that the class shatters with center $0$. For $m\ge jK$ its value is $\lambda j$, so Theorem~\ref{thm:binary}(b) gives $F_\lambda(\Hc,m)\ge jK/2\ge(d+1)K/8$ for affine halfspaces. With $F_\lambda\le Kd_m\le(d+1)K$, this determines $F_\lambda$ up to the factor $8$ for $m\ge jK$, hence the rate up to absolute constants for $n+1\ge jK$, and the gain is of constant order.

\paragraph{Dictators and the order $\log K$.} For dictators the number of patterns limits the shattered panel systems, which produces the gain of order $\log K$ in Corollary~\ref{cor:gain}(ii). The count of patterns enters through a bound that holds for every set of patterns.

\begin{lemma}[Entropy bound]\label{lem:entropy}
For every finite nonempty $V\subseteq\{0,1\}^U$ and $\lambda\ge1$, $\tau_\lambda(V)\le\log_2|V|/h(1/K)\le K\log_2|V|/\log_2K$, where $h$ is the binary entropy in bits.
\end{lemma}

\noindent\emph{Proof.} Let $X$ have law $\mu\in\Pc(V)$. Given the other coordinates, $X_x$ is random only on an edge $\{u,w\}$ of $G(V)$ in direction $x$. Thus $\sum_{x\in U}H(X_x\mid X_{U\setminus\{x\}})=\sum_{\{u,w\}}(\mu_u+\mu_w)\,h(q_{uw})$ over the edges of $G(V)$ with $\mu_u+\mu_w>0$, where $q_{uw}=\min\{\mu_u,\mu_w\}/(\mu_u+\mu_w)$. By Han's inequality \citep{Han78} this sum is at most $H(X)\le\log_2|V|$. As $h$ is concave with $h(0)=0$ and increasing on $[0,1/2]$, on these edges the cost $f_K(\mu_u,\mu_w)=(\mu_u+\mu_w)\min\{1,Kq_{uw}\}$ is at most $(\mu_u+\mu_w)h(q_{uw})/h(1/K)$, as $K\ge2$, and the other edges cost $0$. Summing over the edges and maximizing over $\mu$ gives the first bound, and $h(1/K)\ge\log_2K/K$ gives the second. \jmlrQED

Dictators on $p\ge2$ features are $p+1$ functions of VC dimension $\lfloor\log_2(p+1)\rfloor$, so Lemma~\ref{lem:entropy} gives $F_\lambda(\Hc,m)\le2K\log_2p/\log_2K$. They realize every $V\subseteq\{0,1\}^r$ with at most $p$ nonzero patterns $v^1,\dots,v^q$ in which no two coordinates agree on all patterns, on the points $z_k$ with $(z_k)_i=v^i(k)$ for $i\le q$ and $0$ otherwise. For $r=\Lambda j$ with $\Lambda=\lceil\lambda\rceil+1$, the $(j-1)$-subsets and the $j$-subsets of $r$ points form a panel system with $a=r-j+1\ge\lambda j$ and $b=j$, of value $\lambda j$ on at most $2(e\Lambda)^j$ patterns, which dictators shatter with center $0$ when $2(e\Lambda)^j\le p$ and $r\le m$. With the star on $\min\{m,p\}$ points this gives $F_\lambda(\Hc,m)\ge\frac1{33}\min\{m,p,K\log_2p/\log_2K\}$. By Theorem~\ref{thm:binary}(b) and Lemma~\ref{lem:entropy}, a projection that shatters a panel system of value $\lambda j$ has at least $K^{j/2}$ patterns, while the Hamming layers need at most $2(e\Lambda)^j$, so $p+1$ patterns allow values of order $K\log p/\log K$ once $K\log_2p/\log_2K\le\min\{p,m\}$. In this regime, at $m=n+1$, dividing $K\lfloor\log_2(p+1)\rfloor$ by $F_\lambda$ gives the gain of order $\log K$.

\paragraph{Group-sparse supports.} For the class of Section~\ref{sec:panels}, order the patterns of $\Hc|_U$ into layers by the number of groups they meet. A pattern $\one_S$ in layer $j$ has at most $jL\le kL$ neighbors in layers up to $j$, namely the patterns $\one_{S\triangle\{x\}}$ with $x$ in one of its groups, and at most $j\le k$ neighbors in layer $j-1$, namely the patterns $\one_{S\setminus\{x\}}$ with $x$ alone in its group. In a two-sided core with $\delta_A>kL$ and $\delta_W>k$, no vertex of $A$ lies in the highest layer that the core meets, and then no vertex of $W$ does either; removing that layer leaves a core with the same degrees, so no such core exists, and every shattered panel system has $a\le kL$ or $b\le k$. Hence $\pi_\lambda(\Hc,m)\le\kappa_\lambda(\Hc,m)\le\min\{m,k\max\{L,\lambda\}\}$. With $\Psi(m):=\min\{m,N,k\max\{L,K\}\}$, $N\ge2$ and $t\ge1/2$, this bound and the lower bounds through the VC dimension and the $k$-sparse subclass place the minimax risk between $\Psi(n+1)/(6e(n+1))$ and $4\Psi(n+1)/(n+1)$, and when $N,n+1\ge k\max\{L,K\}$ the gain satisfies $\min\{K,L\}/(12e)\le B_n(\Hc,\lambda)\le4e\min\{K,L\}$, so every order between a constant and $K$ occurs. The same layers give an explicit learner, which abstains on a new key whose group contains a positive example and answers $0$ on every other new key; its leave-one-out cost is at most $k(L+\lambda)$, so $M_n(\Hc,\lambda)\le2\Psi(n+1)/(n+1)$. If $\min\{K,L\}/\log_2K\to\infty$ and $N,n+1\ge k\max\{L,K\}\log_2K$, the bounds $F_\lambda\le Kd_m$ and $F_\lambda\le s$ and the entropy bound exceed $F_\lambda(\Hc,n+1)$ by unbounded factors (proofs in Appendices~\ref{app:criterion}, \ref{app:panelexamples} and~\ref{app:applications}).

\subsection{Small projections of complexes with large systole}\label{sec:ov-plateau}

Theorem~\ref{thm:noparam} needs classes of VC dimension $d$ whose minimax risk is at most of order $\max\{d,\lambda\}/n$ for all $n\le2r$ and, when the complexes are thick enough, at least of order $\lambda d/n$ once $n$ exceeds the size of the domain, which requires a bound on every projection on at most $2r+1$ points. A finite simplicial complex $Y$ is a family of finite sets, its faces, that is closed under taking subsets, and its indicator class is $\Hc_Y=\{\one_F:F\in Y\}$. It is flag if every clique of its $1$-skeleton $G_Y$ is a face. A subcomplex is full if it contains every face of $Y$ whose vertices it contains, and the systole $\sys(Y)$ is the smallest number of edges of a full subcomplex homeomorphic to a circle \citep{JS06,ABJLMS09}. A graph is chordal if it has no induced cycle of length at least $4$. The witnesses are the indicator classes of complexes with four properties.

\begin{lemma}[Thick flag complexes with large systole]\label{lem:complexes}
For all integers $k,q,r\ge2$ there is a finite flag simplicial complex $X_{k,q,r}$ whose largest faces have $k$ vertices, in which every $(k-1)$-subset of a face with $k$ vertices lies in at least $q$ faces with $k$ vertices, and whose systole is at least $2(r+1)$.
\end{lemma}

\noindent The complex $X_{k,q,r}$ is the development of a simplex of groups of \citet[Section~3]{ABJLMS09} with the cyclic group $\Z_q$ on each face of codimension one, and Lemma~\ref{lem:complex} derives the four properties from the facts listed in Section~\ref{sec:C-complex}. For $k=2$, $X_{2,q,r}$ is a triangle-free graph of girth at least $2(r+1)$ in which every endpoint of an edge has degree at least $q$ (Remark~\ref{rem:C-k2}). It induces a forest on every set of at most $2r+1$ vertices, where Lemma~\ref{lem:small} bounds the core number by $\max\{2,\lambda\}$, while on all vertices the singletons and the edges form a panel system with $a\ge q$ and $b=2$. This class has VC dimension at most $2$, so $\kappa_\lambda(\Hc,m)\ge\min\{m,\bar\kappa_\lambda(\Hc)\}/4$ for $\lambda\ge1$ (Proposition~\ref{prop:kappaglobal}); the gap of order $\min\{\lambda,d\}$ needed for Theorem~\ref{thm:noparam} comes from faces with $d$ vertices.

\begin{lemma}[Small projections]\label{lem:small}
Let $Y$ be a finite flag complex whose faces have at most $k\ge1$ vertices, and let $U$ be a set of fewer than $\sys(Y)$ vertices. Then $G_Y[U]$ is chordal, $\Hc_Y|_U$ consists of the indicators of the cliques of $G_Y[U]$, every two-sided core of $G(\Hc_Y|_U)$ has $\delta_A\le k$ or $\delta_W\le1$, and $\kappa_\lambda(\Hc_Y|_U)\le\max\{k,\lambda\}$ for every $\lambda\ge0$.
\end{lemma}

\begin{proof}
An induced cycle of $G_Y[U]$ of length $\ell\ge4$ would span a full subcomplex that is a circle with $\ell\le|U|<\sys(Y)$ edges, so $G_Y[U]$ is chordal, and as $Y$ is flag its faces inside $U$ are the cliques of $G_Y[U]$ (Lemma~\ref{lem:systole}). For $U\ne\emptyset$, the chordal graph has a vertex $v$ with pairwise adjacent neighbors \citep{Dir61}. Every clique $F\ni v$ lies in the clique of $v$ and its neighbors, of at most $k$ vertices, so $F$ has at most $k$ neighbors in $G(\Hc_Y|_U)$, all but $F\setminus\{v\}$ containing $v$. In a two-sided core with $\delta_A\ge k+1$ and $\delta_W\ge2$, no member of $A$ contains $v$, and so no member of $W$ does, as it would have at most one neighbor in $A$. It is then a core of the clique complex of $G_Y[U\setminus\{v\}]$, which induction on $|U|$ excludes (Lemma~\ref{lem:chordalcore}). So every core has $\min\{\delta_A,\lambda\delta_W\}\le\max\{k,\lambda\}$.
\end{proof}

\noindent The indicator class $\Hc_{k,q,r}$ of $X_{k,q,r}$ has VC dimension $k$, Lemma~\ref{lem:small} gives $\kappa_\lambda(\Hc_{k,q,r},m)\le\max\{k,\lambda\}$ for $m\le2r+1$, and on all vertices the $(k-1)$-faces and $k$-faces form a panel system with $a\ge q$ and $b=k$ (Proposition~\ref{prop:plateau}). By Theorem~\ref{thm:binary}(b), $(n+1)M_n(\Hc_{d,q,r},\lambda)\le2\max\{d,\lambda\}$ for $n\le2r$, while for $q\ge d+1$ the domain has $N_0\ge2r+2$ points and $N_0M_{N_0-1}(\Hc_{d,q,r},\lambda)\ge dKq/(e(d\lambda+q))$ (Proposition~\ref{prop:plateau}). For large $r$, admissibility at $n=2r$ gives $D(\Hc_{d,q,r})\le2\max\{d,\lambda\}/c$, and at $n=N_0-1$ it forces $2C\max\{d,\lambda\}/c\ge dKq/(e(d\lambda+q))$, so $q\to\infty$ gives Theorem~\ref{thm:noparam}(a). Part~(b) is~(a) with $d\ge\lambda$, and part~(c) combines Theorem~\ref{thm:binary}(b) with $\kappa_\lambda(\Hc,m)\ge\min\{m,\bar\kappa_\lambda(\Hc)\}/(2\min\{\lambda,d\})$.

\section{Open directions}\label{sec:open}

Above $t=1/2$, the largest value of a shattered panel system on at most $n+1$ points determines the minimax risk of a nontrivial class within absolute constants, and these panel systems decide how much abstention removes from the cost $1/(1-t)$ of an error, from a factor of order $1/(1-t)$ to a constant on natural classes. First, which classes, such as monotone conjunctions, have a rule with risk of order $\pi_\lambda(\Hc,n+1)/n$ that avoids the linear program behind~\eqref{eq:transfer}, as the explicit rule for group-sparse supports does (Section~\ref{sec:ov-panels})? Second, which profiles $m\mapsto\pi_\lambda(\Hc,m)$ occur for classes of a given VC dimension? Third, which weighted panel systems determine the minimax risk under a fixed marginal distribution?

\label{main-text-end}%

\bibliography{references}

\appendix
\counterwithin{theorem}{section}
\providecommand{\ThmMain}{Theorem~\ref{thm:binary}}
\providecommand{\ThmNoParam}{Theorem~\ref{thm:noparam}}
\providecommand{\CorClasses}{Corollary~\ref{cor:natural}}
\providecommand{\CorGain}{Corollary~\ref{cor:gain}}
\providecommand{\PropFull}{Proposition~\ref{prop:exact}}
\providecommand{\LemComplex}{Lemma~\ref{lem:complexes}}
\providecommand{\LemChordal}{Lemma~\ref{lem:small}}
\providecommand{\EqTransfer}{\eqref{eq:transfer}}
\providecommand{\mm}{\mathrm{mm}}

\section{Setting, the leave-one-out value and the transfer}\label{sec:A}\label{app:minimax}

This section fixes the notation of the appendix, proves the basic properties of the leave-one-out value and proves the transfer~\eqref{eq:transfer}.

\subsection{Setting}\label{sec:A-setting}

We use the setting of Section~\ref{sec:loss}: $\X$ is a nonempty countable set, $\Hc\subseteq\{0,1\}^{\X}$ a nonempty class, $\lambda\in[0,\infty)$, $K=1+\lambda$, $t=\lambda/(1+\lambda)$, and $\ell_\lambda(\nu,y)=\nu(\abst)+K\nu(1-y)$ for $\nu\in\Pc(\{0,1,\abst\})$ and $y\in\{0,1\}$. A rule is a map $g:\X\to\Pc(\{0,1,\abst\})$ with risk $R_\lambda(g;P,h)=\sum_xP(\{x\})\ell_\lambda(g(x),h(x))$, and a learner maps every sample $S\in(\X\times\{0,1\})^n$ to a rule $g_S$. With $S_h(x_{1:n})=((x_i,h(x_i)))_{i\le n}$ we write $\Risk^\lambda_n(g;P,h)=\E_{x_{1:n}\sim P^n}R_\lambda(g_{S_h(x_{1:n})};P,h)$, so that $M_n(\Hc,\lambda)=\inf_g\sup_{P,h\in\Hc}\Risk^\lambda_n(g;P,h)$. As $\X$ is countable, $\Risk^\lambda_n(g;P,h)$ is a countable sum of nonnegative terms, and no measurability condition is needed. A learner with internal randomness, independent of the data, has the risk of its average over that randomness because $\ell_\lambda(\cdot,y)$ is affine, so randomized learners are covered. At $\lambda=\infty$, $R_\infty(g;P,h)\ge R_\lambda(g;P,h)$ for every rule and every finite $\lambda$, since both sides agree when the probability of error is $0$; hence $M_n(\Hc,\infty)\ge M_n(\Hc,\lambda)$. The geometric classes are defined on $\Q^d$, and the lower bounds use finitely supported distributions.

A learner never abstains if $g_S(x)(\abst)=0$ for all $S$ and $x$, and $M^{\mathrm{na}}_n(\Hc,\lambda)$ is the infimum defining $M_n(\Hc,\lambda)$ restricted to such learners. For nontrivial $\Hc$ the abstention gain $B_n(\Hc,\lambda)=M^{\mathrm{na}}_n(\Hc,\lambda)/M_n(\Hc,\lambda)$ is well defined since $M_n(\Hc,\lambda)>0$ (Proposition~\ref{prop:transfer}), and it equals $(1+\lambda)M_n(\Hc,0)/M_n(\Hc,\lambda)$ (Lemma~\ref{lem:nonabst}). For the other classes both minimax risks vanish by the next lemma, and $B_n(\Hc,\lambda):=1$.

\begin{lemma}\label{lem:A-trivial}
$\Hc$ is not nontrivial if and only if $\Hc\subseteq\{h,1-h\}$ for some $h\in\Hc$. In that case some learner never errs and never abstains, so $M_n(\Hc,\lambda)=M^{\mathrm{na}}_n(\Hc,\lambda)=0$ for all $n\ge1$ and $\lambda\in[0,\infty)$, and $M_n(\Hc,\infty)=0$.
\end{lemma}

\begin{proof}
Two members of $\{h,1-h\}$ are equal or disagree everywhere. If $h'\in\Hc\setminus\{h,1-h\}$, then $h'\ne h$ and $h'\ne1-h$ give points $x',x$ with $h(x')\ne h'(x')$ and $h(x)=h'(x)$. If $\Hc\subseteq\{h,1-h\}$, the learner that outputs $\delta_{h(x)}$ if $y_1=h(x_1)$ and $\delta_{1-h(x)}$ otherwise returns the target.
\end{proof}

\subsection{One-inclusion graphs and the leave-one-out value}\label{sec:A-graphs}

\paragraph{One-inclusion graphs.} Let $U$ be a finite set and $V\subseteq\{0,1\}^U$ nonempty. For $v\in\{0,1\}^U$ and $x\in U$, $v^{(x)}$ is $v$ with coordinate $x$ flipped, $|v|$ is the number of ones of $v$, $\oplus$ is coordinatewise addition modulo $2$ and $\one_S$ is the indicator of $S\subseteq U$. The edges of $G(V)$ are the pairs $\{v,v^{(x)}\}\subseteq V$, and $x$ is the direction of the edge; every vertex has at most one edge in each direction. $E(V)$ is the edge set, $N_V(v)$ the set of neighbors of $v$, $\deg_V(v)=|N_V(v)|\le|U|$ and $\Delta(V)=\max_v\deg_V(v)$. For $A\subseteq V$, $E_V(A)$ is the set of edges inside $A$, and for disjoint $A,W\subseteq V$, $E_V(A,W)$ is the set of edges between $A$ and $W$. The VC dimension $d_V$ is the largest $|T|$ with $T\subseteq U$ and $V|_T=\{0,1\}^T$. For a finite $U\subseteq\X$, $\Hc|_U=\{h|_U:h\in\Hc\}$; since subsets of shattered sets are shattered, some $T$ with $|T|=d_m$ is shattered by $\Hc$, and $d_{\Hc|_U}\le d_m$ if $|U|\le m$. The star number is $s=\sup_U\Delta(\Hc|_U)$ over finite $U\subseteq\X$: a vertex of degree $D$ in $G(\Hc|_U)$ and the directions of its edges form a star set of size $D$ in the sense of \citet{HY15}, and conversely \citep{Han24}.

\paragraph{The value.} With $f_K(a,b)=\min\{a+b,Ka,Kb\}$ and $\mu\in\Pc(V)$, let $\Phi_\lambda(\mu;V)=\sum_{\{u,w\}\in E(V)}f_K(\mu_u,\mu_w)$, so that $\tau_\lambda(V)=\max_\mu\Phi_\lambda(\mu;V)$, the maximum of a continuous function on a compact set, and $\tau_\lambda(V)=0$ if $E(V)=\emptyset$. The maximum defining $F_\lambda(\Hc,m)$ exists by Lemma~\ref{lem:basic}, and $F_\lambda(\Hc)=\sup_mF_\lambda(\Hc,m)$. The limits $\tau_\infty(V)$ and $F_\infty(\Hc,m)$ as $\lambda\to\infty$ exist by Lemma~\ref{lem:basic}.

\paragraph{Densities, cores and panel systems.} The bipartite density $\rho_\lambda(V)$ for $\lambda\in(0,\infty)$, two-sided cores $(A,W)$ with their minimum degrees $\delta_A,\delta_W$, the core number $\kappa_\lambda(V)$, panel systems $(\Ac,\Wc)$ with their parameters $a(\Ac,\Wc),b(\Ac,\Wc)$, shattering with a center and $\pi_\lambda(V)$ are defined in Section~\ref{sec:panels}; we write $E_V(A,W)$ for $E(A,W)$. If $E(V)\ne\emptyset$, $\kappa_\lambda(V)$ is a maximum over a nonempty finite set, since an edge $\{u,w\}$ gives the core $(\{u\},\{w\})$. For a class, $\pi_\lambda(\Hc,m)=\max_{|U|\le m}\pi_\lambda(\Hc|_U)$, $\bar\kappa_\lambda(\Hc)=\sup_m\kappa_\lambda(\Hc,m)$ and $\bar\pi_\lambda(\Hc)=\sup_m\pi_\lambda(\Hc,m)$.

\begin{lemma}[Edge costs]\label{lem:fK}
Let $a,b,c\ge0$, $1\le K\le K'$ and $\lambda=K-1$.
\begin{enumerate}
\item[(i)] $f_K$ is symmetric, $0\le f_K(a,b)\le a+b$, $f_K(a,0)=0$ and $f_K(ca,cb)=c\,f_K(a,b)$.
\item[(ii)] $\min\{a,b\}=f_1(a,b)\le f_K(a,b)\le f_{K'}(a,b)$, and $f_K(a,b)/K\ge f_{K'}(a,b)/K'$.
\item[(iii)] If $K\le2$, then $f_K(a,b)=K\min\{a,b\}$.
\item[(iv)] If $a\ge b$, then $f_K(a,b)=b+\min\{a,\lambda b\}$.
\item[(v)] For every $\nu\in\Pc(\{0,1,\abst\})$, $a\,\ell_\lambda(\nu,0)+b\,\ell_\lambda(\nu,1)\ge f_K(a,b)$, with equality for one of $\nu=\delta_\abst,\delta_0,\delta_1$.
\end{enumerate}
\end{lemma}

\begin{proof}
(i) and the first part of (ii) hold because $a+b$, $Ka$ and $Kb$ are nonnegative, homogeneous of degree $1$ and nondecreasing in $K$, and $\min\{a+b,a,b\}=\min\{a,b\}$; and $f_K(a,b)/K=\min\{(a+b)/K,a,b\}$ is nonincreasing in $K$. (iii) $a+b\ge2\min\{a,b\}\ge K\min\{a,b\}$. (iv) If $a\ge b$, then $f_K(a,b)=\min\{a+b,Kb\}=b+\min\{a,\lambda b\}$. (v) The left side is affine in $\nu$, so its minimum is attained at a vertex, and $\delta_\abst$, $\delta_0$, $\delta_1$ give $a+b$, $Kb$, $Ka$.
\end{proof}

\begin{lemma}[Basic facts]\label{lem:basic}
Let $V\subseteq\{0,1\}^U$ be finite and nonempty, and let $m\ge0$.
\begin{enumerate}
\item[(i)] $\tau_\lambda(V)\le\Delta(V)\le|U|$.
\item[(ii)] $\tau_\lambda(V)$ and $F_\lambda(\Hc,m)$ are nondecreasing in $\lambda\in[0,\infty)$, and $\tau_\lambda(V)/(1+\lambda)$ and $F_\lambda(\Hc,m)/(1+\lambda)$ are nonincreasing. So $\tau_0(V)\le\tau_\lambda(V)\le(1+\lambda)\tau_0(V)$, and $(1+\lambda)F_0(\Hc,m)/F_\lambda(\Hc,m)$ is nondecreasing in $\lambda$ when $F_0(\Hc,m)>0$.
\item[(iii)] The maxima defining $F_\lambda(\Hc,m)$, $\kappa_\lambda(\Hc,m)$, $\pi_\lambda(\Hc,m)$ and $\max_{|U|\le m}\Delta(\Hc|_U)$ are over finitely many values, all four are nondecreasing in $m$, and $F_\lambda(\Hc,m)\le m$.
\item[(iv)] If $\lambda\ge1$, then $\Delta(V)K/(\lambda+\Delta(V))\le\tau_\lambda(V)$. Hence $\tau_\infty(V)=\Delta(V)$, $F_\infty(\Hc,m)=\max_{|U|\le m}\Delta(\Hc|_U)$, and $F_\lambda(\Hc,m)\le F_\infty(\Hc,m)$ for every $\lambda\in[0,\infty)$.
\end{enumerate}
\end{lemma}

\begin{proof}
(i) By Lemma~\ref{lem:fK}(i), $\Phi_\lambda(\mu;V)\le\sum_{\{u,w\}\in E(V)}(\mu_u+\mu_w)=\sum_v\mu_v\deg_V(v)\le\Delta(V)$. (ii) By Lemma~\ref{lem:fK}(ii), for each $\mu$, $\Phi_\lambda(\mu;V)$ is nondecreasing and $\Phi_\lambda(\mu;V)/K$ nonincreasing in $\lambda$, and so are their maxima over $\mu$ and $U$. (iii) Renaming the points of $U$ by $1,\dots,|U|$ preserves the one-inclusion graph and the panel systems with their centers, so the four quantities of $\Hc|_U$ depend only on a subset of $\{0,1\}^k$ with $k=|U|\le m$, and there are finitely many. Monotonicity holds by definition, and $F_\lambda(\Hc,m)\le m$ by (i). (iv) Let $D=\Delta(V)$; if $D=0$, then $\tau_\lambda(V)=0$ for all $\lambda$. Otherwise let $c$ have degree $D$, and put mass $\lambda/(\lambda+D)$ on $c$ and $1/(\lambda+D)$ on each neighbor of $c$. As $\lambda\ge1$, Lemma~\ref{lem:fK}(iv) gives each edge at $c$ the cost $K/(\lambda+D)$. So $\tau_\lambda(V)$, nondecreasing by (ii), lies between $DK/(\lambda+D)$ and $D$ and tends to $D$. By (iii), $F_\lambda(\Hc,m)$ is a maximum of finitely many nondecreasing functions of $\lambda$ with limits $\Delta(\Hc|_U)$, which gives the rest.
\end{proof}

\paragraph{Rules and priors.} A rule on $V$ is a family $q=(q_e)_{e\in E(V)}$ with $q_e\in\Pc(\{0,1,\abst\})$, used on the edge $e$ when the coordinate $x_e$ of its direction is held out. For $v\in e$ the cost of $q_e$ to $v$ is $c_v(q_e)=\ell_\lambda(q_e,v(x_e))$, and the leave-one-out cost of $v$, summed over the held-out coordinates, is $L_q(v)=\sum_{e\ni v}c_v(q_e)$.

\begin{lemma}[Duality]\label{lem:lp}
Let $V\subseteq\{0,1\}^U$ be finite and nonempty and $\lambda\in[0,\infty)$.
\begin{enumerate}
\item[(i)] For every rule $q$ and $\mu\in\Pc(V)$, $\sum_v\mu_vL_q(v)=\sum_{e=\{u,w\}\in E(V)}(\mu_uc_u(q_e)+\mu_wc_w(q_e))\ge\Phi_\lambda(\mu;V)$, with equality for some rule. So $\Phi_\lambda(\mu;V)$ is the smallest Bayes cost of a rule under the prior $\mu$.
\item[(ii)] $\min_q\max_{v\in V}L_q(v)=\tau_\lambda(V)$. In particular some rule $q^V$ has $L_{q^V}(v)\le\tau_\lambda(V)$ for every $v\in V$.
\end{enumerate}
\end{lemma}

\begin{proof}
(i) The identity regroups the sum by edges. For $e=\{u,w\}$ with $u(x_e)=0$ and $w(x_e)=1$, the term of $e$ is $\mu_u\ell_\lambda(q_e,0)+\mu_w\ell_\lambda(q_e,1)\ge f_K(\mu_u,\mu_w)$ by Lemma~\ref{lem:fK}(v), with equality for a suitable point mass $q_e$. (ii) The map $\Psi(q,\mu)=\sum_v\mu_vL_q(v)$ is continuous and affine in each argument on the product of the compact convex sets $\Pc(\{0,1,\abst\})^{E(V)}$ and $\Pc(V)$. By the minimax theorem \citep{Sio58}, or linear programming duality \citep{Sch86}, $\min_q\max_\mu\Psi=\max_\mu\min_q\Psi$, with both extrema attained. Since $\max_\mu\Psi(q,\mu)=\max_vL_q(v)$ and $\min_q\Psi(q,\mu)=\Phi_\lambda(\mu;V)$ by (i), $\min_q\max_vL_q(v)=\max_\mu\Phi_\lambda(\mu;V)=\tau_\lambda(V)$.
\end{proof}

\subsection{The transfer}\label{sec:A-transfer}

The upper bound in~\eqref{eq:transfer} is the leave-one-out argument of \citet{HLW94}, stated by \citet[Remark~2.2]{ACSZ23} for every loss with values in $[0,1]$, here $\ell_\lambda/K$, and the lower bound follows the comparison of transductive and expected risks of \citet[Proposition~32]{ADDST24}. We write out the parts specific to $\ell_\lambda$: the learner built from the rules of Lemma~\ref{lem:lp}, the pointwise leave-one-out bound, and the Bayes priors.

\paragraph{The learners.} For every finite $U\subseteq\X$ fix a rule $q^U=q^{\Hc|_U}$ as in Lemma~\ref{lem:lp}(ii); the finite subsets of $\X$ form a countable family. Given $S=((x_i,y_i))_{i\le n}$ and $x\in\X$, let $U=\{x_1,\dots,x_n\}\cup\{x\}$ and $C=\{v\in\Hc|_U:v(x_i)=y_i\text{ for all }i\}$. If $C\ne\emptyset$ and all $v\in C$ take the same value $v(x)$, the learners $g^\lambda$ and $g^\infty$ output $\delta_{v(x)}$, and if $C=\emptyset$ both abstain. Otherwise $C$ contains two patterns that differ at $x$; since all members of $C$ agree on $U\setminus\{x\}$, $C$ is an edge of $G(\Hc|_U)$ in direction $x$, and $g^\lambda$ outputs $q^U_C$ while $g^\infty$ abstains. Both learners depend on $S$ only through the set $\{(x_i,y_i):i\le n\}$. On a sample labeled by $h^\star\in\Hc$, $h^\star|_U\in C$, so $g^\infty$ never errs.

\begin{lemma}[Pointwise leave-one-out bound]\label{lem:A-loo}
Let $h^\star\in\Hc$, $z=(z_1,\dots,z_{n+1})\in\X^{n+1}$, $U=\{z_1,\dots,z_{n+1}\}$ and $S^{-j}=((z_i,h^\star(z_i)))_{i\ne j}$. Then $\sum_{j=1}^{n+1}\ell_\lambda\big(g^\lambda_{S^{-j}}(z_j),h^\star(z_j)\big)\le\tau_\lambda(\Hc|_U)$, and $g^\infty$ abstains on $(S^{-j},z_j)$ for at most $\Delta(\Hc|_U)$ indices $j$.
\end{lemma}

\begin{proof}
Let $v^\star=h^\star|_U$. On input $(S^{-j},z_j)$ the learners use the set $U$ and the consistent set $C_j=\{v\in\Hc|_U:v(z_i)=h^\star(z_i),\ i\ne j\}\ni v^\star$. If $z_j=z_i$ for some $i\ne j$, all members of $C_j$ take the value $h^\star(z_j)$ at $z_j$. Otherwise $C_j\subseteq\{v^\star,v^{\star(z_j)}\}$. So both learners output $\delta_{h^\star(z_j)}$ unless $j\in J$, the set of indices with $z_j\ne z_i$ for all $i\ne j$ and $v^{\star(z_j)}\in\Hc|_U$; for $j\in J$, $C_j$ is the edge $e_j=\{v^\star,v^{\star(z_j)}\}$, and these edges at $v^\star$ are distinct because the points $z_j$, $j\in J$, are. Hence the sum is $\sum_{j\in J}c_{v^\star}(q^U_{e_j})\le L_{q^U}(v^\star)\le\tau_\lambda(\Hc|_U)$, and $|J|\le\deg(v^\star)\le\Delta(\Hc|_U)$.
\end{proof}

\noindent For $x_0\in\X$ and $b\in\{0,1\}$ let $\Hc^b_{x_0}=\{h\in\Hc:h(x_0)=b\}$.

\begin{lemma}[Splitting at a point]\label{lem:A-split}
Let $U\subseteq\X$ be finite, $x_0\in U$, $V=\Hc|_U$, $c_K=\min\{1,K/2\}$ and $V_b=\Hc^b_{x_0}|_{U\setminus\{x_0\}}$ for $b\in\{0,1\}$. Then $\tau_\lambda(V)\le\max_{b:\,V_b\ne\emptyset}\tau_\lambda(V_b)+c_K$.
\end{lemma}

\begin{proof}
Let $V^b=\{v\in V:v(x_0)=b\}$; restriction to $U\setminus\{x_0\}$ is a bijection $r_b:V^b\to V_b$. The edges in a direction $x\ne x_0$ lie inside some $V^b$, and $r_b$ maps those inside $V^b$ bijectively onto $E(V_b)$. The edges in direction $x_0$ form a matching. Fix $\mu\in\Pc(V)$ and let $M_b=\mu(V^b)$. If $M_b>0$, the edges inside $V^b$ contribute $M_b\Phi_\lambda(\mu^b;V_b)\le M_b\tau_\lambda(V_b)$ by Lemma~\ref{lem:fK}(i), where $\mu^b(r_b(v))=\mu_v/M_b$; if $M_b=0$ they contribute $0$. A matching edge $\{u,w\}$ contributes $f_K(\mu_u,\mu_w)\le c_K(\mu_u+\mu_w)$ by Lemma~\ref{lem:fK}(i) if $K\ge2$ and by Lemma~\ref{lem:fK}(iii) if $K<2$, and matching edges are disjoint. Hence $\Phi_\lambda(\mu;V)\le\sum_{b:M_b>0}M_b\tau_\lambda(V_b)+c_K\le\max_{b:V_b\ne\emptyset}\tau_\lambda(V_b)+c_K$.
\end{proof}

\begin{lemma}[Bayes risk with an anchor]\label{lem:anchor}
Let $n\ge1$, $x_0\in\X$ and $b\in\{0,1\}$ with $\Hc^b_{x_0}\ne\emptyset$, let $U'\subseteq\X\setminus\{x_0\}$ be finite with $|U'|\le n$, and $p=1/(n+1)$. Then
\[
M_n(\Hc,\lambda)\ \ge\ p(1-p)^n\,\tau_\lambda(\Hc^b_{x_0}|_{U'})\ \ge\ \frac{\tau_\lambda(\Hc^b_{x_0}|_{U'})}{e(n+1)}.
\]
\end{lemma}

\begin{lemma}[Uniform sampling]\label{lem:uniform}
Let $n\ge1$ and let $U\subseteq\X$ be finite and nonempty. Then $M_n(\Hc,\lambda)\ge\frac1{|U|}\big(1-\frac1{|U|}\big)^n\tau_\lambda(\Hc|_U)$.
\end{lemma}

\begin{proof}[Proof of Lemmas~\ref{lem:anchor} and~\ref{lem:uniform}]
For Lemma~\ref{lem:anchor} let $\Hc'=\Hc^b_{x_0}$ and let $P$ put mass $p$ on each point of $U'$ and $1-p|U'|$ on $x_0$. For Lemma~\ref{lem:uniform} let $U'=U$, $p=1/|U|$, $\Hc'=\Hc$ and $P$ uniform on $U$. Let $V'=\Hc'|_{U'}$; if $E(V')=\emptyset$ there is nothing to prove. Otherwise let $\mu\in\Pc(V')$ attain $\tau_\lambda(V')$ and choose $h_v\in\Hc'$ with $h_v|_{U'}=v$ for $v\in V'$. Fix a learner $g$. As $\sup_{P',h}\Risk^\lambda_n(g;P',h)\ge\sum_v\mu_v\Risk^\lambda_n(g;P,h_v)$, it suffices to bound the right side from below.

Let $Z=(x_1,\dots,x_n)\sim P^n$ be independent of $v\sim\mu$. In the case of Lemma~\ref{lem:uniform} every sample point lies in $U'$ and carries the label $v(x_i)$; in the case of Lemma~\ref{lem:anchor} a sample point outside $U'$ is $x_0$ and carries the label $b$. In both cases the sample $S$ is a function of $(Z,v)$. Dropping the term of $x_0$ in the second case, $\sum_v\mu_v\Risk^\lambda_n(g;P,h_v)\ge p\sum_{x\in U'}\E\,\ell_\lambda(g_S(x),v(x))$. Fix $x\in U'$, and let $N_x$ be the event that $x$ is not a sample point, of probability $(1-p)^n$. On $N_x$ the sample is a function of $Z$ and $v|_{U'\setminus\{x\}}$, so the action at $x$ is $\nu_x(Z,v|_{U'\setminus\{x\}})$ for a function $\nu_x$. For $\zeta\in N_x$, independence gives $\E[\ell_\lambda(g_S(x),v(x))\mid Z=\zeta]=\sum_t\sum_{v\in V':\,v|_{U'\setminus\{x\}}=t}\mu_v\,\ell_\lambda(\nu_x(\zeta,t),v(x))$. The inner sum runs over at most two patterns, which differ only at $x$. If there are two, they form an edge $\{u,w\}$ of $G(V')$ in direction $x$ and the inner sum is at least $f_K(\mu_u,\mu_w)$ by Lemma~\ref{lem:fK}(v); every edge in direction $x$ arises once. Hence $\E\,\ell_\lambda(g_S(x),v(x))\ge(1-p)^n\sum f_K(\mu_u,\mu_w)$ over the edges in direction $x$, and summing over $x\in U'$ gives the bound $p(1-p)^n\Phi_\lambda(\mu;V')=p(1-p)^n\tau_\lambda(V')$. In Lemma~\ref{lem:anchor}, $p(1-p)^n=\frac1{n+1}(1+\frac1n)^{-n}\ge\frac1{e(n+1)}$.
\end{proof}

\begin{proposition}[Transfer]\label{prop:transfer}
Let $n\ge1$, $\lambda\in[0,\infty)$, $K=1+\lambda$, $c_K=\min\{1,K/2\}$ and $\varphi_n=\frac1{n+1}\big(\frac n{n+1}\big)^n$.
\begin{enumerate}
\item[(i)] For every nonempty $\Hc$, $M_n(\Hc,\lambda)\le F_\lambda(\Hc,n+1)/(n+1)\le1$ and $M_n(\Hc,\infty)\le F_\infty(\Hc,n+1)/(n+1)$.
\item[(ii)] If $\Hc$ is nontrivial, then $M_n(\Hc,\infty)\ge F_\infty(\Hc,n+1)/(2e(n+1))$ and, with $F=F_\lambda(\Hc,n+1)$,
\[
M_n(\Hc,\lambda)\ \ge\ \varphi_n\max\{F-c_K,\ c_K\}\ \ge\ \Big(\frac n{n+1}\Big)^n\frac{F}{2(n+1)}\ \ge\ \frac{F}{2e(n+1)};
\]
in particular $M_n(\Hc,\lambda)\ge\varphi_nc_K>0$.
\end{enumerate}
\end{proposition}

\begin{proof}
(i) Let $h^\star\in\Hc$, let $z_1,\dots,z_{n+1}$ be i.i.d.\ from $P$, and $c_j(z)=\ell_\lambda(g^\lambda_{S^{-j}}(z_j),h^\star(z_j))$. As the test point is independent of the sample, $\Risk^\lambda_n(g^\lambda;P,h^\star)=\E\,c_{n+1}(z)$. Since $g^\lambda$ depends on the sample only through the set of labeled pairs, $c_j(z)=c_{n+1}(z')$ for $z'$ equal to $z$ with the entries $j$ and $n+1$ exchanged, and $P^{n+1}$ is invariant under this exchange. With Lemma~\ref{lem:A-loo},
\[
\Risk^\lambda_n(g^\lambda;P,h^\star)=\frac1{n+1}\E\sum_{j=1}^{n+1}c_j(z)\le\frac1{n+1}\E\,\tau_\lambda\big(\Hc|_{\{z_1,\dots,z_{n+1}\}}\big)\le\frac{F_\lambda(\Hc,n+1)}{n+1},
\]
which is at most $1$ by Lemma~\ref{lem:basic}(iii). As $g^\infty$ never errs, $R_\infty(g^\infty_S;P,h^\star)$ is the probability of abstention, and the same argument with $c_j(z)$ the indicator that $g^\infty$ abstains on $(S^{-j},z_j)$ gives $\Risk^\infty_n(g^\infty;P,h^\star)\le F_\infty(\Hc,n+1)/(n+1)$ by Lemmas~\ref{lem:A-loo} and~\ref{lem:basic}(iv).

(ii) Take $h,h',x,x'$ from the definition of nontriviality, $x_0=x$ and $b=h(x)$. Then $x'\ne x_0$, and $\Hc^b_{x_0}|_{\{x'\}}=\{0,1\}^{\{x'\}}$ is a single edge of value $\max_{a\in[0,1]}\min\{1,Ka,K(1-a)\}=c_K$, so Lemma~\ref{lem:anchor} with $U'=\{x'\}$ gives $M_n(\Hc,\lambda)\ge\varphi_nc_K$. Next, Lemma~\ref{lem:basic}(iii) gives $U$ with $|U|\le n+1$ and $\tau_\lambda(\Hc|_U)=F$. If $U=\emptyset$, then $F=0$. Otherwise pick $x_0\in U$; Lemma~\ref{lem:A-split} gives $b$ with $\Hc^b_{x_0}|_{U\setminus\{x_0\}}\ne\emptyset$ and value at least $F-c_K$, and Lemma~\ref{lem:anchor} with $U'=U\setminus\{x_0\}$ gives $M_n(\Hc,\lambda)\ge\varphi_n(F-c_K)$. The second inequality follows from $\max\{F-c,c\}\ge F/2$ for $c>0$, and the third from $(1+1/n)^{-n}\ge1/e$. Finally $M_n(\Hc,\infty)\ge M_n(\Hc,\lambda)\ge F_\lambda(\Hc,n+1)/(2e(n+1))$ for every finite $\lambda$, and letting $\lambda\to\infty$ gives $M_n(\Hc,\infty)\ge F_\infty(\Hc,n+1)/(2e(n+1))$, which is the first claim.
\end{proof}

\subsection{Learners that never abstain}\label{sec:A-nonabst}

\begin{lemma}[Forbidding abstention]\label{lem:nonabst}
Let $\Hc$ be nonempty, $n\ge1$, $\lambda\in[0,\infty)$ and $K=1+\lambda$.
\begin{enumerate}
\item[(i)] $M^{\mathrm{na}}_n(\Hc,\lambda)=K\,M_n(\Hc,0)$.
\item[(ii)] $M_n(\Hc,0)\le M_n(\Hc,\lambda)\le M^{\mathrm{na}}_n(\Hc,\lambda)$, and $M_n(\Hc,\lambda)=M^{\mathrm{na}}_n(\Hc,\lambda)$ if $\lambda\le1$.
\item[(iii)] $\lambda\mapsto M_n(\Hc,\lambda)$ is nondecreasing and $\lambda\mapsto M_n(\Hc,\lambda)/(1+\lambda)$ is nonincreasing on $[0,\infty)$.
\item[(iv)] For finite nonempty $V\subseteq\{0,1\}^U$, the smallest value of $\max_vL_q(v)$ over the rules $q$ with $q_e(\abst)=0$ for all $e$ is $K\tau_0(V)$.
\end{enumerate}
\end{lemma}

\begin{proof}
Let $\nu\in\Pc(\{0,1,\abst\})$ and $y\in\{0,1\}$. (a) $\ell_\lambda(\nu,y)$ is nondecreasing in $\lambda$, and $\ell_\lambda(\nu,y)/K=\nu(\abst)/K+\nu(1-y)$ is nonincreasing. (b) If $\nu(\abst)=0$, then $\ell_\lambda(\nu,y)=K\ell_0(\nu,y)$. (c) Moving the abstention mass to the two labels in equal parts gives $\tilde\nu=(\nu(0)+\frac12\nu(\abst),\nu(1)+\frac12\nu(\abst),0)$ with $\ell_\lambda(\tilde\nu,y)=\ell_\lambda(\nu,y)+(K/2-1)\nu(\abst)$, which is at most $\ell_\lambda(\nu,y)$ when $\lambda\le1$. Every risk $\Risk^\lambda_n(g;P,h)$ is a sum of losses with nonnegative weights that do not depend on $g$ or $\lambda$, so an inequality between losses for all $\nu$ and $y$ carries over to risks, to suprema over $(P,h)$ and to infima over a common set of learners. For a learner $g$, let $\tilde g$ apply $\nu\mapsto\tilde\nu$ to every output; it never abstains.

(i) By (b), $M^{\mathrm{na}}_n(\Hc,\lambda)=KM^{\mathrm{na}}_n(\Hc,0)$. Restricting the learners gives $M^{\mathrm{na}}_n(\Hc,0)\ge M_n(\Hc,0)$, and (c) at $\lambda=0$ gives $\Risk^0_n(\tilde g;P,h)\le\Risk^0_n(g;P,h)$, hence $M^{\mathrm{na}}_n(\Hc,0)\le M_n(\Hc,0)$. (ii) The first inequality follows from (a), the second from restricting the learners, and for $\lambda\le1$, (c) gives $\Risk^\lambda_n(\tilde g;P,h)\le\Risk^\lambda_n(g;P,h)$, hence $M^{\mathrm{na}}_n(\Hc,\lambda)\le M_n(\Hc,\lambda)$. (iii) follows from (a). (iv) By (b), for rules without abstention every cost $c_v(q_e)$ at $\lambda$ is $K$ times the cost at $0$. At $\lambda=0$ the smallest value is at least $\tau_0(V)$ by Lemma~\ref{lem:lp}(ii), as fewer rules are allowed, and at most $\tau_0(V)$, since replacing each $q_e$ of an optimal rule by $\tilde q_e$ increases no cost by (c).
\end{proof}

\section{Proof of Theorem~\ref{thm:binary} and its constants}\label{sec:B}\label{app:criterion}

This section gives the lemmas behind the proof of Theorem~\ref{thm:binary} in Section~\ref{sec:proof}, shows that its constants are close to best possible, and draws two consequences for a fixed class. Throughout, $U$ is a finite set and $V\subseteq\{0,1\}^U$ is nonempty; the notation is that of Section~\ref{sec:A}.

\restatemain{thm:binary}

\subsection{Part (a)}\label{sec:B-a}

\begin{lemma}[Small penalties]\label{lem:smalllambda}
Let $0\le\lambda\le1$. Then $\tau_\lambda(V)=K\tau_0(V)$ and
\[
\tau_0(V)=\max_{\emptyset\ne B\subseteq V}\frac{|E_V(B)|}{|B|}\ \le\ d_V.
\]
For every finite set $T$, $\tau_0(\{0,1\}^T)\ge|T|/2$.
\end{lemma}

\begin{proof}
By Lemma~\ref{lem:fK}(iii), $\Phi_\lambda(\mu;V)=K\Phi_0(\mu;V)$ for every $\mu$, so $\tau_\lambda(V)=K\tau_0(V)$. Let $\rho^\ast$ be the maximum in the display. The uniform prior on $B$ gives each edge of $E_V(B)$ the cost $f_1(1/|B|,1/|B|)=1/|B|$, so $\tau_0(V)\ge\rho^\ast$. Conversely, $f_1(a,b)=\min\{a,b\}=\int_0^\infty\one[a>u]\,\one[b>u]\,du$, so with $L_u=\{v:\mu_v>u\}$,
\[
\Phi_0(\mu;V)=\int_0^\infty|E_V(L_u)|\,du\ \le\ \rho^\ast\int_0^\infty|L_u|\,du=\rho^\ast\sum_v\mu_v=\rho^\ast,
\]
where $|E_V(L_u)|\le\rho^\ast|L_u|$ holds trivially if $L_u=\emptyset$. The one-inclusion graph of a set $B$ of patterns has at most $d_B|B|$ edges \citep{HLW94,Hau95}, and $d_B\le d_V$, so $\rho^\ast\le d_V$. The cube $\{0,1\}^T$ has $|T|2^{|T|-1}$ edges and $2^{|T|}$ vertices, which gives the last claim.
\end{proof}

\noindent Equality $\tau_0(\{0,1\}^T)=|T|/2$ holds by Lemma~\ref{lem:cube}, so $\tau_\lambda(\{0,1\}^U)=K|U|/2$ for $\lambda\le1$.

\begin{proof}[Proof of Theorem~\ref{thm:binary}(a)]
Let $0\le\lambda\le1$. For $|U|\le m$, Lemma~\ref{lem:smalllambda} gives $\tau_\lambda(\Hc|_U)=K\tau_0(\Hc|_U)\le Kd_{\Hc|_U}\le Kd_m$. A shattered set $U$ with $|U|=d_m\le m$ has $\Hc|_U=\{0,1\}^U$ and $\tau_\lambda(\Hc|_U)\ge Kd_m/2$. Hence $(1+\lambda)d_m/2\le F_\lambda(\Hc,m)\le(1+\lambda)d_m$. The identity $M_n(\Hc,\lambda)=(1+\lambda)M_n(\Hc,0)$ is Lemma~\ref{lem:nonabst}(i),(ii).
\end{proof}

\subsection{Part (b): densities, cores and panel systems}\label{sec:B-b}

\begin{figure}[ht]
\centering
\begingroup
\input{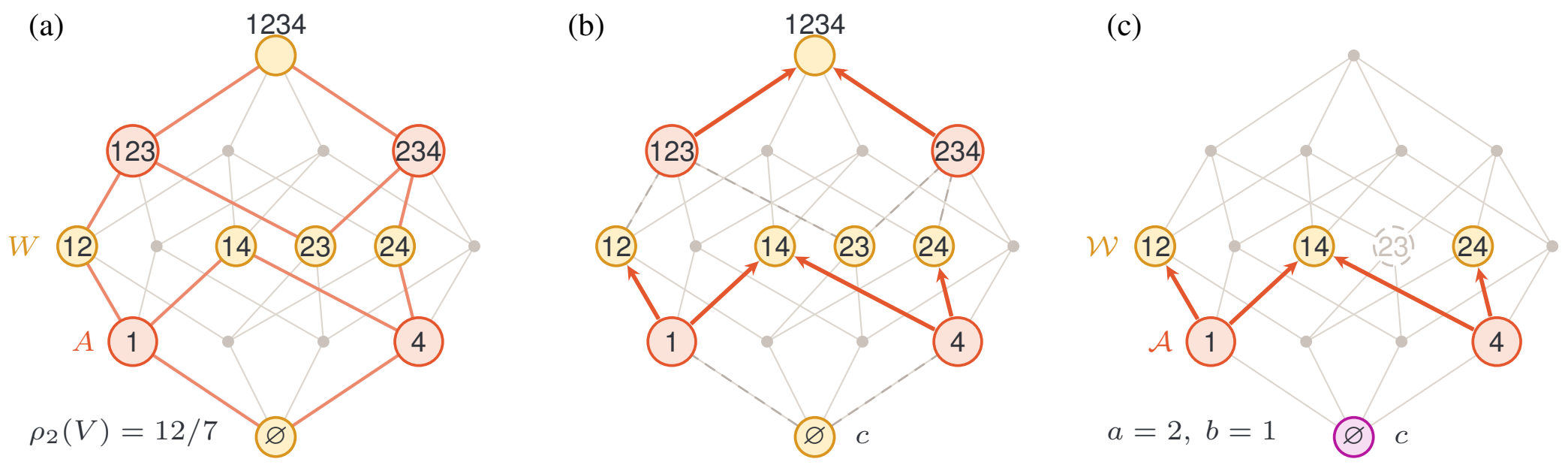}
\tikzset{
  pscube/.pic={
    \coordinate (e)    at (0,0);
    \coordinate (s1)   at (-2.25,1);  \coordinate (s2) at (-0.75,1);
    \coordinate (s3)   at (0.75,1);   \coordinate (s4) at (2.25,1);
    \coordinate (s12)  at (-3.125,2); \coordinate (s13) at (-1.875,2); \coordinate (s14) at (-0.625,2);
    \coordinate (s23)  at (0.625,2);  \coordinate (s24) at (1.875,2);  \coordinate (s34) at (3.125,2);
    \coordinate (s123) at (-2.25,3);  \coordinate (s124) at (-0.75,3);
    \coordinate (s134) at (0.75,3);   \coordinate (s234) at (2.25,3);
    \coordinate (s1234) at (0,4);
    \foreach \a/\b in {e/s1,e/s2,e/s3,e/s4,
      s1/s12,s1/s13,s1/s14,s2/s12,s2/s23,s2/s24,s3/s13,s3/s23,s3/s34,s4/s14,s4/s24,s4/s34,
      s12/s123,s12/s124,s13/s123,s13/s134,s14/s124,s14/s134,s23/s123,s23/s234,s24/s124,s24/s234,
      s34/s134,s34/s234,s123/s1234,s124/s1234,s134/s1234,s234/s1234}
      \draw[psgrid] (\a) -- (\b);
    \foreach \v in {s2,s3,s13,s34,s124,s134} \node[psdot] at (\v) {};
  }}
\begin{tikzpicture}[x=0.56cm,y=0.84cm,font=\sffamily\fontsize{7}{8}\selectfont,
  A/.style={psA,minimum size=12pt},
  W/.style={psW,minimum size=9.8pt},
  off/.style={psoff,minimum size=9.8pt},
  pair/.style={psedge},
  up/.style={psarrow},
  down/.style={psdown},
  tag/.style={pstag}]
\begin{scope}
\pic {pscube};
\foreach \a/\w in {s1/e,s1/s12,s1/s14,s4/e,s4/s14,s4/s24,s123/s12,s123/s23,s123/s1234,s234/s23,s234/s24,s234/s1234}
  \draw[pair] (\a) -- (\w);
\foreach \v/\t in {s1/1,s4/4,s123/123,s234/234} \node[A] at (\v) {\t};
\foreach \v/\t in {e/$\varnothing$,s12/12,s14/14,s23/23,s24/24,s1234/} \node[W] at (\v) {\t};
\node[psInk,above=1.5pt] at (s1234) {1234};
\node[tag] at (-3.9,4.45) {(a)};
\node[psLowerS,anchor=east] at (-2.6,1) {$A$};
\node[psnote,anchor=west] at (-3.9,0.05) {$\rho_2(V)=12/7$};
\node[psUpperS,anchor=east] at (-3.45,2) {$W$};
\end{scope}
\begin{scope}[xshift=4.75cm]
\pic {pscube};
\foreach \a/\w in {s1/e,s4/e,s123/s12,s123/s23,s234/s23,s234/s24} \draw[down] (\a) -- (\w);
\foreach \v/\t in {s1/1,s4/4,s123/123,s234/234} \node[A] (b\v) at (\v) {\t};
\foreach \v/\t in {e/$\varnothing$,s12/12,s14/14,s23/23,s24/24,s1234/} \node[W] (b\v) at (\v) {\t};
\node[psInk,above=1.5pt] at (s1234) {1234};
\foreach \a/\w in {s1/s12,s1/s14,s4/s14,s4/s24,s123/s1234,s234/s1234} \draw[up] (b\a) -- (b\w);
\node[psInk,right=1pt] at (be.east) {$c$};
\node[tag] at (-3.9,4.45) {(b)};
\end{scope}
\begin{scope}[xshift=9.5cm]
\pic {pscube};
\foreach \v in {s123,s234,s1234} \node[psdot] at (\v) {};
\node[off] at (s23) {23};
\foreach \v/\t in {s1/1,s4/4} \node[A] (c\v) at (\v) {\t};
\foreach \v/\t in {s12/12,s14/14,s24/24} \node[W] (c\v) at (\v) {\t};
\node[psC,minimum size=9.8pt] (ce) at (e) {$\varnothing$};
\foreach \a/\w in {s1/s12,s1/s14,s4/s14,s4/s24} \draw[up] (c\a) -- (c\w);
\node[psInk,right=1pt] at (ce.east) {$c$};
\node[psnote,anchor=west] at (-3.9,0.05) {$a=2,\ b=1$};
\node[tag] at (-3.9,4.45) {(c)};
\node[psLowerS,anchor=east] at (-2.6,1) {$\mathcal A$};
\node[psUpperS,anchor=east] at (-3.45,2) {$\mathcal W$};
\end{scope}
\end{tikzpicture}%
\endgroup
\caption{The proof of Theorem~\ref{thm:binary}(b) on $V=\{\varnothing,1,4,12,14,23,24,123,234,1234\}\subseteq\{0,1\}^4$, patterns written as supports, with $\lambda=2$; small dots mark the other patterns. (a) A pair $(A,W)$ attaining $\rho_2(V)$, with vertex sizes for the two-level prior of Step~1; it is already a two-sided core, with $\delta_A=3$ and $\delta_W=2$. (b) Step~3 with center $c=0$: six of the twelve edges point away from $c$, four into layer $2$ and two into layer $4$. (c) Peeling layer $2$ removes $23$ and leaves a panel system of value $\min\{a,\lambda b\}=2\ge\rho_2(V)/2$.}
\label{fig:B-steps}
\end{figure}

Figure~\ref{fig:B-steps} follows the three steps of Section~\ref{sec:proof} on ten patterns in $\{0,1\}^4$ with $\lambda=2$; the lemmas below give each step in general.

\begin{lemma}[Peeling]\label{lem:peel}
Let $P$ and $Q$ be disjoint finite sets, $E$ a set of pairs $\{p,q\}$ with $p\in P$ and $q\in Q$, and $\lambda,r>0$ with $|E|>0$ and $|E|\ge r(|P|+|Q|/\lambda)$. Then there are nonempty $P'\subseteq P$ and $Q'\subseteq Q$ such that every $p\in P'$ lies in at least $r$ pairs of $E$ whose other end is in $Q'$, and every $q\in Q'$ lies in at least $r/\lambda$ pairs of $E$ whose other end is in $P'$.
\end{lemma}

\begin{proof}
We run the peeling algorithm of \citet{Cha00} with weight $1$ on $P$ and $1/\lambda$ on $Q$. Repeatedly delete a vertex of $P$ that lies in fewer than $r$ pairs of $E$ joining it to undeleted vertices, or a vertex of $Q$ that lies in fewer than $r/\lambda$ such pairs, until no such vertex remains, and let $P',Q'$ be the undeleted vertices. If $P'\ne\emptyset=Q'$, every vertex of $P'$ lies in $0<r$ such pairs, so the process would not have stopped; likewise $P'=\emptyset\ne Q'$ is impossible. If $P'=Q'=\emptyset$, every pair of $E$ is removed when its first end is deleted, and each deletion removes fewer than $r$ pairs (a vertex of $P$) or fewer than $r/\lambda$ pairs (a vertex of $Q$). As $|E|>0$, at least one vertex is deleted, and $|E|<r|P|+(r/\lambda)|Q|\le|E|$, a contradiction. So $P'$ and $Q'$ are nonempty and have the stated degrees.
\end{proof}

\begin{lemma}[Parity layer cake]\label{lem:layer}
If $1\le\lambda<\infty$, then $\tau_\lambda(V)\le2\rho_\lambda(V)$.
\end{lemma}

\begin{proof}
Fix $\mu\in\Pc(V)$. For $u>0$ let $A_u=\{v:\mu_v\ge u\}$ and $B_u=\{v:\mu_v\ge u/\lambda\}$, and let $e(A_u,B_u)$ be the number of ordered pairs $(p,q)\in A_u\times B_u$ of adjacent patterns. For an edge $\{v,w\}$ with $a=\mu_v\ge b=\mu_w$, the pair $(v,w)$ is counted when $u\le\min\{a,\lambda b\}$ and the pair $(w,v)$ when $u\le\min\{b,\lambda a\}=b$, since $\lambda a\ge a\ge b$. So the edge contributes $\min\{a,\lambda b\}+b=f_K(a,b)$ to $\int_0^\infty e(A_u,B_u)\,du$ by Lemma~\ref{lem:fK}(iv), and $\Phi_\lambda(\mu;V)=\int_0^\infty e(A_u,B_u)\,du$.

Let $V^{\mathrm{ev}}$ and $V^{\mathrm{odd}}$ be the patterns of $V$ with an even and an odd number of ones. Adjacent patterns lie in different classes, so the pairs counted by $e(A_u,B_u)$ correspond one to one to the edges of $E_V(A_u\cap V^{\mathrm{ev}},B_u\cap V^{\mathrm{odd}})$ and of $E_V(A_u\cap V^{\mathrm{odd}},B_u\cap V^{\mathrm{ev}})$. In both cases the two sets are disjoint, so the definition of $\rho=\rho_\lambda(V)$, which holds trivially when both sets are empty, gives
\[
e(A_u,B_u)\le\rho\Big(|A_u\cap V^{\mathrm{ev}}|+\frac{|B_u\cap V^{\mathrm{odd}}|}\lambda\Big)+\rho\Big(|A_u\cap V^{\mathrm{odd}}|+\frac{|B_u\cap V^{\mathrm{ev}}|}\lambda\Big)=\rho\Big(|A_u|+\frac{|B_u|}\lambda\Big).
\]
Since $\int_0^\infty|A_u|\,du=\sum_v\mu_v=1$ and $\int_0^\infty|B_u|\,du=\sum_v\lambda\mu_v=\lambda$, integration gives $\Phi_\lambda(\mu;V)\le2\rho$.
\end{proof}

\begin{lemma}[Two-level priors]\label{lem:twolevel}
Let $1\le\lambda<\infty$, and let $A,W\subseteq V$ be disjoint and not both empty. The prior with mass $\theta=1/(|A|+|W|/\lambda)$ on each vertex of $A$ and $\theta/\lambda$ on each vertex of $W$ has $\Phi_\lambda\ge\frac K\lambda\,\theta\,|E_V(A,W)|$. Hence $\tau_\lambda(V)\ge\frac K\lambda\rho_\lambda(V)$.
\end{lemma}

\begin{proof}
The masses sum to $\theta|A|+\theta|W|/\lambda=1$. Since $\theta\ge\theta/\lambda$, Lemma~\ref{lem:fK}(iv) gives each edge between $A$ and $W$ the cost $f_K(\theta,\theta/\lambda)=\theta/\lambda+\theta=K\theta/\lambda$, and the other edges cost at least $0$. Maximizing over $(A,W)$ gives the second claim.
\end{proof}

\noindent By Lemmas~\ref{lem:layer} and~\ref{lem:twolevel}, the best two-level prior is optimal up to the factor $2\lambda/K<2$.

\begin{lemma}[Cores and densities]\label{lem:corerho}
If $0<\lambda<\infty$, then $\frac12\kappa_\lambda(V)\le\rho_\lambda(V)\le\kappa_\lambda(V)$.
\end{lemma}

\begin{proof}
If $E(V)=\emptyset$, all three quantities are $0$. Otherwise let $(A,W)$ be a two-sided core attaining $\kappa=\kappa_\lambda(V)$. Counting $E_V(A,W)$ from each side, $|E_V(A,W)|\ge\delta_A|A|\ge\kappa|A|$ and $|E_V(A,W)|\ge\delta_W|W|\ge\kappa|W|/\lambda$, so $|E_V(A,W)|\ge\kappa\max\{|A|,|W|/\lambda\}\ge\frac\kappa2(|A|+|W|/\lambda)$ and $\rho_\lambda(V)\ge\kappa/2$. Conversely let $(A,W)$ attain $\rho=\rho_\lambda(V)$, which is positive because an edge $\{u,w\}$ gives the pair $(\{u\},\{w\})$ a positive ratio. Lemma~\ref{lem:peel} with $P=A$, $Q=W$, $E=E_V(A,W)$ and $r=\rho$ gives nonempty $A'\subseteq A$ and $W'\subseteq W$ in which every vertex of $A'$ has at least $\rho$ neighbors in $W'$ and every vertex of $W'$ at least $\rho/\lambda$ neighbors in $A'$. These numbers are positive integers, so $(A',W')$ is a two-sided core with $\delta_{A'}\ge\rho$ and $\lambda\delta_{W'}\ge\rho$, and $\kappa_\lambda(V)\ge\rho$.
\end{proof}

\begin{lemma}[Panel systems and cores]\label{lem:panelcore}
Let $\lambda\in[0,\infty)$.
\begin{enumerate}
\item[(i)] If $V$ shatters a panel system $(\Ac,\Wc)$ with center $c$, then $(A,W)=(\{c\oplus\one_S:S\in\Ac\},\{c\oplus\one_T:T\in\Wc\})$ is a two-sided core of $G(V)$ with $\delta_A\ge a(\Ac,\Wc)$ and $\delta_W\ge b(\Ac,\Wc)$. Hence $\pi_\lambda(V)\le\kappa_\lambda(V)\le\Delta(V)$, and $\pi_\lambda(\Hc,m)\le\kappa_\lambda(\Hc,m)\le\min\{s,m\}$ for every $m\ge0$.
\item[(ii)] Let $\lambda\ge1$. If $\Hc$ shatters a set $T$ with $|T|=t\ge1$, then $\pi_\lambda(\Hc,t)\ge t/2$. Consequently $d_m\le2\pi_\lambda(\Hc,m)$ for every $m\ge0$.
\item[(iii)] Let $\lambda\ge1$. Then $\kappa_\lambda(\Hc,m)\ge d_m$ for every $m\ge0$.
\end{enumerate}
\end{lemma}

\begin{proof}
(i) If the rank is $k$, the patterns of $A$ lie at Hamming distance $k-1$ from $c$ and those of $W$ at distance $k$, so $A\cap W=\emptyset$; both sets are nonempty, and $S\mapsto c\oplus\one_S$ is injective. If $S\subset T$ and $|T|=|S|+1$, then $c\oplus\one_S$ and $c\oplus\one_T$ differ exactly at the point of $T\setminus S$, so they are adjacent. Hence every vertex of $A$ has at least $a(\Ac,\Wc)\ge1$ neighbors in $W$ and every vertex of $W$ at least $b(\Ac,\Wc)\ge1$ neighbors in $A$, and $\kappa_\lambda(V)\ge\min\{a(\Ac,\Wc),\lambda b(\Ac,\Wc)\}$; so $\pi_\lambda(V)\le\kappa_\lambda(V)$. If $E(V)\ne\emptyset$, every two-sided core and every $a\in A$ satisfy $\min\{\delta_A,\lambda\delta_W\}\le\delta_A\le\deg_V(a)\le\Delta(V)$, and if $E(V)=\emptyset$, then $\kappa_\lambda(V)=0$. For the class, $\Delta(\Hc|_U)\le\min\{s,|U|\}$.

(ii) Here $\Hc|_T=\{0,1\}^T$. Let $k=\lceil t/2\rceil$, let $\Ac$ be all $(k-1)$-subsets and $\Wc$ all $k$-subsets of $T$. A $(k-1)$-set lies in $t-k+1$ members of $\Wc$ and a $k$-set contains $k$ members of $\Ac$, so $(\Ac,\Wc)$ is a panel system on $T$, shattered with center $0$, of value $\min\{t-k+1,\lambda k\}\ge\min\{\lfloor t/2\rfloor+1,\lceil t/2\rceil\}\ge t/2$. For the consequence, if $d_m\ge1$ apply this to a shattered set with $d_m\le m$ points.

(iii) If $d_m\ge1$, let $U$ be shattered with $|U|=d_m$, and let $A$ and $W$ be the patterns of $\Hc|_U=\{0,1\}^U$ with an even and with an odd number of ones. Every neighbor of a pattern lies in the other class, so $(A,W)$ is a two-sided core with $\delta_A=\delta_W=d_m$, and $\kappa_\lambda(\Hc,m)\ge\min\{d_m,\lambda d_m\}=d_m$.
\end{proof}

\begin{lemma}[Dense pairs give panel systems]\label{lem:rhopi}
If $0<\lambda<\infty$, then $\rho_\lambda(V)\le2\pi_\lambda(V)$.
\end{lemma}

\begin{proof}
Let $\rho=\rho_\lambda(V)$; if $\rho=0$ there is nothing to prove. Otherwise let disjoint $A,W$ attain $\rho$, so that $|E_V(A,W)|=\rho(|A|+|W|/\lambda)>0$.

\emph{Orientation.} For $c\in\{0,1\}^U$ let $E_c$ be the set of edges $\{a,w\}\in E_V(A,W)$, $a\in A$, $w\in W$, with $|w\oplus c|=|a\oplus c|+1$. Adjacent patterns satisfy $|w\oplus c|=|a\oplus c|\pm1$, and $|v\oplus\one_U|=|U|-|v|$, so every edge of $E_V(A,W)$ lies in exactly one of $E_0$ and $E_{\one_U}$. Fix $c\in\{0,\one_U\}$ with $|E_c|\ge|E_V(A,W)|/2=\frac\rho2(|A|+|W|/\lambda)$.

\emph{Levels.} For $1\le j\le|U|$ let $A_{j-1}=\{a\in A:|a\oplus c|=j-1\}$, $W_j=\{w\in W:|w\oplus c|=j\}$, and let $E_j$ be the set of edges of $E_c$ joining $A_{j-1}$ and $W_j$. Then $E_c$ is the disjoint union of the sets $E_j$, every edge of $G(V)$ between $A_{j-1}$ and $W_j$ lies in $E_j$, and $\sum_j(|A_{j-1}|+|W_j|/\lambda)\le|A|+|W|/\lambda$. If every $j$ with $E_j\ne\emptyset$ had $|E_j|<\frac\rho2(|A_{j-1}|+|W_j|/\lambda)$, summing over these $j$, of which there is at least one, would give $|E_c|<\frac\rho2(|A|+|W|/\lambda)$. So some $j$ has $|E_j|>0$ and $|E_j|\ge\frac\rho2(|A_{j-1}|+|W_j|/\lambda)$.

\emph{Peeling.} Lemma~\ref{lem:peel} with $P=A_{j-1}$, $Q=W_j$, $E=E_j$ and $r=\rho/2$ gives nonempty $A'\subseteq A_{j-1}$ and $W'\subseteq W_j$ in which every vertex of $A'$ has at least $\rho/2$ neighbors in $W'$ and every vertex of $W'$ at least $\rho/(2\lambda)$ neighbors in $A'$. Let $\Ac=\{\supp(a\oplus c):a\in A'\}$ and $\Wc=\{\supp(w\oplus c):w\in W'\}$, families of $(j-1)$-sets and $j$-sets. For $a\in A_{j-1}$ and $w\in W_j$, $a$ and $w$ are adjacent if and only if $\supp(a\oplus c)\subset\supp(w\oplus c)$: if they differ only at $x$, then $w\oplus c$ is $a\oplus c$ with the coordinate $x$ raised from $0$ to $1$, and conversely a $(j-1)$-set and a $j$-set containing it differ in one point. As $v\mapsto\supp(v\oplus c)$ is injective, $(\Ac,\Wc)$ is a panel system of rank $j$ with $a(\Ac,\Wc)\ge\rho/2$ and $b(\Ac,\Wc)\ge\rho/(2\lambda)$, both positive integers, and $V$ shatters it with center $c$ because $c\oplus\one_{\supp(v\oplus c)}=v$. Its value is at least $\min\{\rho/2,\lambda\rho/(2\lambda)\}=\rho/2$.
\end{proof}

\begin{proposition}[The chain]\label{prop:chain}
Let $1\le\lambda<\infty$ and $K=1+\lambda$.
\begin{enumerate}
\item[(i)] For every finite nonempty $V\subseteq\{0,1\}^U$,
\[
\tfrac K{2\lambda}\pi_\lambda(V)\le\tfrac K{2\lambda}\kappa_\lambda(V)\le\tfrac K\lambda\rho_\lambda(V)\le\tau_\lambda(V)\le2\rho_\lambda(V)\le\min\{2\kappa_\lambda(V),4\pi_\lambda(V)\},
\]
and $\pi_\lambda(V)\le\kappa_\lambda(V)\le2\rho_\lambda(V)\le4\pi_\lambda(V)$.
\item[(ii)] For every nonempty $\Hc$ and $m\ge0$, $\frac K{2\lambda}\pi_\lambda(\Hc,m)\le F_\lambda(\Hc,m)\le4\pi_\lambda(\Hc,m)$, $\frac K{2\lambda}\kappa_\lambda(\Hc,m)\le F_\lambda(\Hc,m)\le2\kappa_\lambda(\Hc,m)$, and $\pi_\lambda(\Hc,m)\le\kappa_\lambda(\Hc,m)\le4\pi_\lambda(\Hc,m)$.
\item[(iii)] Let $n\ge1$, $\pi=\pi_\lambda(\Hc,n+1)$ and $\kappa=\kappa_\lambda(\Hc,n+1)$. Then $M_n(\Hc,\lambda)\le4\pi/(n+1)$ and $M_n(\Hc,\lambda)\le2\kappa/(n+1)$. If $\Hc$ is nontrivial, then $M_n(\Hc,\lambda)\ge\pi/(4e(n+1))$ and $M_n(\Hc,\lambda)\ge\kappa/(4e(n+1))$.
\end{enumerate}
\end{proposition}

\begin{proof}
(i) combines Lemmas~\ref{lem:panelcore}(i), \ref{lem:corerho}, \ref{lem:twolevel}, \ref{lem:layer} and~\ref{lem:rhopi}. (ii) For $|U|\le m$, (i) gives $\tau_\lambda(\Hc|_U)\le2\kappa_\lambda(\Hc|_U)\le2\kappa_\lambda(\Hc,m)$ and $\tau_\lambda(\Hc|_U)\le4\pi_\lambda(\Hc,m)$. A set $U$ with $|U|\le m$ attaining $\kappa_\lambda(\Hc,m)$ (Lemma~\ref{lem:basic}(iii)) gives $F_\lambda(\Hc,m)\ge\tau_\lambda(\Hc|_U)\ge\frac K{2\lambda}\kappa_\lambda(\Hc,m)$, and likewise for $\pi_\lambda$. The last chain is (i) maximized over $U$. (iii) Proposition~\ref{prop:transfer} and (ii) with $m=n+1$, using $K/(2\lambda)\ge\frac12$.
\end{proof}

\begin{proof}[Proof of Theorem~\ref{thm:binary}(b)]
This is Proposition~\ref{prop:chain} for $m\ge1$.
\end{proof}

\subsection{Part (c)}\label{sec:B-c}

\begin{lemma}[Errors forbidden]\label{lem:Finfty}
For every nonempty $\Hc$ and $m\ge0$, $F_\infty(\Hc,m)=\max_{|U|\le m}\Delta(\Hc|_U)=\min\{s,m\}$.
\end{lemma}

\begin{proof}
The first equality is Lemma~\ref{lem:basic}(iv), which passes the limit $\lambda\to\infty$ through the maximum over the finitely many projection types on at most $m$ points (Lemma~\ref{lem:basic}(iii)).
Each $\Delta(\Hc|_U)$ with $|U|\le m$ is at most $|U|\le m$ and at most $s$. Conversely let $D=\min\{s,m\}\ge1$. Some finite $U$ has $\Delta(\Hc|_U)\ge D$: if $s<\infty$ the supremum of integers defining $s$ is attained, and if $s=\infty$ some $U$ has $\Delta(\Hc|_U)\ge m$. Let $v\in\Hc|_U$ have degree at least $D$, let $x_1,\dots,x_D$ be the directions of $D$ of its edges, and $T=\{x_1,\dots,x_D\}$. The patterns $v|_T$ and $(v^{(x_i)})|_T=(v|_T)^{(x_i)}$ lie in $\Hc|_T$, so $v|_T$ has degree $D$ in $G(\Hc|_T)$, and $|T|=D\le m$.
\end{proof}

\begin{proof}[Proof of Theorem~\ref{thm:binary}(c)]
Proposition~\ref{prop:transfer} at $\lambda=\infty$ and Lemma~\ref{lem:Finfty} with $m=n+1$.
\end{proof}

\subsection{Near-optimal constants}\label{sec:B-tight}\label{app:constants}

For a finite set $T$, the star $S(T)=\{0\}\cup\{\one_{\{x\}}:x\in T\}\subseteq\{0,1\}^T$ has one-inclusion graph $K_{1,|T|}$ with center $0$. For $\lambda\ge1$ and $|T|=t\ge1$,
\begin{equation}\label{eq:B-star}
\tau_\lambda(S(T))=tK/(t+\lambda).
\end{equation}
Indeed, Lemma~\ref{lem:twolevel} with $A=\{0\}$ and $W$ the leaves gives $\tau_\lambda(S(T))\ge\frac K\lambda\cdot\frac t{1+t/\lambda}$. Conversely, if $\mu$ has mass $\mu_0$ at the center, the edge to the leaf $\one_{\{x\}}$ costs at most $\mu_0+\mu_{\one_{\{x\}}}$ and at most $K\mu_{\one_{\{x\}}}$, so $\Phi_\lambda(\mu;S(T))\le\min\{1+(t-1)\mu_0,K(1-\mu_0)\}$, and the two terms, nondecreasing and decreasing in $\mu_0$, meet at $\mu_0=\lambda/(t+\lambda)$ with the value $tK/(t+\lambda)$.

\begin{proposition}[The factors around $\kappa_\lambda$]\label{prop:B-coretight}
Let $\lambda\ge1$ be an integer and $K=1+\lambda$.
\begin{enumerate}
\item[(i)] If $|T|=\lambda$, then $\tau_\lambda(S(T))=K/2=\frac K{2\lambda}\kappa_\lambda(S(T))=\frac K{2\lambda}\pi_\lambda(S(T))$.
\item[(ii)] Let $T,S$ be disjoint finite sets with $|T|=\lambda-1$ and $|S|=s\ge1$, and $V=\{0,1\}^T\times S(S)$. Then $\kappa_\lambda(V)\le\lambda$ and $\tau_\lambda(V)\ge\lambda-1+sK/(s+\lambda)$, so $\tau_\lambda(V)/\kappa_\lambda(V)\ge(\lambda-1+sK/(s+\lambda))/\lambda$, which tends to $2$ as $s\to\infty$.
\end{enumerate}
Both hold also for the class $\Hc=V$ on the domain $U$ of $V$, with $F_\lambda(\Hc,|U|)$ and $\kappa_\lambda(\Hc,|U|)$ in place of $\tau_\lambda(V)$ and $\kappa_\lambda(V)$, so the factors $K/(2\lambda)$ and $2$ around $\kappa_\lambda$ in Proposition~\ref{prop:chain}(i),(ii) cannot be improved.
\end{proposition}

\begin{proof}
(i) By~\eqref{eq:B-star}, $\tau_\lambda(S(T))=K/2$. Each leaf has the center as its only neighbor, so every two-sided core $(A,W)$ contains the center. If the center lies in $A$, then $A$ is the center alone, $\delta_W=1$ and $\delta_A=|W|\le\lambda$; if it lies in $W$, then $\delta_A=1$. The core of the center and all leaves has value $\lambda$, so $\kappa_\lambda(S(T))=\lambda$. The panel system $(\{\emptyset\},\{\{x\}:x\in T\})$, shattered with center $0$, has $a=\lambda$ and $b=1$, so $\pi_\lambda(S(T))=\lambda$ by Lemma~\ref{lem:panelcore}(i). For the class, every projection is a star $S(U')$ with $|U'|\le\lambda$ or one pattern, of value at most $K/2$ by~\eqref{eq:B-star} and core number at most $\Delta(S(U'))\le\lambda$, so $F_\lambda(\Hc,\lambda)=K/2$ and $\kappa_\lambda(\Hc,\lambda)=\lambda$.

(ii) \emph{Value.} For $V_i\subseteq\{0,1\}^{U_i}$ with $U_1\cap U_2=\emptyset$, an edge of $G(V_1\times V_2)$ is an edge of $G(V_1)$ or of $G(V_2)$ in one component, with the other component fixed. By the homogeneity in Lemma~\ref{lem:fK}(i), a product prior therefore satisfies $\Phi_\lambda(\mu^1\otimes\mu^2;V_1\times V_2)=\Phi_\lambda(\mu^1;V_1)+\Phi_\lambda(\mu^2;V_2)$. The uniform prior on $\{0,1\}^T$ gives each of its $|T|2^{|T|-1}$ edges the cost $2^{1-|T|}$, as $K\ge2$, so with~\eqref{eq:B-star}, $\tau_\lambda(V)\ge|T|+sK/(s+\lambda)$.

\emph{Cores.} Let $V'=\{0,1\}^{T'}\times S(S')$ with $|T'|\le\lambda-1$ and $S'$ finite, which covers $V$ and its projections; we show $\kappa_\lambda(V')\le\lambda$. If $S'=\emptyset$, $\kappa_\lambda(V')\le\Delta(V')<\lambda$. Otherwise write the patterns as $(q,z)$ with $q\in Q=\{0,1\}^{T'}$. A pattern $(q,\one_{\{y\}})$ has the neighbors $(q,0)$ and $(q\oplus\one_{\{x\}},\one_{\{y\}})$, $x\in T'$, so it has degree at most $\lambda$ and one neighbor in $Q\times\{0\}$, and $(q,0)$ has $|T'|$ neighbors in $Q\times\{0\}$. A two-sided core with $\min\{\delta_A,\lambda\delta_W\}>\lambda$ has $\delta_A\ge\lambda+1$ and $\delta_W\ge2$. Then $A\subseteq Q\times\{0\}$ by the degrees, $W\subseteq Q\times\{0\}$ because each vertex of $W$ has two neighbors in $A$, and each vertex of $A$ has at most $|T'|<\delta_A$ neighbors in $W$, a contradiction. The ratio bound follows, and its right side tends to $(\lambda-1+K)/\lambda=2$. For the class, $\kappa_\lambda(\Hc,|U|)\le\lambda$ by the bound for projections, and $F_\lambda(\Hc,|U|)\ge\tau_\lambda(V)$.
\end{proof}

\begin{proposition}[The constants of the transfer]\label{prop:B-transfertight}
Let $n\ge1$, $c_K=\min\{1,K/2\}$ and $\varphi_n=\frac1{n+1}(\frac n{n+1})^n$.
\begin{enumerate}
\item[(i)] If $\X$ is countably infinite and $\Hc=\{0,1\}^{\X}$, then $F_\lambda(\Hc,n+1)=c_K(n+1)$ and $M_n(\Hc,\lambda)=c_K=F_\lambda(\Hc,n+1)/(n+1)$ for every $\lambda\in[0,\infty)$.
\item[(ii)] Let $\X=\{1,2,3\}$, $\Hc_3=\{\one_{\{1\}},\one_{\{2\}},\one_{\{3\}}\}$ and $\lambda\ge1$. Then $\Hc_3$ is nontrivial, $F_\lambda(\Hc_3,n+1)=2K/(K+1)$, and $\varphi_n\le M_n(\Hc_3,\lambda)\le\varphi_n+2^{-(n+1)}$.
\end{enumerate}
Consequently, for $\lambda\ge1$ the largest $c_n(K)$ with $M_n(\Hc,\lambda)\ge c_n(K)F_\lambda(\Hc,n+1)/(n+1)$ for all nontrivial $\Hc$ satisfies $\frac12(\frac n{n+1})^n\le c_n(K)\le\big((\frac n{n+1})^n+(n+1)2^{-(n+1)}\big)\frac{K+1}{2K}$, and both bounds tend to $\frac1{2e}$ as $n,K\to\infty$. So the constant $1$ of the upper bound in~\eqref{eq:transfer} is attained, and $2e$ is the limit of the best constants of the lower bound.
\end{proposition}

\begin{proof}
(i) Every projection on a finite $U$ is $\{0,1\}^U$, of value $c_K|U|$ by Lemma~\ref{lem:cube}, and $\X$ is infinite, so $F_\lambda(\Hc,n+1)=c_K(n+1)$ and $M_n(\Hc,\lambda)\le c_K$ by Proposition~\ref{prop:transfer}(i). Lemma~\ref{lem:uniform} with $|U|=u$ gives $M_n(\Hc,\lambda)\ge c_K(1-1/u)^n$ for every $u\ge1$; let $u\to\infty$.

(ii) $\one_{\{1\}}$ and $\one_{\{2\}}$ agree at $3$ and differ at $1$. A projection on one point is a single edge, of value $1$; on two points $\{i,j\}$ it is the star $S(\{i,j\})$, of value $2K/(K+1)\ge1$ by~\eqref{eq:B-star}; on three points its patterns are pairwise at distance $2$, of value $0$. So $F_\lambda(\Hc_3,n+1)=2K/(K+1)<2$, and Proposition~\ref{prop:transfer}(ii) with $c_K=1$ gives $M_n(\Hc_3,\lambda)\ge\varphi_n\max\{F_\lambda(\Hc_3,n+1)-1,1\}=\varphi_n$.
For the upper bound, consider the learner that outputs the target when a label $1$ occurs or the sample contains two distinct points (then the target is $\one_{\{c\}}$ for the third point $c$), and otherwise, when all sample points equal one point $j$ with label $0$, predicts $0$ at $j$ and abstains elsewhere. It never errs, and for the target $\one_{\{c\}}$ its risk is $\sum_{j\ne c}\psi(P(\{j\}))$ with $\psi(p)=p^n(1-p)$. As $\psi'(p)=p^{n-1}(n-(n+1)p)$, $\psi\le\psi(\frac n{n+1})=\varphi_n$, and $\psi(p)\le\psi(\frac12)=2^{-(n+1)}$ for $p\le\frac12$; one of the two masses $P(\{j\})$, $j\ne c$, is at most $\frac12$. Finally, the lower bound on $c_n(K)$ is Proposition~\ref{prop:transfer}(ii), and the upper bound is $M_n(\Hc_3,\lambda)(n+1)/F_\lambda(\Hc_3,n+1)$ bounded by (ii), with $\varphi_n(n+1)=(n/(n+1))^n$. Both tend to $\frac1{2e}$ as $(n/(n+1))^n\to e^{-1}$, $(n+1)2^{-(n+1)}\to0$ and $(K+1)/(2K)\to\frac12$.
\end{proof}

\subsection{A fixed class}\label{sec:B-fixed}

\begin{lemma}[Core numbers of a fixed class]\label{lem:kappaeventual}
Let $1\le\lambda<\infty$ and let $\Hc$ have finite VC dimension $d$. Then $m\mapsto\kappa_\lambda(\Hc,m)$ is nondecreasing, $\kappa_\lambda(\Hc,m)\le2\lambda d\le2Kd$, and every value of $\kappa_\lambda(\Hc,m)$ has the form $\min\{a,\lambda b\}$ with integers $a,b\ge0$. Hence there is $m_0$ with $\kappa_\lambda(\Hc,m)=\bar\kappa_\lambda(\Hc)$ for all $m\ge m_0$. Moreover $\bar\pi_\lambda(\Hc)\le\bar\kappa_\lambda(\Hc)\le4\bar\pi_\lambda(\Hc)$, and if $\Hc$ is nontrivial, then for every $n$ with $n+1\ge m_0$,
\[
\frac{\bar\pi_\lambda(\Hc)}{4e(n+1)}\ \le\ M_n(\Hc,\lambda)\ \le\ \frac{8\,\bar\pi_\lambda(\Hc)}{n+1}.
\]
\end{lemma}

\begin{proof}
Monotonicity is Lemma~\ref{lem:basic}(iii). Let $V=\Hc|_U$. For disjoint $A,W\subseteq V$, not both empty, the edge bound in the proof of Lemma~\ref{lem:smalllambda} gives $|E_V(A,W)|\le|E_V(A\cup W)|\le d_V(|A|+|W|)\le\lambda d_V(|A|+|W|/\lambda)$, as $\lambda\ge1$. So $\rho_\lambda(V)\le\lambda d_V\le\lambda d$, and $\kappa_\lambda(V)\le2\lambda d$ by Lemma~\ref{lem:corerho}. The value $\kappa_\lambda(V)$ is $0$ or $\min\{\delta_A,\lambda\delta_W\}$ with integers $\delta_A,\delta_W$. If such a value equals $a$, then $a\le2\lambda d$; if it equals $\lambda b$, then $b\le2d$. So the possible values lie in the finite set $\{0,1,\dots,\lfloor2\lambda d\rfloor\}\cup\{\lambda b:b\le2d\}$, and a nondecreasing sequence in a finite set equals its supremum from some index on. Taking suprema over $m$ in Proposition~\ref{prop:chain}(ii) gives $\bar\pi_\lambda\le\bar\kappa_\lambda\le4\bar\pi_\lambda$. For $n+1\ge m_0$, $\kappa_\lambda(\Hc,n+1)=\bar\kappa_\lambda(\Hc)$, and Proposition~\ref{prop:chain}(iii) gives $\bar\kappa_\lambda/(4e(n+1))\le M_n(\Hc,\lambda)\le2\bar\kappa_\lambda/(n+1)$, which with $\bar\pi_\lambda\le\bar\kappa_\lambda\le4\bar\pi_\lambda$ is the claim.
\end{proof}

\begin{corollary}[Panel systems decide the gain]\label{cor:B-criterion}
Let $1\le\lambda<\infty$, let $\Hc$ have VC dimension $d$ with $1\le d<\infty$, let $g\ge1$, and write $F_\lambda(\Hc)=\sup_mF_\lambda(\Hc,m)$.
\begin{enumerate}
\item[(i)] If $F_\lambda(\Hc)\ge Kd/g$, then $\Hc$ shatters, with some center, a panel system with $a\ge Kd/(4g)$ and $b\ge d/(4g)$.
\item[(ii)] If $\Hc$ shatters a panel system with $a\ge Kd/g$ and $b\ge d/g$, then $F_\lambda(\Hc)\ge Kd/(2g)$.
\end{enumerate}
\end{corollary}

\begin{proof}
(i) By Lemmas~\ref{lem:panelcore}(i) and~\ref{lem:kappaeventual}, $\pi_\lambda(\Hc,m)\le\kappa_\lambda(\Hc,m)\le2\lambda d$ for all $m$. The values $\min\{a,\lambda b\}\le2\lambda d$ with integers $a,b\ge0$ lie in the finite set of the proof of Lemma~\ref{lem:kappaeventual}, so $\bar\pi_\lambda(\Hc)=\pi_\lambda(\Hc,m)$ for some $m$, and $\bar\pi_\lambda(\Hc)$ is the value of a panel system that $\Hc$ shatters. By Proposition~\ref{prop:chain}(ii), $F_\lambda(\Hc,m)\le4\pi_\lambda(\Hc,m)\le4\bar\pi_\lambda(\Hc)$ for every $m$, so $\bar\pi_\lambda(\Hc)\ge F_\lambda(\Hc)/4\ge Kd/(4g)$. The attaining panel system has $a\ge Kd/(4g)$ and $\lambda b\ge Kd/(4g)$, hence $b\ge d/(4g)$ because $K\ge\lambda$.

(ii) Let the panel system live on $U$. By Lemma~\ref{lem:panelcore}(i), $\kappa_\lambda(\Hc|_U)\ge\min\{Kd/g,\lambda d/g\}=\lambda d/g$, and Proposition~\ref{prop:chain}(ii) with $m=|U|$ gives $F_\lambda(\Hc)\ge F_\lambda(\Hc,|U|)\ge\frac K{2\lambda}\cdot\frac{\lambda d}g=\frac{Kd}{2g}$.
\end{proof}

\section{No parameter independent of the sample size}\label{sec:C}\label{app:noparameter}

This appendix proves Theorem~\ref{thm:noparam}, Lemma~\ref{lem:complexes}, the two auxiliary facts used in the proof of Lemma~\ref{lem:small}, and the other claims of Section~\ref{sec:ov-plateau} and of the paragraph after Theorem~\ref{thm:noparam}.

\paragraph{Graphs and complexes.} Graphs are finite and simple, and chordal graphs, finite simplicial complexes $Y$, flagness, full subcomplexes and the systole are as in Section~\ref{sec:ov-plateau}, with $\sys(Y)=\infty$ if no full subcomplex is homeomorphic to a circle. For a graph $G$, $G[S]$ is the subgraph induced by a vertex set $S$, $N_G(v)$ is the set of neighbors of $v$, $N_G[v]=N_G(v)\cup\{v\}$, $\Cl(G)$ is the set of cliques, including $\emptyset$ and the singletons, and $\omega(G)$ is the largest size of a clique. An induced cycle of length $\ell\ge3$ is a sequence of distinct vertices $u_1,\dots,u_\ell$ in which $u_i$ and $u_j$ are adjacent if and only if $j-i\equiv\pm1\pmod\ell$, and $v$ is simplicial if $N_G(v)$ is a clique. The cliques and induced cycles of $G[S]$ are those of $G$ inside $S$, so induced subgraphs of chordal graphs are chordal. For a finite simplicial complex $Y$, $\Vc_Y=\bigcup_{F\in Y}F$ is its vertex set, a $j$-face has $j$ elements, $k(Y)$ is the largest size of a face, the edges of $G_Y$ are the $2$-faces, and $\Hc_Y$ is a class on the domain $\Vc_Y$. Faces are cliques of $G_Y$, so $Y$ is flag if and only if $Y=\Cl(G_Y)$, and the full subcomplex spanned by $S\subseteq\Vc_Y$ is $\{F\in Y:F\subseteq S\}$. For $U\subseteq\Vc_Y$
\begin{equation}\label{eq:C-proj}
\Hc_Y|_U=\{\one_F\in\{0,1\}^U:F\in Y,\ F\subseteq U\},\qquad\VC(\Hc_Y)=k(Y).
\end{equation}
Indeed $\one_F|_U=\one_{F\cap U}$ with $F\cap U\in Y$, and a face inside $U$ is its own trace. If $S$ is shattered, its all-ones pattern is the trace of a face $F\supseteq S$, so $S\in Y$; conversely all subsets of a face are faces, so faces are shattered.

\paragraph{Cores.} For a finite nonempty $V\subseteq\{0,1\}^U$, $N_V(v)$ and $\deg_V(v)=|N_V(v)|$ refer to $G(V)$, and $v^{(x)}$ is the pattern that differs from $v$ exactly at $x$. A pattern has at most one neighbor in each direction, so every two-sided core $(A,W)$ and $a\in A$ satisfy $\min\{\delta_A,\lambda\delta_W\}\le\delta_A\le\deg_V(a)\le\Delta(V)\le|U|$. As $\kappa_\lambda(V)=0$ when $G(V)$ has no edge, for every $\lambda\in[0,\infty)$
\begin{equation}\label{eq:C-kappaDelta}
\kappa_\lambda(V)\ \le\ \Delta(V)\ \le\ |U|.
\end{equation}
The star number is $s(\Hc)=\sup_U\Delta(\Hc|_U)$ over finite $U\subseteq\X$ (Section~\ref{sec:panels}), and $\bar\kappa_\lambda(\Hc)=\sup_{m\ge0}\kappa_\lambda(\Hc,m)$.

\subsection{Clique complexes of chordal graphs}\label{sec:C-chordal}

\begin{lemma}[\citealp{Dir61}]\label{lem:dirac}
Every chordal graph with at least one vertex has a simplicial vertex.
\end{lemma}

\begin{lemma}[No large two-sided cores]\label{lem:chordalcore}
Let $G$ be a chordal graph on a finite set $U$ with $\omega(G)\le k$, where $k\ge1$, and let $V=\{\one_F:F\in\Cl(G)\}\subseteq\{0,1\}^U$. Then every two-sided core $(A,W)$ of $G(V)$ has $\delta_A\le k$ or $\delta_W\le1$. Consequently $\kappa_\lambda(V)\le\max\{k,\lambda\}$ for every $\lambda\in[0,\infty)$.
\end{lemma}

\begin{proof}
We identify $\one_F$ with $F$; two cliques are adjacent in $G(V)$ if and only if one is obtained from the other by adding one vertex. We show by induction on $|U|$ that no two-sided core has $\delta_A\ge k+1$ and $\delta_W\ge2$. If $U=\emptyset$, $V$ has one element and there is no two-sided core. Let $|U|\ge1$, suppose that $(A,W)$ is a two-sided core with $\delta_A\ge k+1$ and $\delta_W\ge2$, and let $v$ be a simplicial vertex of $G$ (Lemma~\ref{lem:dirac}). Then $N_G[v]$ is a clique, so $|N_G[v]|\le k$.

No member of $A$ contains $v$: a clique $F\ni v$ satisfies $F\subseteq N_G[v]$, and its neighbors in $G(V)$ are the sets $F\setminus\{x\}$ with $x\in F$ and the cliques $F\cup\{y\}$ with $y\notin F$, where $y$ is adjacent to $v$ and so $y\in N_G[v]\setminus F$. Hence $\deg_V(F)\le|F|+|N_G[v]\setminus F|=|N_G[v]|\le k<\delta_A$. No member of $W$ contains $v$: every neighbor of a clique $F\ni v$ other than $F\setminus\{v\}$ contains $v$ and is not in $A$, so $F$ has at most one neighbor in $A$, while $\delta_W\ge2$.

Let $G'=G[U\setminus\{v\}]$ and $V'=\{\one_F:F\in\Cl(G')\}\subseteq\{0,1\}^{U\setminus\{v\}}$. Then $G'$ is chordal with $\omega(G')\le k$, its cliques are the cliques of $G$ that avoid $v$, and two of them are adjacent in $G(V')$ if and only if they are adjacent in $G(V)$. All members of $A\cup W$ avoid $v$, and $\delta_A,\delta_W$ count only neighbors in $A\cup W$, so $(A,W)$ is a two-sided core of $G(V')$ with $\delta_A\ge k+1$ and $\delta_W\ge2$, contradicting the induction hypothesis.

For the consequence, $\kappa_\lambda(V)=0$ if $G(V)$ has no edge, and otherwise every two-sided core has $\min\{\delta_A,\lambda\delta_W\}\le\delta_A\le k$ or $\min\{\delta_A,\lambda\delta_W\}\le\lambda\delta_W\le\lambda$.
\end{proof}

\begin{lemma}[Sets below the systole]\label{lem:systole}
Let $Y$ be a finite simplicial complex and $U\subseteq\Vc_Y$ with $|U|<\sys(Y)$. Then $G_Y[U]$ is chordal. If $Y$ is flag, the faces of $Y$ inside $U$ are the cliques of $G_Y[U]$, and $\Hc_Y|_U=\{\one_F:F\in\Cl(G_Y[U])\}$.
\end{lemma}

\begin{proof}
Suppose that $G_Y[U]$ has an induced cycle $u_1,\dots,u_\ell$ with $\ell\ge4$, and let $S=\{u_1,\dots,u_\ell\}$. It is an induced cycle of $G_Y$, so no three of its vertices are pairwise adjacent, and as faces are cliques, the faces inside $S$ are $\emptyset$, the singletons and the pairs $\{u_i,u_{i+1}\}$, with indices modulo $\ell$. The full subcomplex spanned by $S$ is therefore a cycle with $\ell$ edges, homeomorphic to a circle, and $\sys(Y)\le\ell\le|U|$, a contradiction. If $Y$ is flag, its faces are the cliques of $G_Y$, so the faces inside $U$ are the cliques of $G_Y[U]$, and~\eqref{eq:C-proj} gives the last claim.
\end{proof}

\noindent Lemmas~\ref{lem:chordalcore} and~\ref{lem:systole} give Lemma~\ref{lem:small}: under its hypotheses, $G_Y[U]$ is chordal, $\Hc_Y|_U=\{\one_F:F\in\Cl(G_Y[U])\}$, and $\omega(G_Y[U])\le k$ because cliques of $G_Y[U]$ are faces of $Y$.

\subsection{Complexes with large systole}\label{sec:C-complex}

The complexes of Lemma~\ref{lem:complexes} are developments of simplices of finite groups constructed by \citet[Section~3]{ABJLMS09}. We use the following definitions and results of that section (page numbers of the journal version). Let $\Delta_0$ be a simplex, viewed as the complex of its nonempty faces, with geometric realization $|\Delta_0|$, and for $p\in|\Delta_0|$ let $\sigma_p$ be the smallest face with $p\in|\sigma_p|$.
\begin{enumerate}
\item[(A1)] (\S3.1, p.~1241.) An extended simplex of groups $\bG^+$ over $\Delta_0$ has groups $\bG(\sigma)$ for the faces $\sigma$, including $\Pi=\bG(\emptyset)$, and compatible injective homomorphisms $\bG(\sigma)\to\bG(\tau)$ for $\tau\subseteq\sigma$, through which $\bG(\sigma)\subseteq\bG(\tau)\subseteq\Pi$ whenever $\tau\subseteq\sigma$.
\item[(A2)] (\S3.1, p.~1242.) The development is $\Dev(\bG^+)=(|\Delta_0|\times\Pi)/{\sim}$, where $(p,g)\sim(p',h)$ if and only if $p=p'$ and $g^{-1}h\in\bG(\sigma_p)$. Its simplices are the images $[\sigma,g]$ of the sets $|\sigma|\times\{g\}$, and $[p,g]$ is the class of $(p,g)$. For the $\bG^+$ of (A3), $\Dev(\bG^+)$ is a simplicial complex.
\item[(A3)] (\S3.4, p.~1251.) For every integer $r\ge1$, every simplex $\Delta_0$ of dimension at least $1$ and all groups $\bG(s)$ on its codimension-one faces $s$, there is an $r$-retractible (Definition~3.19, p.~1246) extended simplex of groups over $\Delta_0$ with these groups and $\bG(\Delta_0)=\{1\}$, built from minimal $r$-retractible extensions (Theorem~3.24(1), p.~1249).
\item[(A4)] (Definition~3.26 and the paragraph after it, p.~1252, restating Theorem~3.24(2), p.~1249.) In (A3), $\Pi$ is an $r$-retra-product of a finite family of the chosen groups, and $r$-retra-products of finite groups are finite.
\item[(A5)] (\S3.5, p.~1253, after \citealp{JS06}.) The link of a simplex $\sigma$ of $Y$ consists of the simplices $\tau$ disjoint from $\sigma$ with $\tau\cup\sigma\in Y$, and $Y$ is $\ell$-large if $Y$ and the links of all its simplices have systole at least $\ell$.
\item[(A6)] (Proposition~3.30, p.~1253, which rests on \citealp[Corollary~1.5]{JS06}.) If $\bG^+$ is $r$-retractible, then $\Dev(\bG^+)$ is $2(r+1)$-large.
\end{enumerate}

\paragraph{The complexes $X_{k,q,r}$.} Let $k,q,r\ge2$ be integers, $\Delta_0$ the simplex on $\{v_1,\dots,v_k\}$, of dimension $k-1\ge1$, and $\bG^+$ the extended simplex of groups of (A3) with parameter $r$ and $\bG(s)=\Z_q$ on every codimension-one face $s$. By (A2), (A3) and (A6), $\Dev(\bG^+)$ is a $2(r+1)$-large simplicial complex; by (A4), $\Pi$ is finite; by (A3) and (A1), $\bG(\Delta_0)=\{1\}$ and each $\bG(s)$ is a subgroup of $\Pi$ of order $q$. Let $X_{k,q,r}$ be the complex whose vertices are the points $[v,g]$, with $v$ a vertex of $\Delta_0$ and $g\in\Pi$, and whose faces are $\emptyset$ and the vertex sets of the simplices $[\sigma,g]$. As the simplices of $\Dev(\bG^+)$ correspond to their vertex sets, $X_{k,q,r}$ has the same full subcomplexes, links and systoles, and is $2(r+1)$-large. Let $\Vc_{k,q,r}$ be its vertex set and $\Hc_{k,q,r}=\Hc_{X_{k,q,r}}$.

\begin{lemma}[Properties of $X_{k,q,r}$]\label{lem:complex}
Let $k,q,r\ge2$ be integers and $X=X_{k,q,r}$.
\begin{enumerate}
\item[(a)] $X$ is a finite simplicial complex with $k(X)=k$, its $k$-faces are the vertex sets of the simplices $[\Delta_0,g]$, $g\in\Pi$, and every face lies in a $k$-face.
\item[(b)] $X$ is flag.
\item[(c)] Every $(k-1)$-subset of a $k$-face lies in at least $q$ of the $k$-faces.
\item[(d)] $\sys(X)\ge2(r+1)$. Hence, for every set $U$ of at most $2r+1$ vertices, $G_X[U]$ is chordal and its cliques are the faces of $X$ inside $U$.
\item[(e)] The set $\Ac$ of $(k-1)$-faces and the set $\Wc$ of $k$-faces form a panel system of rank $k$ on $\Vc_{k,q,r}$ with $a(\Ac,\Wc)\ge q$ and $b(\Ac,\Wc)=k$, which $\Hc_{k,q,r}$ shatters with center $0$, and $A=\{\one_P:P\in\Ac\}$, $W=\{\one_F:F\in\Wc\}$ form a two-sided core of $G(\Hc_{k,q,r})$ with $\delta_A\ge q$ and $\delta_W=k$.
\item[(f)] If $q\ge k+1$, then $|\Vc_{k,q,r}|\ge2r+2$.
\end{enumerate}
\end{lemma}

\begin{proof}
(a) By (A2) the map $[p,g]\mapsto p$ is well defined, so $[v,g]=[w,h]$ implies $v=w$. Hence the vertex set $\{[v,g]:v\in\sigma\}$ of $[\sigma,g]$ has $|\sigma|$ elements, it has $k$ elements only if $\sigma=\Delta_0$, it lies in the vertex set of $[\Delta_0,g]$, and its subsets, including the singletons, are the vertex sets of the simplices $[\sigma',g]$, $\sigma'\subseteq\sigma$. There are at most $k|\Pi|$ vertices.

(b) As $2(r+1)\ge6$, $X$ is $6$-large. Let $S$ be a clique of $G_X$ of minimal size that is not a face. Cliques with at most two elements are faces, so $|S|\ge3$, and all proper subsets of $S$ are faces. Pick distinct $a,b,c\in S$ and let $\sigma=S\setminus\{a,b,c\}$. In the link of $\sigma$, or in $X$ if $\sigma=\emptyset$, all subsets of $\{a,b,c\}$ with one or two elements are simplices, since their unions with $\sigma$ are proper subsets of $S$, while $\{a,b,c\}$ is not, since its union with $\sigma$ is $S$. So the full subcomplex of this link, or of $X$, spanned by $\{a,b,c\}$ is a circle with three edges, which contradicts $6$-largeness.

(c) By (a) a $k$-face is the vertex set of some $[\Delta_0,g]$, and its $(k-1)$-subsets are the sets $\{[v,g]:v\in s\}$ for the codimension-one faces $s$. Let $h\in g\bG(s)$. For a vertex $v\in s$, $\sigma_v=\{v\}$ and $g^{-1}h\in\bG(s)\subseteq\bG(\{v\})$ by (A1), so $[v,g]=[v,h]$ by (A2), and the vertex set of $[\Delta_0,h]$ contains $\{[v,g]:v\in s\}$. For distinct $h,h'\in g\bG(s)$ and $p$ in the interior of $|\Delta_0|$, $\sigma_p=\Delta_0$ and $h^{-1}h'\notin\bG(\Delta_0)=\{1\}$, so $[p,h]\ne[p,h']$ and $[\Delta_0,h]\ne[\Delta_0,h']$. Distinct simplices of a simplicial complex have distinct vertex sets, which gives $|\bG(s)|=q$ distinct $k$-faces.

(d) The first claim is part of $2(r+1)$-largeness, and Lemma~\ref{lem:systole} with (b) gives the second, as $|U|<\sys(X)$.

(e) By (a), $\Wc\ne\emptyset$ and every member of $\Ac$ lies in a member of $\Wc$, hence in at least $q$ of them by (c); every member of $\Wc$ contains exactly $k$ members of $\Ac$. This gives the panel system, and $\Hc_{k,q,r}$ contains $0\oplus\one_S=\one_S$ for every face $S$. For $P\in\Ac$ and $F\in\Wc$, $\one_P$ and $\one_F$ are adjacent if and only if $|P\triangle F|=1$, that is, $P\subset F$, and $A\cap W=\emptyset$ as the sizes differ. So every member of $A$ has at least $q$ neighbors in $W$ and every member of $W$ has exactly $k$ neighbors in $A$.

(f) As $q\ge k+1$ and $k\ge2$, the core of (e) has $\delta_A\ge k+1$ and $\delta_W\ge2$. If $|\Vc_{k,q,r}|<\sys(X)$, Lemma~\ref{lem:small} with $Y=X$ and $U=\Vc_{k,q,r}$, which applies by (a) and (b), would give $\delta_A\le k$ or $\delta_W\le1$. Hence $|\Vc_{k,q,r}|\ge\sys(X)\ge2(r+1)$.
\end{proof}

\noindent Lemma~\ref{lem:complex}(a)--(d) is Lemma~\ref{lem:complexes}.

\subsection{The plateau classes}\label{sec:C-plateau}\label{app:kappagap}

\begin{proposition}[Plateau classes]\label{prop:plateau}
Let $k\ge2$, $q\ge k+1$ and $r\ge2$ be integers, $\Hc=\Hc_{k,q,r}$, a class on the finite domain $\Vc_{k,q,r}$, and $N_0=|\Vc_{k,q,r}|$. Then $\Hc$ is nontrivial with VC dimension $k$, $N_0\ge2r+2$, and
\begin{enumerate}
\item[(i)] $\kappa_\lambda(\Hc,m)\le\max\{k,\lambda\}$ for every $\lambda\in[0,\infty)$ and every integer $0\le m\le2r+1$;
\item[(ii)] $\bar\kappa_\lambda(\Hc)\ge\min\{q,\lambda k\}$ for every $\lambda\in[0,\infty)$, and $\tau_\lambda(\Hc)\ge kKq/(k\lambda+q)$ for every $\lambda\in[1,\infty)$;
\item[(iii)] for every $\lambda\in[1,\infty)$, $(n+1)M_n(\Hc,\lambda)\le2\max\{k,\lambda\}$ for $1\le n\le2r$, and $N_0M_{N_0-1}(\Hc,\lambda)\ge kKq/(e(k\lambda+q))$.
\end{enumerate}
\end{proposition}

\begin{proof}
By~\eqref{eq:C-proj} and Lemma~\ref{lem:complex}(a), $\VC(\Hc)=k\ge2$, and a shattered pair $\{x,x'\}$ gives $h,h'\in\Hc$ with $h(x)=h'(x)=h(x')=0\ne h'(x')$, so $\Hc$ is nontrivial. Lemma~\ref{lem:complex}(f) gives $N_0\ge2r+2$.

(i) For $U\subseteq\Vc_{k,q,r}$ with $|U|\le m$, $|U|<\sys(X_{k,q,r})$ by Lemma~\ref{lem:complex}(d), and Lemma~\ref{lem:small}, which applies by Lemma~\ref{lem:complex}(a),(b), gives $\kappa_\lambda(\Hc|_U)\le\max\{k,\lambda\}$; take the maximum over $U$.

(ii) For the core $(A,W)$ of Lemma~\ref{lem:complex}(e), $\bar\kappa_\lambda(\Hc)\ge\kappa_\lambda(\Hc|_{\Vc_{k,q,r}})\ge\min\{\delta_A,\lambda\delta_W\}\ge\min\{q,\lambda k\}$. Let $\lambda\ge1$. The edges between $A$ and $W$ are the pairs $\{\one_P,\one_F\}$ with $P\subset F$, so $|E(A,W)|=k|W|$, and counting them from the side of $A$ gives $q|A|\le k|W|$. Hence $\rho_\lambda(\Hc)\ge k|W|/(|A|+|W|/\lambda)\ge k|W|/(k|W|/q+|W|/\lambda)=k\lambda q/(k\lambda+q)$, and Lemma~\ref{lem:twolevel} gives $\tau_\lambda(\Hc)\ge(K/\lambda)\rho_\lambda(\Hc)\ge kKq/(k\lambda+q)$.

(iii) For $1\le n\le2r$, \eqref{eq:transfer}, Theorem~\ref{thm:binary}(b) and (i) with $m=n+1\le2r+1$ give $(n+1)M_n(\Hc,\lambda)\le F_\lambda(\Hc,n+1)\le2\kappa_\lambda(\Hc,n+1)\le2\max\{k,\lambda\}$. For $n=N_0-1\ge1$ and $U=\Vc_{k,q,r}$, $|U|=n+1$ and $\Hc|_U=\Hc$, so Lemma~\ref{lem:uniform} and (ii) give $M_n(\Hc,\lambda)\ge(n+1)^{-1}(n/(n+1))^n\tau_\lambda(\Hc)\ge kKq/(e(n+1)(k\lambda+q))$, since $(n/(n+1))^n=(1+1/n)^{-n}\ge1/e$.
\end{proof}

\begin{remark}[The case $k=2$]\label{rem:C-k2}
Let $q,r\ge2$ and $G$ the $1$-skeleton of $X=X_{2,q,r}$. By Lemma~\ref{lem:complex}(a),(b), the faces of $X$ are $\emptyset$, the singletons and the edges of $G$, and $G$ has no triangle, as its vertex set would be a $3$-face. By Lemma~\ref{lem:complex}(c), every endpoint of an edge has at least $q$ neighbors. A shortest cycle of $G$ has no chord, which would split it into two shorter cycles, so it spans a full subcomplex homeomorphic to a circle and has at least $\sys(X)\ge2(r+1)$ edges. Hence $G[U]$ is a forest whenever $|U|\le2r+1$, and the panel system of Lemma~\ref{lem:complex}(e) consists of the singletons and the edges, with $a\ge q$ and $b=2$.
\end{remark}

\subsection{The core number on all points}\label{sec:C-global}\label{app:kappapos}

\begin{proposition}[Comparison with $\bar\kappa_\lambda$]\label{prop:kappaglobal}
Let $\lambda\ge1$ and let $\Hc$ have VC dimension $d\ge1$, possibly $d=\infty$, and star number $s$. Write $\bar\kappa=\bar\kappa_\lambda(\Hc)$. Then $\kappa_\lambda(\Hc,m)\ge\min\{m,\bar\kappa\}/(2\min\{\lambda,d\})$ for every integer $m\ge1$, with $\min\{\lambda,\infty\}=\lambda$. Moreover (i) $\kappa_\lambda(\Hc,m)\ge\min\{m,s,\lambda\}$ for every $m\ge1$; (ii) $\bar\kappa\le s$; (iii) $\bar\kappa\le2\lambda d$ if $d<\infty$. In particular, every nonempty class $\Hc$ of VC dimension $d\in\{0,1,\dots\}\cup\{\infty\}$ satisfies $\min\{m,\bar\kappa_\lambda(\Hc)\}\le4\min\{K,d\}\,\kappa_\lambda(\Hc,m)$ for every integer $m\ge0$, with $\min\{K,\infty\}=K$.
\end{proposition}

\begin{proof}
(i) If $s=0$ there is nothing to prove. Otherwise $D=\min\{m,s\}\ge1$, and some finite $U$ has $\Delta(\Hc|_U)\ge D$, because $D\le s$ and $s$ is a supremum of integers, attained when finite. Let $v\in\Hc|_U$ have degree at least $D$, let $x_1,\dots,x_D$ be the directions of $D$ of its edges and $T=\{x_1,\dots,x_D\}$. The patterns $c=v|_T$ and $c_i=v^{(x_i)}|_T$ lie in $\Hc|_T$, are pairwise distinct, and $c$ is adjacent to every $c_i$. So $(\{c\},\{c_1,\dots,c_D\})$ is a two-sided core with $\delta_A=D$ and $\delta_W=1$, and $\kappa_\lambda(\Hc,m)\ge\kappa_\lambda(\Hc|_T)\ge\min\{D,\lambda\}$, as $|T|=D\le m$.

(ii) By~\eqref{eq:C-kappaDelta}, $\kappa_\lambda(\Hc|_U)\le\Delta(\Hc|_U)\le s$ for every finite $U$.

(iii) Let $V=\Hc|_U$ with $U$ finite. If $G(V)$ has no edge, $\kappa_\lambda(V)=0$. Otherwise write $d_V=\VC(V)\le d$. For disjoint $A,W\subseteq V$, not both empty, the edge bound $|E_V(B)|\le d_V|B|$ of \citet{HLW94} (see also \citealp{Hau95}) for the edges of $G(V)$ inside $B=A\cup W$ gives $|E(A,W)|\le d_V(|A|+|W|)\le\lambda d_V(|A|+|W|/\lambda)$. Hence $\rho_\lambda(V)\le\lambda d_V$, and Lemma~\ref{lem:corerho} gives $\kappa_\lambda(V)\le2\rho_\lambda(V)\le2\lambda d_V\le2\lambda d$.

\emph{The bound.} Lemma~\ref{lem:panelcore} gives $\kappa_\lambda(\Hc,m)\ge d_m$, which settles $d=\infty$, where $d_m=m$.
Let $1\le d<\infty$. If $m\le\min\{s,\lambda\}$, (i) gives $\kappa_\lambda(\Hc,m)\ge m$. If $s\le\lambda$ and $s<m$, (i) and (ii) give $\kappa_\lambda(\Hc,m)\ge s\ge\bar\kappa$. In the remaining case $\lambda<\min\{s,m\}$, and (i) gives $\kappa_\lambda(\Hc,m)\ge\lambda$. If $m\le d$, then $\kappa_\lambda(\Hc,m)\ge d_m=m$. If $m>d$, then $\kappa_\lambda(\Hc,m)\ge\max\{\lambda,d\}$ and, by (iii), $\min\{m,\bar\kappa\}\le2\lambda d=2\max\{\lambda,d\}\min\{\lambda,d\}\le2\min\{\lambda,d\}\,\kappa_\lambda(\Hc,m)$.
In all cases the bound holds, since $2\min\{\lambda,d\}\ge1$. For the last claim, a class of VC dimension $0$ has one member, since two members that differ at $x$ shatter $\{x\}$, so $\bar\kappa_\lambda(\Hc)=0$; both sides vanish if $d=0$ or $m=0$, and otherwise the bound applies with $\min\{\lambda,d\}\le\min\{K,d\}$.
\end{proof}

\noindent For $\lambda\ge1$ and a class $\Hc$ of finite VC dimension, $\kappa_\lambda(\Hc,m)$ is nondecreasing in $m$ and equals $\bar\kappa_\lambda(\Hc)$ for all large $m$ (Lemma~\ref{lem:kappaeventual}); let $m_0(\Hc)$ be the smallest $m_0\ge0$ with $\kappa_\lambda(\Hc,m)=\bar\kappa_\lambda(\Hc)$ for all $m\ge m_0$.

\begin{corollary}\label{cor:C-cutoff}
There is no function $M$ from triples $(d,\lambda,\bar\kappa)$ to $[0,\infty)$ such that $m_0(\Hc)\le M(d,\lambda,\bar\kappa_\lambda(\Hc))$ for every $\lambda\ge1$ and every class $\Hc$ of finite VC dimension $d$.
\end{corollary}

\begin{proof}
Fix integers $k,\lambda\ge2$, let $q=\lambda k\ge k+1$ and $\Hc_r=\Hc_{k,q,r}$ for $r\ge2$, of VC dimension $k$. Since the domain $\Vc_{k,q,r}$ is finite, $\kappa_\lambda(\Hc_r,m)$ is the same for all $m\ge|\Vc_{k,q,r}|$, so $\bar\kappa_\lambda(\Hc_r)$ is attained by a projection. By Propositions~\ref{prop:plateau}(ii) and~\ref{prop:kappaglobal}(iii), $0<\lambda k\le\bar\kappa_\lambda(\Hc_r)\le2\lambda k$, so $\bar\kappa_\lambda(\Hc_r)$ is the value $\min\{a,\lambda b\}\le2\lambda k$ of a two-sided core, with integers $a,b\ge1$; it is $a\le2\lambda k$ or $\lambda b$ with $b\le2k$, and so lies in the finite set $\{1,\dots,2\lambda k\}\cup\{\lambda b:1\le b\le2k\}$, which does not depend on $r$. By Proposition~\ref{prop:plateau}(i), $\kappa_\lambda(\Hc_r,2r+1)\le\max\{k,\lambda\}<\lambda k\le\bar\kappa_\lambda(\Hc_r)$, so $m_0(\Hc_r)>2r+1$ for every $r$, while a function $M$ as in the statement would bound $m_0(\Hc_r)$ by its largest value at the finitely many triples $(k,\lambda,\kappa)$ with $\kappa$ in this set.
\end{proof}

\subsection{Proof of Theorem~\ref{thm:noparam}}\label{sec:C-proof}

\restatemain{thm:noparam}

\noindent Fix $\lambda\in[1,\infty)$ and $K=1+\lambda$. For a family $\Kc$ of nontrivial classes and a map $D:\Kc\to[0,\infty]$, constants $0<c\le C<\infty$ are admissible if for all $\Hc\in\Kc$ and $n\ge1$
\begin{equation}\label{eq:C-adm}
c\,\min\{D(\Hc),n+1\}\ \le\ (n+1)\,M_n(\Hc,\lambda)\ \le\ C\,\min\{D(\Hc),n+1\}.
\end{equation}
Let $\Kc_d$ be the family of nontrivial classes of VC dimension $d$ and $\Kc_{\mathrm{all}}$ that of all nontrivial classes. Constants admissible for $D$ on $\Kc$ are admissible for the restriction of $D$ to any $\Kc'\subseteq\Kc$.

\begin{proof}[Proof of Theorem~\ref{thm:noparam}(a),(b)]
(a) Let $d\ge2$, let $(c,C)$ be admissible for $D:\Kc_d\to[0,\infty]$, and let $q\ge d+1$ be an integer. Choose an integer $r\ge2$ with $c(2r+1)>2\max\{d,\lambda\}$, and let $\Hc=\Hc_{d,q,r}\in\Kc_d$ (Proposition~\ref{prop:plateau}). By~\eqref{eq:C-adm} with $n=2r$ and Proposition~\ref{prop:plateau}(iii),
\[
c\,\min\{D(\Hc),2r+1\}\ \le\ (2r+1)M_{2r}(\Hc,\lambda)\ \le\ 2\max\{d,\lambda\}\ <\ c(2r+1),
\]
so $D(\Hc)<2r+1$ and $D(\Hc)\le2\max\{d,\lambda\}/c$. With $N_0=|\Vc_{d,q,r}|$ and $n=N_0-1\ge1$, \eqref{eq:C-adm} and Proposition~\ref{prop:plateau}(iii) give
\[
\frac{dKq}{e(d\lambda+q)}\ \le\ N_0M_{N_0-1}(\Hc,\lambda)\ \le\ C\min\{D(\Hc),N_0\}\ \le\ C\,D(\Hc)\ \le\ \frac{2C\max\{d,\lambda\}}{c}.
\]
So $C/c\ge dKq/(2e(d\lambda+q)\max\{d,\lambda\})$ for every integer $q\ge d+1$, and as $c$, $C$, $D$ do not depend on $q$, letting $q\to\infty$ gives $C/c\ge dK/(2e\max\{d,\lambda\})$. This is at least $\min\{K,d\}/(2e)$: if $d\ge\lambda$ it equals $K/(2e)$, and if $d<\lambda$ it is $dK/(2e\lambda)\ge d/(2e)$.

(b) Constants admissible for $D$ on $\Kc_{\mathrm{all}}$ are admissible for its restriction to $\Kc_d$, and (a) with an integer $d\ge\max\{2,\lambda\}$ gives $C/c\ge dK/(2e\max\{d,\lambda\})=K/(2e)$.
\end{proof}

\begin{proof}[Proof of Theorem~\ref{thm:noparam}(c)]
Let $\Hc$ be nontrivial, $n\ge1$ and $m=n+1$. A nontrivial class has $h,h'\in\Hc$ and $x'$ with $h(x')\ne h'(x')$, so $\{x'\}$ is shattered and $d\ge1$.

\emph{The core number on $\Kc_d$.} By~\eqref{eq:transfer}, Theorem~\ref{thm:binary}(b) with $K/(2\lambda)\ge\frac12$, and Proposition~\ref{prop:kappaglobal},
\[
M_n(\Hc,\lambda)\ \ge\ \frac{F_\lambda(\Hc,m)}{2em}\ \ge\ \frac{\kappa_\lambda(\Hc,m)}{4em}\ \ge\ \frac{\min\{m,\bar\kappa_\lambda(\Hc)\}}{8e\min\{\lambda,d\}\,m}.
\]
Conversely, $\kappa_\lambda(\Hc,m)\le\bar\kappa_\lambda(\Hc)$ by definition and $\kappa_\lambda(\Hc,m)\le m$ by~\eqref{eq:C-kappaDelta}, so~\eqref{eq:transfer} and Theorem~\ref{thm:binary}(b) give $mM_n(\Hc,\lambda)\le F_\lambda(\Hc,m)\le2\kappa_\lambda(\Hc,m)\le2\min\{m,\bar\kappa_\lambda(\Hc)\}$. Hence $c=1/(8e\min\{\lambda,d\})$ and $C=2$ are admissible for $\bar\kappa_\lambda$ on $\Kc_d$, with $C/c=16e\min\{\lambda,d\}\le16e\min\{K,d\}$.

\emph{The VC dimension on $\Kc_{\mathrm{all}}$.} By~\eqref{eq:transfer}, Theorem~\ref{thm:binary}(b) and Lemma~\ref{lem:panelcore},
\[
mM_n(\Hc,\lambda)\ \ge\ \frac{F_\lambda(\Hc,m)}{2e}\ \ge\ \frac{K\kappa_\lambda(\Hc,m)}{4e\lambda}\ \ge\ \frac{d_m}{4e}.
\]
Conversely, $f_K(a,b)\le\min\{Ka,Kb\}=Kf_1(a,b)$, so $\tau_\lambda(V)\le K\tau_0(V)$ for every finite nonempty $V$, and Theorem~\ref{thm:binary}(a) at $\lambda=0$ gives $F_\lambda(\Hc,m)\le KF_0(\Hc,m)\le Kd_m$; so $mM_n(\Hc,\lambda)\le Kd_m$ by~\eqref{eq:transfer}. Hence $c=1/(4e)$ and $C=K$ are admissible for the VC dimension on $\Kc_{\mathrm{all}}$, with $C/c=4eK$.
\end{proof}

\noindent By parts~(a) and~(c), on the classes of VC dimension $d\ge2$ the infimum of the admissible ratios $C/c$ over all maps $D$ lies between $\min\{K,d\}/(2e)$ and $16e\min\{K,d\}$, and on all nontrivial classes between $K/(2e)$ and $4eK$.

\section{Rates of natural classes}\label{sec:D}\label{app:applications}

This appendix proves Corollary~\ref{cor:natural}, the statement on sample sizes after it, and the claims of Section~\ref{sec:ov-panels} on the entropy bound, affine halfspaces and dictators, by bounding $F_\lambda(\Hc,m)$ from both sides class by class (Sections~\ref{sec:D-sat}--\ref{sec:D-log}) and applying the transfer~\eqref{eq:transfer} (Section~\ref{sec:D-proof}).

\paragraph{Notation.} Throughout, $U$ is a finite set and $V\subseteq\{0,1\}^U$ is nonempty; $\Phi_\lambda(\mu;V)=\sum_{\{u,w\}\in E(V)}f_K(\mu_u,\mu_w)$ is the objective of Section~\ref{sec:A}, so that $\tau_\lambda(V)=\max_\mu\Phi_\lambda(\mu;V)$. All finite parameters that count points, sets, groups, blocks, layers or dimensions ($k$, $m$, $n$, $N$, $p$, $D$, $d$, $r$, $j$, $t$, $u$) are integers. The projection on $U=\emptyset$ is a single pattern, with $\tau_\lambda=0$ and VC dimension $0$; every bound below holds for it trivially, so in the proofs $U$ is nonempty. For $x\in U$ let $e_x:=\one_{\{x\}}$, and for $c\in\{0,1\}^U$ let $v\oplus c$ be coordinatewise addition modulo $2$. For a finite nonempty set $T$, the \emph{star} on $T$ is $S(T):=\{0\}\cup\{e_x:x\in T\}\subseteq\{0,1\}^T$ and the \emph{cube} is $Q(T):=\{0,1\}^T$. For integers $t\ge1$ let $\sigma_\lambda(t):=tK/(t+\max\{1,\lambda\})$, and $\sigma_\lambda(0):=0$. Let $\Delta_m(\Hc):=\max_{U\subseteq\X,\,|U|\le m}\Delta(\Hc|_U)$, so that the star number is $s=\sup_m\Delta_m(\Hc)$. Logarithms $\log_2$ are in base $2$, and $h(q):=-q\log_2q-(1-q)\log_2(1-q)$, with $h(0)=h(1)=0$, is the binary entropy.

\subsection{Tools}\label{sec:D-tools}

\begin{lemma}[Invariance and monotonicity]\label{lem:Dinv}
Let $\lambda\in[0,\infty)$.
\begin{enumerate}
\item[(i)] If $\phi:U\to U'$ is a bijection and $c\in\{0,1\}^U$, then $v\mapsto(v\oplus c)\circ\phi^{-1}$ is an isomorphism from $G(V)$ onto $G(V')$, where $V'$ is the image of $V$, and $\tau_\lambda(V')=\tau_\lambda(V)$. If all elements of $V$ agree on $U\setminus W$, then $v\mapsto v|_W$ is an isomorphism from $G(V)$ onto $G(V|_W)$ and $\tau_\lambda(V|_W)=\tau_\lambda(V)$.
\item[(ii)] If $\emptyset\ne V'\subseteq V$, then $\tau_\lambda(V')\le\tau_\lambda(V)$.
\item[(iii)] If $\emptyset\ne\Hc'\subseteq\Hc$, then $F_\lambda(\Hc',m)\le F_\lambda(\Hc,m)$ for every $m\ge1$.
\end{enumerate}
\end{lemma}

\begin{proof}
(i) Both maps are bijections that preserve Hamming distances (in the second case elements of $V$ differ only inside $W$), hence adjacency, and $\tau_\lambda(V)$ depends only on $G(V)$. (ii) Extend $\mu'\in\Pc(V')$ by zero to $V$. Every edge of $G(V')$ is an edge of $G(V)$ and every term $f_K$ is nonnegative (Lemma~\ref{lem:fK}), so $\Phi_\lambda(\mu';V')\le\Phi_\lambda(\mu';V)\le\tau_\lambda(V)$. (iii) $\Hc'|_U\subseteq\Hc|_U$ for every $U$; apply (ii).
\end{proof}

\begin{lemma}[Product additivity]\label{lem:product}
Let $U_1,U_2$ be disjoint finite sets, let $V_i\subseteq\{0,1\}^{U_i}$ be nonempty, and identify $\{0,1\}^{U_1\cup U_2}$ with $\{0,1\}^{U_1}\times\{0,1\}^{U_2}$. Then $\tau_\lambda(V_1\times V_2)=\tau_\lambda(V_1)+\tau_\lambda(V_2)$ for every $\lambda\in[0,\infty)$.
\end{lemma}

\begin{proof}
Two elements $(v_1,v_2)$ and $(w_1,w_2)$ of $V_1\times V_2$ are adjacent if and only if $v_2=w_2$ and $\{v_1,w_1\}\in E(V_1)$, or $v_1=w_1$ and $\{v_2,w_2\}\in E(V_2)$. So $E(V_1\times V_2)$ is the disjoint union of $E_1=\{\{(v_1,z),(w_1,z)\}:\{v_1,w_1\}\in E(V_1),\,z\in V_2\}$ and the analogous set $E_2$. For the lower bound let $\mu_i$ attain $\tau_\lambda(V_i)$; under $\mu_1\otimes\mu_2$, homogeneity of $f_K$ (Lemma~\ref{lem:fK}) shows that $E_1$ contributes $\sum_z\mu_2(z)\Phi_\lambda(\mu_1;V_1)=\tau_\lambda(V_1)$ and $E_2$ contributes $\tau_\lambda(V_2)$. For the upper bound fix $\mu\in\Pc(V_1\times V_2)$ and let $\nu(z)=\sum_{v\in V_1}\mu(v,z)$. By homogeneity the edges of $E_1$ with second component $z$ contribute $0$ if $\nu(z)=0$ and $\nu(z)\Phi_\lambda(\mu(\cdot,z)/\nu(z);V_1)\le\nu(z)\tau_\lambda(V_1)$ otherwise, so $E_1$ contributes at most $\tau_\lambda(V_1)$; likewise $E_2$ contributes at most $\tau_\lambda(V_2)$.
\end{proof}

\begin{lemma}[Stars]\label{lem:star}
Let $\lambda\in[0,\infty)$ and $|T|=t\ge1$. Then $\tau_\lambda(S(T))=\sigma_\lambda(t)$. Moreover $\sigma_\lambda$ is nondecreasing, $\min\{t,K\}/2\le\sigma_\lambda(t)\le\min\{t,K\}$, and $\sigma_\lambda(1)=1$ if $\lambda\ge1$.
\end{lemma}

\begin{proof}
$G(S(T))$ has the $t$ edges $\{0,e_x\}$, since $e_x$ and $e_y$ differ in two coordinates for $x\ne y$. Write $a=\mu_0$ and $b_x=\mu_{e_x}$, so that $\sum_xb_x=1-a$.
Let $\lambda\ge1$. Since $f_K(a,b)\le\min\{a+b,Kb\}=b+\min\{a,\lambda b\}$, we get $\Phi_\lambda\le(1-a)+\sum_x\min\{a,\lambda b_x\}\le(1-a)+\min\{ta,\lambda(1-a)\}$. If $a\le\lambda/(t+\lambda)$, then $ta\le\lambda(1-a)$ and the bound is $1+(t-1)a\le tK/(t+\lambda)$; otherwise it is at most $K(1-a)\le tK/(t+\lambda)$. Conversely, $a=\lambda/(t+\lambda)$ and $b_x=1/(t+\lambda)$ satisfy $a\ge b_x$, and Lemma~\ref{lem:fK} gives $f_K(a,b_x)=b_x+\min\{a,\lambda b_x\}=K/(t+\lambda)$ on each of the $t$ edges.
Let $\lambda\le1$. Then $f_K=K\min$ (Lemma~\ref{lem:fK}), so $\Phi_\lambda=K\sum_x\min\{a,b_x\}\le K\min\{ta,1-a\}\le Kt/(t+1)$, and the uniform measure attains $Kt/(t+1)$.
For the properties let $c=\max\{1,\lambda\}\in[1,K]$; $t\mapsto t/(t+c)$ is increasing, $\sigma_\lambda(t)\le K$, and $\sigma_\lambda(t)\le t$ as $K\le t+c$. If $t\le K$, then $\sigma_\lambda(t)\ge tK/(K+c)\ge t/2$; if $t\ge K$, then $\sigma_\lambda(t)\ge tK/(2t)=K/2$. For $\lambda\ge1$, $\sigma_\lambda(1)=K/(1+\lambda)=1$.
\end{proof}

\begin{lemma}[Cubes and regular graphs]\label{lem:cube}
Let $\lambda\in[0,\infty)$. If $G(V)$ is regular of degree $g$, then $\tau_\lambda(V)=g\min\{1,K/2\}$. In particular $\tau_\lambda(Q(T))=|T|\min\{1,K/2\}$ for every finite nonempty $T$.
\end{lemma}

\begin{proof}
$G(V)$ has $g|V|/2$ edges, and the uniform measure gives each of them $f_K(1/|V|,1/|V|)=\min\{2,K\}/|V|$, in total $g\min\{1,K/2\}$. Conversely, if $\lambda\ge1$, then $\tau_\lambda(V)\le\Delta(V)=g$ by Lemma~\ref{lem:basic}. If $\lambda\le1$, then $f_K(a,b)=K\min\{a,b\}\le\frac K2(a+b)$ by Lemma~\ref{lem:fK}, so $\Phi_\lambda(\mu;V)\le\frac K2\sum_{\{u,w\}\in E(V)}(\mu_u+\mu_w)=\frac K2\sum_{v\in V}g\mu_v=\frac K2g$ for every $\mu\in\Pc(V)$. Every vertex of $Q(T)$ has degree $|T|$.
\end{proof}

\paragraph{Star products.} Let $T_1,\dots,T_r$ be nonempty, pairwise disjoint, finite subsets of $\X$ with union $U$, and let $c\in\{0,1\}^U$. A \emph{partial transversal} is a set $Y\subseteq U$ with $|Y\cap T_j|\le1$ for every $j$, and $\Hc$ \emph{star-shatters} $(c;T_1,\dots,T_r)$ if $c\oplus\one_Y\in\Hc|_U$ for every partial transversal $Y$.

\begin{lemma}[Star products]\label{lem:starprod}
\begin{enumerate}
\item[(i)] Let $\Hc$ star-shatter $(c;T_1,\dots,T_r)$ with union $U$. Then $F_\lambda(\Hc,|U|)\ge\tau_\lambda(\Hc|_U)\ge\sum_{j=1}^r\sigma_\lambda(|T_j|)$ for every $\lambda\in[0,\infty)$. For $1\le k\le r$, the partial transversals with $k-1$ elements and those with $k$ elements form a panel system of rank $k$ on $U$ with $a\ge r-k+1$ and $b=k$, and $\Hc$ shatters it with center $c$.
\item[(ii)] Let $\lambda\ge1$, let $r,u\ge1$ be integers, $r'=\min\{r,u\}$, and let $t_1,\dots,t_{r'}\ge1$ be integers with $\sum_jt_j=u$ and $\max_jt_j\le\lceil u/r'\rceil$. Then $\sum_j\sigma_\lambda(t_j)\ge\min\{u,rK\}/2$. Such integers exist: if $u=qr'+s$ with $0\le s<r'$, take $s$ of them equal to $q+1$ and the others equal to $q\ge1$.
\end{enumerate}
\end{lemma}

\begin{proof}
(i) The set $P=\{c\oplus\one_Y:Y\text{ a partial transversal}\}$ lies in $\Hc|_U$, and $v\mapsto v\oplus c$ maps it isomorphically onto $P_0=\{\one_Y:Y\text{ a partial transversal}\}$ (Lemma~\ref{lem:Dinv}(i)). Under $\{0,1\}^U=\prod_j\{0,1\}^{T_j}$, the restriction of $\one_Y$ to $T_j$ is $0$ or $e_y$ with $y\in T_j$, and every choice of one such pattern for each block is the restriction of a partial transversal; hence $P_0=\prod_jS(T_j)$. Lemmas~\ref{lem:product} (by induction on $r$) and~\ref{lem:star} give $\tau_\lambda(P_0)=\sum_j\sigma_\lambda(|T_j|)$, and Lemma~\ref{lem:Dinv}(ii) gives $\tau_\lambda(\Hc|_U)\ge\tau_\lambda(P)$. For the panel system let $\Ac$ and $\Wc$ be the families of partial transversals with $k-1$ and $k$ elements, both nonempty as $k\le r$. A set $S\in\Ac$ misses $r-k+1$ blocks, and $S\cup\{y\}\in\Wc$ for every $y$ in them, so $a\ge r-k+1\ge1$. Each $T\in\Wc$ has exactly $k$ subsets with $k-1$ elements, all partial transversals, so $b=k$. By definition $c\oplus\one_S\in\Hc|_U$ for all $S\in\Ac\cup\Wc$.

(ii) If $r'=u$, all $t_j=1$ and the sum is $u\,\sigma_\lambda(1)=u$ by Lemma~\ref{lem:star}. Otherwise $r'=r<u$. For $\lambda\ge1$ and $t\le t_{\max}:=\max_jt_j<u/r+1$ we have $\sigma_\lambda(t)=tK/(t+\lambda)\ge tK/(t_{\max}+\lambda)$, so $\sum_j\sigma_\lambda(t_j)\ge uK/(u/r+1+\lambda)=uK/(u/r+K)$. This is at least $uK/(2K)=u/2$ if $u/r\le K$, and at least $uK/(2u/r)=rK/2$ otherwise.
\end{proof}

\begin{lemma}[A priori bounds]\label{lem:apriori}
\begin{enumerate}
\item[(i)] For every $\lambda\in[0,\infty)$, $\tau_\lambda(V)\le Kd_V$. Hence $F_\lambda(\Hc,m)\le\min\{Kd_m,\Delta_m(\Hc)\}$.
\item[(ii)] For $\lambda\ge1$, $F_\lambda(\Hc,m)\ge\max\{d_m,\sigma_\lambda(\Delta_m(\Hc))\}\ge\max\{d_m,\min\{\Delta_m(\Hc),K\}/2\}$.
\end{enumerate}
\end{lemma}

\begin{proof}
(i) Since $f_K(a,b)\le K\min\{a,b\}=Kf_1(a,b)$ (Lemma~\ref{lem:fK}), $\Phi_\lambda(\mu;V)\le K\Phi_0(\mu;V)\le K\tau_0(V)$ for every $\mu\in\Pc(V)$. Regard $V$ as a class on the finite domain $U$; it has VC dimension $d_V$, $V|_U=V$, and Theorem~\ref{thm:binary}(a) at $\lambda=0$ gives $\tau_0(V)\le F_0(V,|U|)\le d_V$. For $|U|\le m$, every set shattered by $\Hc|_U$ is shattered by $\Hc$ and has at most $m$ points, so $d_{\Hc|_U}\le d_m$, and $\tau_\lambda(\Hc|_U)\le\Delta(\Hc|_U)\le\Delta_m(\Hc)$ by Lemma~\ref{lem:basic}.
(ii) If $d_m\ge1$ and $T$ is a shattered set with $d_m$ points, then $\Hc|_T=Q(T)$ and $F_\lambda(\Hc,m)\ge d_m$ by Lemma~\ref{lem:cube}. If $\Delta_m:=\Delta_m(\Hc)\ge1$, take $|U|\le m$ and $c\in\Hc|_U$ with neighbors $c\oplus e_x$, $x\in T$, $|T|=\Delta_m$. Translation by $c$ maps $\{c\}\cup\{c\oplus e_x:x\in T\}$ onto $\{0\}\cup\{e_x:x\in T\}$, whose elements agree outside $T$, so by Lemma~\ref{lem:Dinv}(i) it is isomorphic to $S(T)$, and Lemmas~\ref{lem:Dinv}(ii) and~\ref{lem:star} give $\tau_\lambda(\Hc|_U)\ge\sigma_\lambda(\Delta_m)\ge\min\{\Delta_m,K\}/2$.
\end{proof}

\begin{lemma}[Hamming balls]\label{lem:hball}
For a finite set $U$ and $k\ge1$ let $B(U,k):=\{\one_A:A\subseteq U,\ |A|\le k\}$. If $\lambda\ge1$ and $|U|=u\ge1$, then $\frac12\min\{u,kK\}\le\tau_\lambda(B(U,k))\le\min\{u,kK\}$.
\end{lemma}

\begin{proof}
$\tau_\lambda(B(U,k))\le u$ by Lemma~\ref{lem:basic}. $B(U,k)$ shatters exactly the subsets of $U$ with at most $k$ elements, so its VC dimension is $\min\{k,u\}$ and Lemma~\ref{lem:apriori}(i) gives $\tau_\lambda(B(U,k))\le kK$. For the lower bound partition $U$ into $r'=\min\{k,u\}$ blocks $T_1,\dots,T_{r'}$ with sizes as in Lemma~\ref{lem:starprod}(ii) for $r=k$. A partial transversal has at most $r'\le k$ elements, so $B(U,k)$, as a class on $U$, star-shatters $(0;T_1,\dots,T_{r'})$, and Lemma~\ref{lem:starprod} gives $\tau_\lambda(B(U,k))\ge\sum_j\sigma_\lambda(|T_j|)\ge\frac12\min\{u,kK\}$.
\end{proof}

\noindent The next theorem restates Lemma~\ref{lem:entropy} with its consequence for classes and gives the details of its proof.

\begin{theorem}[Entropy bound]\label{thm:entropy}
Let $\lambda\ge1$. Then
\[
\tau_\lambda(V)\le\frac{\log_2|V|}{h(1/K)}\le\frac{K\log_2|V|}{\log_2K},
\]
and consequently $F_\lambda(\Hc,m)\le\log_2|\Hc|/h(1/K)$ for every finite class $\Hc$ and every $m\ge1$.
\end{theorem}

\begin{proof}
We use three facts on the Shannon entropy $H(\cdot)$, in bits, of random variables with finitely many values \citep[Theorems~2.5.1, 2.6.4 and~2.6.5]{CT06}: the chain rule, $H(Y)\le\log_2|\supp(Y)|$, and $H(Y\mid Z,Z')\le H(Y\mid Z)$. Fix $\mu\in\Pc(V)$, let $Z=(Z_x)_{x\in U}$ be a random element of $V$ with law $\mu$, and write $Z_{-x}=(Z_y)_{y\ne x}$.

\emph{Step 1.} Fix $x\in U$ and a value $y$ of $Z_{-x}$ of positive probability. The fiber $\{v\in V:v|_{U\setminus\{x\}}=y\}$ is a single pattern, and then $H(Z_x\mid Z_{-x}=y)=0$, or an edge $\{u,w\}$ of $G(V)$ in direction $x$, and then the event has probability $\mu_u+\mu_w$ and $Z_x$ takes its two values with conditional probabilities $\mu_u/(\mu_u+\mu_w)$ and $\mu_w/(\mu_u+\mu_w)$. Every edge in direction $x$ is such a fiber. Let $E^+$ be the set of edges with $\mu_u+\mu_w>0$ and $q_{uw}=\min\{\mu_u,\mu_w\}/(\mu_u+\mu_w)$. Since $h$ is symmetric about $1/2$, $\sum_{x\in U}H(Z_x\mid Z_{-x})=\sum_{\{u,w\}\in E^+}(\mu_u+\mu_w)\,h(q_{uw})$.

\emph{Step 2 (Han's inequality).} Enumerate $U=\{x_1,\dots,x_k\}$. Since $(Z_{x_1},\dots,Z_{x_{i-1}})$ is a subvector of $Z_{-x_i}$, conditioning gives $H(Z_{x_i}\mid Z_{-x_i})\le H(Z_{x_i}\mid Z_{x_1},\dots,Z_{x_{i-1}})$, and the chain rule gives
\[
\sum_{x\in U}H(Z_x\mid Z_{-x})\le\sum_{i=1}^kH(Z_{x_i}\mid Z_{x_1},\dots,Z_{x_{i-1}})=H(Z)\le\log_2|V| .
\]
This is the form of Han's inequality \citep{Han78} used in Section~\ref{sec:ov-panels}.

\emph{Step 3.} Let $\{u,w\}\in E^+$ and $q=q_{uw}\in[0,1/2]$. Then $f_K(\mu_u,\mu_w)=\min\{\mu_u+\mu_w,K\min\{\mu_u,\mu_w\}\}=(\mu_u+\mu_w)\min\{1,Kq\}$, and $h(1/K)>0$ as $0<1/K\le1/2$. If $q\ge1/K$, then $h(q)\ge h(1/K)$ because $h$ is increasing on $[0,1/2]$, so $\min\{1,Kq\}\le1\le h(q)/h(1/K)$. If $q<1/K$, concavity of $h$ and $h(0)=0$ give $h(q)=h\big(Kq\cdot\frac1K+(1-Kq)\cdot0\big)\ge Kq\,h(1/K)$, so $\min\{1,Kq\}\le Kq\le h(q)/h(1/K)$. Edges outside $E^+$ contribute $f_K(0,0)=0$. With Steps 1 and 2, $\Phi_\lambda(\mu;V)\le\log_2|V|/h(1/K)$, and we maximize over $\mu$. Finally $h(1/K)=\frac1K\log_2K+(1-\frac1K)\log_2\frac K{K-1}\ge\frac1K\log_2K$. For the consequence, $|\Hc|_U|\le|\Hc|$ for every $U$.
\end{proof}

\subsection{Saturating classes}\label{sec:D-sat}

\begin{proposition}[Threshold functions]\label{prop:Dthresholds}
Let $\X\subseteq\R$ be countable with $|\X|\ge2$, for instance $\X=\Q$, and $\Hc_{\rm thr}=\{x\mapsto\one[x\ge\theta]:\theta\in\R\}$. Then $\Hc_{\rm thr}$ has VC dimension $1$, and $\min\{1,K/2\}\le F_\lambda(\Hc_{\rm thr},m)\le\min\{2,K\}$ for all $\lambda\in[0,\infty)$ and $m\ge1$.
\end{proposition}

\begin{proof}
For $U=\{x_1<\dots<x_u\}$ and $\theta\in(x_j,x_{j+1}]$, with $x_0=-\infty$ and $x_{u+1}=+\infty$, the restriction of $\one[x\ge\theta]$ to $U$ is $0^j1^{u-j}$. Two of these $u+1$ patterns are adjacent if and only if their indices $j$ differ by $1$, so $G(\Hc_{\rm thr}|_U)$ is a path and $\Delta(\Hc_{\rm thr}|_U)\le2$. Every point carries both labels and the pattern $(1,0)$ on two points is not realized, so $d=1$, and Lemma~\ref{lem:apriori}(i) gives $F_\lambda\le\min\{K,2\}$. On one point the projection is a cube, so $F_\lambda(\Hc_{\rm thr},m)\ge F_\lambda(\Hc_{\rm thr},1)=\min\{1,K/2\}$ by Lemmas~\ref{lem:cube} and~\ref{lem:basic}.
\end{proof}

\begin{proposition}[Parities]\label{prop:Dparities}
Let $D\ge1$, $\X=\{0,1\}^D$ and $\Hc_{\rm par}=\{x\mapsto\langle a,x\rangle\bmod2:a\in\{0,1\}^D\}$. Then $\Hc_{\rm par}$ has VC dimension $D$ and star number $D$, and $F_\lambda(\Hc_{\rm par},m)=\min\{1,K/2\}\min\{m,D\}$ for all $\lambda\in[0,\infty)$ and $m\ge1$.
\end{proposition}

\begin{proof}
For finite $U\subseteq\{0,1\}^D$ the map $a\mapsto(\langle a,x\rangle)_{x\in U}$ from $\GF^D$ to $\GF^U$ is linear, and its image $W_U=\Hc_{\rm par}|_U$ is a subspace of dimension at most $\min\{|U|,D\}$. For $v\in W_U$ and $x\in U$, $v\oplus e_x\in W_U$ if and only if $e_x\in W_U$. Hence $G(W_U)$ is regular of degree $g_U=|\{x\in U:e_x\in W_U\}|$, and since the vectors $e_x\in W_U$ are linearly independent, $g_U\le\dim W_U\le\min\{|U|,D\}$. Lemma~\ref{lem:cube} gives $\tau_\lambda(W_U)=g_U\min\{1,K/2\}\le\min\{|U|,D\}\min\{1,K/2\}$. For a set $U$ of $\min\{m,D\}$ unit vectors, $\langle a,e_i\rangle=a_i$ gives $W_U=\{0,1\}^U$ and $g_U=|U|$. A shattered set has at most $\dim W_U\le D$ points, so $d=D$, and the star number is $\max_Ug_U=D$.
\end{proof}

\begin{proposition}[Full tables]\label{prop:Dfulltable}
Let $|\X|=N\in\{1,2,\dots\}\cup\{\infty\}$ and $\Hc=\{0,1\}^{\X}$. Then $F_\lambda(\Hc,m)=\min\{1,K/2\}\min\{m,N\}$ for all $\lambda\in[0,\infty)$ and $m\ge1$.
\end{proposition}

\begin{proof}
Every projection $\Hc|_U$ is the cube $Q(U)$, and $|U|\le\min\{m,N\}$ with equality for some $U$. Apply Lemma~\ref{lem:cube}.
\end{proof}

\subsection{Classes of linear growth in $K$}\label{sec:D-lin}

Throughout this subsection $\lambda\ge1$ and $m\ge1$. Every upper bound below is Lemma~\ref{lem:apriori}(i) with the stated VC dimension, combined with $F_\lambda(\Hc,m)\le m$ (Lemma~\ref{lem:basic}).

\begin{proposition}[Sparse sets]\label{prop:Dsparse}
Let $|\X|=N\in\{1,2,\dots\}\cup\{\infty\}$, $k\ge1$ and $\Sc_k=\{\one_A:A\subseteq\X,\ |A|\le k\}$. Then $\Sc_k$ has VC dimension $\min\{k,N\}$ and star number $N$, and $\frac12\min\{m,N,kK\}\le F_\lambda(\Sc_k,m)\le\min\{m,N,kK\}$.
\end{proposition}

\begin{proof}
For finite $U$, $\Sc_k|_U=B(U,k)$. A set $T$ is shattered if and only if $|T|\le k$, so $d=\min\{k,N\}$. In $B(U,k)$ the vertex $0$ has degree $|U|$ and every degree is at most $|U|$, so $s=N$. Every $U$ with $|U|\le m$ has $|U|\le\min\{m,N\}$, and Lemma~\ref{lem:hball} gives the upper bound for all such $U$ and the lower bound for $|U|=\min\{m,N\}$.
\end{proof}

\begin{proposition}[Unions of intervals]\label{prop:Dintervals}
Let $\X\subseteq\R$ be countably infinite, $k\ge1$ and $\mathcal I_k=\{\one[x\in I_1\cup\dots\cup I_j]:0\le j\le k,\ I_i\subseteq\R\text{ intervals}\}$. Then $\mathcal I_k$ has VC dimension $2k$, and $\frac12\min\{m,kK\}\le F_\lambda(\mathcal I_k,m)\le\min\{m,2kK\}$.
\end{proposition}

\begin{proof}
Let $U=\{x_1<\dots<x_u\}$. An interval meets $U$ in the empty set or in a run of consecutive points, so a union of $j\le k$ intervals has at most $k$ maximal runs of ones on $U$, and the pattern with maximal runs $\{x_{a_i},\dots,x_{b_i}\}$, $i\le j$, is the restriction of $\bigcup_i[x_{a_i},x_{b_i}]$; so $\mathcal I_k|_U$ is the set of patterns with at most $k$ maximal runs of ones. A pattern with $r$ runs has length at least $2r-1$, so all patterns on $2k$ points are realized, while $1010\cdots1$ on $2k+1$ points has $k+1$ runs; thus $d=2k$. Every set of at most $k$ points has at most $k$ runs, so $\mathcal I_k|_U\supseteq B(U,k)$, and Lemmas~\ref{lem:Dinv}(ii) and~\ref{lem:hball} with $|U|=m$ give the lower bound.
\end{proof}

\paragraph{Halfspaces.} Let $\X=\Q^d$, $\mathcal{HS}_d=\{x\mapsto\one[\langle w,x\rangle+b\ge0]:w\in\R^d,\ b\in\R\}$ and $\mathcal{HHS}_d=\{x\mapsto\one[\langle w,x\rangle\ge0]:w\in\R^d\}$. Let $\gamma(s)=(s,s^2,\dots,s^d)$ be the moment curve.

\begin{lemma}[Moment curve]\label{lem:moment}
Let $d\ge2$, $j=\lfloor d/2\rfloor$, and $U=\{\gamma(s_1),\dots,\gamma(s_u)\}$ for rationals $s_1<\dots<s_u$. Then $\mathcal{HS}_d|_U\supseteq B(U,j)$.
\end{lemma}

\begin{proof}
The points are distinct because their first coordinates are. Let $A=\{\gamma(s'_1),\dots,\gamma(s'_a)\}\subseteq U$ with $a\le j$, and $p(s)=-\prod_{i=1}^a(s-s'_i)^2=\sum_{l=0}^{2a}c_ls^l$, with $p\equiv-1$ for $a=0$. Since $2a\le d$, the choice $b=c_0$, $w_l=c_l$ for $1\le l\le2a$ and $w_l=0$ for $l>2a$ gives $\langle w,\gamma(s)\rangle+b=p(s)$. As $p$ vanishes at $s'_1,\dots,s'_a$ and is negative elsewhere, this halfspace labels exactly the points of $A$ by $1$. This is the $j$-neighborliness of cyclic polytopes \citep{Gal63}.
\end{proof}

\begin{proposition}[Affine halfspaces]\label{prop:Dhalfspaces}
Let $d\ge2$ and $j=\lfloor d/2\rfloor$. Then $\mathcal{HS}_d$ has VC dimension at most $d+1$, and
\[
\tfrac18\min\{m,(d+1)K\}\le\tfrac12\min\{m,jK\}\le F_\lambda(\mathcal{HS}_d,m)\le\min\{m,(d+1)K\}.
\]
If $U$ is a set of $m\ge jK$ points $\gamma(s)$ with rational $s$, the $(j-1)$-subsets and the $j$-subsets of $U$ form a panel system of rank $j$ with $a=m-j+1$ and $b=j$, which $\mathcal{HS}_d$ shatters with center $0$, and its value is $\lambda j$.
\end{proposition}

\begin{proof}
Let $P\subseteq\Q^d$ have $d+2$ points. By Radon's theorem \citep{Rad21} there is a partition $P=P_1\sqcup P_2$ and a point $z\in\conv(P_1)\cap\conv(P_2)$. No $h\in\mathcal{HS}_d$ labels $P_1$ by $1$ and $P_2$ by $0$, since the convex sets $\{\langle w,x\rangle+b\ge0\}$ and $\{\langle w,x\rangle+b<0\}$ would both contain $z$; so $d_{\mathcal{HS}_d}\le d+1$. For the lower bound take $m$ points $\gamma(s)$ with rational $s$; Lemmas~\ref{lem:moment}, \ref{lem:Dinv}(ii) and~\ref{lem:hball} give $F_\lambda\ge\frac12\min\{m,jK\}$. Since $j\ge(d+1)/4$ for $d\ge2$ (for $d=2$ directly, and $j\ge(d-1)/2\ge(d+1)/4$ for $d\ge3$), $\min\{m,jK\}\ge\frac14\min\{m,(d+1)K\}$.
By Lemma~\ref{lem:moment}, $\one_S\in\mathcal{HS}_d|_U$ for every $S\subseteq U$ with $|S|\le j$. Each $(j-1)$-subset of $U$ lies in $m-j+1$ of the $j$-subsets and each $j$-subset contains $j$ of the $(j-1)$-subsets, and for $m\ge jK$ we have $m-j+1>\lambda j$, so the value is $\lambda j$; Theorem~\ref{thm:binary}(b) then gives again $F_\lambda(\mathcal{HS}_d,m)\ge\frac K{2\lambda}\lambda j=jK/2$.
\end{proof}

\begin{proposition}[Homogeneous halfspaces]\label{prop:Dhomogeneous}
Let $d\ge3$. Then $\mathcal{HHS}_d$ has VC dimension at most $d$, and
\[
\tfrac1{10}\min\{m,dK\}\le\tfrac12\min\{m,\lfloor(d-1)/2\rfloor K\}\le F_\lambda(\mathcal{HHS}_d,m)\le\min\{m,dK\}.
\]
\end{proposition}

\begin{proof}
Points $x_1,\dots,x_{d+1}\in\R^d$ satisfy $\sum_i\alpha_ix_i=0$ for some $\alpha\ne0$, where we may take $I=\{i:\alpha_i>0\}$ nonempty. If some $w$ had $\langle w,x_i\rangle<0$ for $i\in I$ and $\langle w,x_i\rangle\ge0$ for $i\notin I$, then $0=\sum_{i\in I}\alpha_i\langle w,x_i\rangle+\sum_{i\notin I}\alpha_i\langle w,x_i\rangle<0$. So the labeling that is $0$ on $I$ and $1$ elsewhere is not realized, and $d_{\mathcal{HHS}_d}\le d$. For the lower bound let $U'\subseteq\Q^{d-1}$ and $U=\{(z,1):z\in U'\}\subseteq\Q^d$. Since $\langle(w',b),(z,1)\rangle=\langle w',z\rangle+b$, $\mathcal{HHS}_d|_U$ coincides with $\mathcal{HS}_{d-1}|_{U'}$ under $(z,1)\mapsto z$. For the $m$ points $U'$ used in the proof of Proposition~\ref{prop:Dhalfspaces} in dimension $d-1\ge2$, Lemma~\ref{lem:Dinv}(i) gives $F_\lambda(\mathcal{HHS}_d,m)\ge\tau_\lambda(\mathcal{HHS}_d|_U)=\tau_\lambda(\mathcal{HS}_{d-1}|_{U'})\ge\frac12\min\{m,\lfloor(d-1)/2\rfloor K\}$. Finally $\lfloor(d-1)/2\rfloor\ge d/5$ for $d\ge3$ (for $d=3$ directly, and $\lfloor(d-1)/2\rfloor\ge(d-2)/2\ge d/5$ for $d\ge4$).
\end{proof}

\paragraph{Boxes.} Let $\X=\Q^d$ and $\mathcal B_d=\{\one_{\prod_i[a_i,b_i]}:a_i\le b_i\}\cup\{0\}$, where $0$ is the empty box.

\begin{lemma}[Box blocks]\label{lem:boxblocks}
Let $d\ge2$, $1\le r\le\lfloor d/2\rfloor$, let $t_1,\dots,t_r\ge1$ be integers, $t=\max_lt_l$ and $R=t+1$. For $l\in[r]$ and $i\in[t_l]$ let $a_{l,i}=(R+i)e_{2l-1}+(R-i)e_{2l}\in\Q^d$ and $T_l=\{a_{l,i}:i\in[t_l]\}$. Then $\mathcal B_d$ star-shatters $(0;T_1,\dots,T_r)$.
\end{lemma}

\begin{proof}
The points are distinct, and the support of $a_{l,i}$ is $\{2l-1,2l\}$ since $R-i\ge1$. Let $Y$ be a partial transversal; for $Y=\emptyset$ take the empty box. Otherwise let $\mathrm{bb}(Y)=\prod_k[\min_{y\in Y}y_k,\max_{y\in Y}y_k]$, a box that contains $Y$, and let $z=a_{l',i'}\notin Y$. If $Y\cap T_{l'}=\emptyset$, every $y\in Y$ has $y_{2l'-1}=0$, while $z_{2l'-1}=R+i'>0$. If $Y\cap T_{l'}=\{a_{l',i}\}$ with $i\ne i'$, the coordinates $2l'-1$ and $2l'$ of the elements of $Y$ take only the values $\{0,R+i\}$ and $\{0,R-i\}$, so $\mathrm{bb}(Y)$ has upper ends $R+i$ and $R-i$ there; if $i'>i$ then $z_{2l'-1}=R+i'>R+i$, and if $i'<i$ then $z_{2l'}=R-i'>R-i$. In all cases $z\notin\mathrm{bb}(Y)$, so $\one_{\mathrm{bb}(Y)}|_U=\one_Y$.
\end{proof}

\begin{proposition}[Boxes]\label{prop:Dboxes}
Let $d\ge1$. Then $\mathcal B_d$ has VC dimension at most $2d$, and $\frac1{12}\min\{m,2dK\}\le F_\lambda(\mathcal B_d,m)\le\min\{m,2dK\}$.
\end{proposition}

\begin{proof}
Let $P\subseteq\Q^d$ have $2d+1$ points. For each coordinate $i$ pick points of $P$ that minimize and maximize the $i$-th coordinate over $P$, and let $S$ be the set of these at most $2d$ points. Some $z\in P\setminus S$ exists, and every box that contains $S$ contains $\prod_i[\min_{x\in P}x_i,\max_{x\in P}x_i]\ni z$; so $\one_S$ is not realized on $P$, and $d_{\mathcal B_d}\le2d$. For the lower bound, singletons are boxes, so $\mathcal B_d|_U\supseteq B(U,1)$ and Lemma~\ref{lem:hball} with $|U|=m$ gives $F_\lambda\ge\frac12\min\{m,K\}$. For $d\ge2$, Lemma~\ref{lem:boxblocks} with $\min\{\lfloor d/2\rfloor,m\}$ blocks of sizes as in Lemma~\ref{lem:starprod}(ii), for $u=m$ and $r=\lfloor d/2\rfloor$, and Lemma~\ref{lem:starprod}(i) give $F_\lambda\ge\frac12\min\{m,\lfloor d/2\rfloor K\}$. Hence $F_\lambda\ge\frac12\min\{m,j_dK\}$ with $j_d=\max\{\lfloor d/2\rfloor,1\}\ge d/3$ (for $d\le3$ directly, and $\lfloor d/2\rfloor\ge(d-1)/2\ge d/3$ for $d\ge4$), so $F_\lambda\ge\frac12\min\{m,2dK/6\}\ge\frac1{12}\min\{m,2dK\}$.
\end{proof}

\subsection{Classes of growth $K/\log K$}\label{sec:D-log}

Throughout this subsection $\lambda\ge1$, $m\ge1$, and $\Lambda:=\lceil\lambda\rceil+1$.

\paragraph{Dictators.} Let $p\ge1$, $\X=\{0,1\}^p$ and $\mathcal D_p=\{h_0\}\cup\{h_i:i\in[p]\}$ with $h_0\equiv0$ and $h_i(x)=x_i$.

\begin{lemma}[Realization by dictators]\label{lem:Drealize}
Let $V\subseteq\{0,1\}^r$ have pairwise distinct coordinates, that is, for $k\ne k'$ some $v\in V$ has $v(k)\ne v(k')$, and let $|V\setminus\{0\}|\le p$. Then there are distinct $z_1,\dots,z_r\in\{0,1\}^p$ such that, with $U=\{z_1,\dots,z_r\}$ and the bijection $k\mapsto z_k$, $\mathcal D_p|_U\supseteq V$. In particular $F_\lambda(\mathcal D_p,m)\ge\tau_\lambda(V)$ for $r\le m$.
\end{lemma}

\begin{proof}
Enumerate $V\setminus\{0\}=\{v^1,\dots,v^q\}$ and let $(z_k)_i=v^i(k)$ for $i\le q$ and $(z_k)_i=0$ for $q<i\le p$. If $z_k=z_{k'}$, then $v(k)=v(k')$ for all $v\in V$, so $k=k'$. Under $k\mapsto z_k$, $h_i|_U=v^i$ for $i\le q$ and $h_0|_U=0$; apply Lemma~\ref{lem:Dinv}(i),(ii).
\end{proof}

\begin{lemma}[Hamming layers]\label{lem:Dlayers}
Let $1\le j\le r-1$ and $V_{r,j}=\{\one_S:S\subseteq[r],\ |S|\in\{j-1,j\}\}$. Then $V_{r,j}$ has pairwise distinct coordinates, the $(j-1)$-subsets and the $j$-subsets of $[r]$ form a panel system of rank $j$ with $a=r-j+1$ and $b=j$, which $V_{r,j}$ shatters with center $0$, and if $j\le r/2$, then $|V_{r,j}|\le2\binom rj\le2(er/j)^j$.
\end{lemma}

\begin{proof}
For $k\ne k'$, since $j-1\le r-2$, some $j$-subset contains $k$ and not $k'$. Each $(j-1)$-subset lies in $r-j+1$ of the $j$-subsets and each $j$-subset contains $j$ of the $(j-1)$-subsets, and $\one_S=0\oplus\one_S\in V_{r,j}$ for all of them. For $j\le r/2$, $\binom r{j-1}\le\binom rj\le r^j/j!\le(er/j)^j$, using $j!\ge(j/e)^j$.
\end{proof}

\begin{proposition}[Dictators]\label{prop:Ddictators}
\begin{enumerate}
\item[(i)] $|\mathcal D_p|=p+1$, and $\mathcal D_p$ has VC dimension $\lfloor\log_2(p+1)\rfloor$ and star number $p$.
\item[(ii)] If $p\ge2$ and $B:=K\log_2p/\log_2K$, then $\frac1{33}\min\{m,p,B\}\le F_\lambda(\mathcal D_p,m)\le\min\{m,p,2B\}$.
\end{enumerate}
\end{proposition}

\begin{proof}
(i) Distinct members of $\mathcal D_p$ differ at some unit vector $e_i$. A shattered set of size $d$ needs $2^d\le p+1$ functions. For $t=\lfloor\log_2(p+1)\rfloor\ge1$, $Q([t])$ has pairwise distinct coordinates and $2^t-1\le p$ nonzero patterns, so Lemma~\ref{lem:Drealize} gives a set of $t$ points shattered by $\mathcal D_p$. On $U=\{e_1,\dots,e_p\}$ the projection is the star $S(U)$, whose center has degree $p$, and every projection has at most $p+1$ patterns, hence degrees at most $p$; so $s=p$.

(ii) \emph{Upper bound.} $F_\lambda\le m$, $F_\lambda\le\Delta_m(\mathcal D_p)\le p$ by Lemma~\ref{lem:apriori}(i) and part (i), and $F_\lambda\le\log_2(p+1)/h(1/K)$ by Theorem~\ref{thm:entropy}. For $p\ge2$, $\log_2(p+1)\le\log_2p+\log_2\frac32\le1.585\log_2p$, and $h(1/K)\ge\log_2K/K$, so $\log_2(p+1)/h(1/K)\le2B$.

\emph{Lower bound.} Let $t^\ast=\min\{m,p\}$. On $t^\ast$ unit vectors the projection is a star, so $F_\lambda\ge\sigma_\lambda(t^\ast)$ by Lemma~\ref{lem:star}. Let $j^\ast=\max(\{0\}\cup\{j\in\N,\ j\ge1:\Lambda j\le m,\ 2(e\Lambda)^j\le p\})$, suppose $j^\ast\ge1$, and let $r=\Lambda j^\ast\le m$. As $\Lambda\ge2$, $j^\ast\le r/2\le r-1$, and $|V_{r,j^\ast}|\le2(e\Lambda)^{j^\ast}\le p$ by Lemma~\ref{lem:Dlayers}. By Lemmas~\ref{lem:Drealize} and~\ref{lem:Dlayers}, $\mathcal D_p$ shatters with center $0$, on $r\le m$ points, a panel system with $a=\lceil\lambda\rceil j^\ast+1>\lambda j^\ast$ and $b=j^\ast$, of value $\lambda j^\ast$, and Theorem~\ref{thm:binary}(b) gives $F_\lambda(\mathcal D_p,m)\ge\frac K{2\lambda}\lambda j^\ast=Kj^\ast/2$.
Note that $\Lambda\le\lambda+2=K+1\le\frac32K$. \emph{Case $t^\ast\le3eK$.} By Lemma~\ref{lem:star}, $\sigma_\lambda(t^\ast)\ge\min\{t^\ast,K\}/2\ge t^\ast/(6e)\ge\frac1{33}\min\{m,p,B\}$. \emph{Case $m>3eK$ and $p>3eK$.} Then $m/\Lambda>2e$ and $p/2>\frac32eK\ge e\Lambda$, so $j_m:=\lfloor m/\Lambda\rfloor\ge1$, $j_p:=\lfloor\log_2(p/2)/\log_2(e\Lambda)\rfloor\ge1$ and $j^\ast=\min\{j_m,j_p\}$. We use $\lfloor x\rfloor\ge x/2$ for $x\ge1$. If $j^\ast=j_m$, then $Kj^\ast/2\ge Km/(4\Lambda)\ge m/6$. If $j^\ast=j_p$, then $Kj^\ast/2\ge K\log_2(p/2)/(4\log_2(e\Lambda))$. Here $\log_2(e\Lambda)\le\log_2K+\log_2\frac{3e}2\le3.03\log_2K$ as $\log_2K\ge1$, and $p>6e$ gives $\log_2p>4.02$, hence $\log_2(p/2)\ge0.75\log_2p$; so $Kj^\ast/2\ge\frac{0.75}{4\cdot3.03}B\ge B/17$. In both cases $F_\lambda\ge\min\{m/6,B/17\}\ge\frac1{33}\min\{m,p,B\}$.
\end{proof}

\begin{corollary}[Patterns of a shattered panel system]\label{cor:Dpatterns}
Let $\lambda\ge1$ and $j\ge1$. If $\Hc|_U$ shatters a panel system on $U$ of value at least $\lambda j$, then $|\Hc|_U|\ge K^{j/2}$.
\end{corollary}

\begin{proof}
Regard $\Hc|_U$ as a class on the finite domain $U$. Then $\pi_\lambda(\Hc|_U,|U|)\ge\lambda j$, and Theorem~\ref{thm:binary}(b) gives $F_\lambda(\Hc|_U,|U|)\ge Kj/2$. Let $U'\subseteq U$ attain this value (Lemma~\ref{lem:basic}). Theorem~\ref{thm:entropy} gives $Kj/2\le\tau_\lambda(\Hc|_{U'})\le K\log_2|\Hc|_{U'}|/\log_2K$, so $|\Hc|_U|\ge|\Hc|_{U'}|\ge K^{j/2}$.
\end{proof}

\paragraph{Monotone conjunctions and disjunctions.} Let $D\ge2$, $\X=\{0,1\}^D$, $\mathcal{MC}_D=\{h_S:S\subseteq[D]\}$ with $h_S(x)=\prod_{i\in S}x_i$ ($h_\emptyset\equiv1$), and $\mathcal{MD}_D=\{x\mapsto1-\prod_{i\in S}(1-x_i):S\subseteq[D]\}$.

\begin{lemma}[Blocks for conjunctions]\label{lem:Cblocks}
Let $2\le g\le D$, $R=\lfloor D/g\rfloor$, $w=\lfloor g/2\rfloor$ and $G_l=\{(l-1)g+1,\dots,lg\}$ for $l\in[R]$. For $B\subseteq[D]$ let $z_B=\one_{[D]\setminus B}$. For $r\le R$ and $l\in[r]$ let $T_l\subseteq\{z_B:B\subseteq G_l,\ |B|=w\}$ be nonempty. Then $\mathcal{MC}_D$ star-shatters $(0;T_1,\dots,T_r)$. Moreover $\binom g{\lfloor g/2\rfloor}\ge2^{g/2}$ for every $g\ge2$.
\end{lemma}

\begin{proof}
Distinct sets $B$ give distinct points, and $h_S(z_B)=1$ if and only if $S\cap B=\emptyset$. Let $Y$ be a partial transversal and $S=[D]\setminus\bigcup_{z_B\in Y}B$. For $z_B\in Y$, $h_S(z_B)=1$. Let $z_{B'}\notin Y$ with $B'\subseteq G_{l'}$. Then $h_S(z_{B'})=1$ if and only if $B'\subseteq\bigcup_{z_B\in Y}B$, and since the groups are disjoint the right side meets $G_{l'}$ in $\emptyset$ or in a single set $B$ with $|B|=w$. In the first case $B'$ is not contained in it because $B'\ne\emptyset$; in the second, $B'\subseteq B$ and $|B'|=|B|$ would give $z_{B'}=z_B\in Y$. So $h_S|_U=\one_Y$. For the binomial bound, $\binom21=2$ and $\binom31\ge2^{3/2}$, and for $k=\lfloor g/2\rfloor$ Vandermonde's identity gives $\binom{g+2}{k+1}=\binom g{k-1}+2\binom gk+\binom g{k+1}\ge2\binom gk$; induct from $g$ to $g+2$.
\end{proof}

\begin{proposition}[Monotone conjunctions and disjunctions]\label{prop:Dconjunctions}
$|\mathcal{MC}_D|=2^D$, $\mathcal{MC}_D$ and $\mathcal{MD}_D$ have VC dimension $D$, $F_\lambda(\mathcal{MD}_D,m)=F_\lambda(\mathcal{MC}_D,m)$, and if $K\le2^{D/2}$, then
\[
\min\Big\{\frac m6,\frac{DK}{12\log_2K}\Big\}\le F_\lambda(\mathcal{MC}_D,m)\le\min\Big\{m,\frac D{h(1/K)}\Big\}\le\min\Big\{m,\frac{DK}{\log_2K}\Big\},
\]
where the left side is at least $\frac1{12}\min\{m,DK/\log_2K\}$.
\end{proposition}

\begin{proof}
\emph{Size and VC dimension.} $h_{S'}(\one_S)=1$ if and only if $S'\subseteq S$, so distinct $S$ give distinct functions. On the points $\one_{[D]\setminus\{i\}}$, $i\in[D]$, $h_S$ equals $1$ exactly where $i\notin S$, so these points are shattered, and $d\le\log_2|\mathcal{MC}_D|=D$. The bijection $\iota(x)=\one-x$ satisfies $1-h_S(\iota(x))=1-\prod_{i\in S}(1-x_i)$, so $\mathcal{MD}_D|_U$ is the image of $\mathcal{MC}_D|_{\iota(U)}$ under complementing every coordinate and relabeling the points by $\iota$, which preserves shattering and $\tau_\lambda$ (Lemma~\ref{lem:Dinv}(i)).

\emph{Upper bound.} $F_\lambda\le m$, Theorem~\ref{thm:entropy} with $|\mathcal{MC}_D|=2^D$, and $h(1/K)\ge\log_2K/K$.

\emph{Lower bound.} Let $g$ be the least integer $g\ge2$ with $\binom g{\lfloor g/2\rfloor}\ge K$. By Lemma~\ref{lem:Cblocks}, $g\le\lceil2\log_2K\rceil\le D$, using $K\ge2$ and $K\le2^{D/2}$; also $g\le2\log_2K+1\le3\log_2K$. Let $R=\lfloor D/g\rfloor\ge1$ and $t_0=\lceil K\rceil\le\binom g{\lfloor g/2\rfloor}$. By Lemmas~\ref{lem:Cblocks} and~\ref{lem:starprod}(i), any $r\le R$ blocks of $t\le t_0$ points each give $F_\lambda(\mathcal{MC}_D,rt)\ge r\sigma_\lambda(t)$, where $\sigma_\lambda(t)\ge\min\{t,K\}/2$ (Lemma~\ref{lem:star}) and $F_\lambda$ is nondecreasing in $m$ (Lemma~\ref{lem:basic}). If $m\ge Rt_0$, then $F_\lambda\ge R\sigma_\lambda(t_0)\ge RK/2\ge\frac D{2g}\cdot\frac K2\ge\frac{DK}{12\log_2K}$. If $t_0\le m<Rt_0$, then $r=\lfloor m/t_0\rfloor\in[1,R)$ blocks of $t_0$ points give $F_\lambda\ge\lfloor m/t_0\rfloor K/2\ge mK/(4\lceil K\rceil)\ge mK/(4(K+1))\ge m/6$, as $K/(K+1)\ge2/3$. If $m<t_0$, one block of $m$ points gives $\sigma_\lambda(m)\ge\min\{m,K\}/2=m/2$, because $m\le\lceil K\rceil-1<K$.
\end{proof}

\subsection{Proof of Corollary~\ref{cor:natural}}\label{sec:D-proof}

\restatemain{cor:natural}

\noindent Let $\lambda\ge1$, so $\min\{1,K/2\}=1$ and $\min\{2,K\}=2$. For each row of Corollary~\ref{cor:natural}, Propositions~\ref{prop:Dthresholds}--\ref{prop:Dconjunctions} give $a\Psi_\lambda(m)\le F_\lambda(\Hc,m)\le b\Psi_\lambda(m)$ for all $m\ge1$ with the following constants $(a,b)$: $(1,2)$ for threshold functions; $(1,1)$ for parities and full tables; $(1/2,1)$ for $k$-sparse sets; $(1/2,2)$ for unions of at most $k$ intervals, since $\min\{m,2kK\}\le2\min\{m,kK\}$; $(1/8,1)$, $(1/10,1)$ and $(1/12,1)$ for affine halfspaces, homogeneous halfspaces and boxes; $(1/33,2)$ for dictators, since $\min\{m,p,2B\}\le2\min\{m,p,B\}$; and $(1/12,1)$ for monotone conjunctions and disjunctions.

\emph{Nontriviality.} We list $h,h'$ in the class that agree at $x'$ and differ at $x$. Threshold functions: $\one[\,\cdot\ge x]$ and $\one[\,\cdot\ge x']$ for $x<x'$. Full tables, $k$-sparse sets, intervals and boxes ($N\ge2$ in the first two): $0$ and $\one_{\{x\}}$ for $x\ne x'$. Affine halfspaces: $0$ ($w=0$, $b=-1$) and $\one[x_1\ge0]$, with $(x',x)=(-e_1,0)$. Homogeneous halfspaces: $\one[x_1\ge0]$ and $\one[-x_1\ge0]$; parities ($a=0$ and $a=e_1$), dictators ($h_0$ and $h_1$) and disjunctions ($S=\emptyset$ and $S=\{1\}$): all with $(x',x)=(0,e_1)$. Conjunctions: $h_\emptyset$ and $h_{\{1\}}$, with $(x',x)=(\one,0)$.

\emph{Rates.} By the transfer~\eqref{eq:transfer} (Proposition~\ref{prop:transfer}), each of these classes satisfies $\frac a{2e}\Psi_\lambda(n+1)/(n+1)\le M_n(\Hc,\lambda)\le b\Psi_\lambda(n+1)/(n+1)$ for $n\ge1$, and the lower constant $a/(2e)$ is $1/(2e)$ for threshold functions, parities and full tables, $1/(4e)$ for $k$-sparse sets and intervals, $1/(16e)$, $1/(20e)$ and $1/(24e)$ for affine halfspaces, homogeneous halfspaces and boxes, $1/(66e)$ for dictators and $1/(24e)$ for conjunctions and disjunctions.

\emph{Sample sizes.} Consider a row with $\Psi_\lambda(m)=\min\{m,\Psi\}$, where $\Psi$ does not depend on $m$ ($\Psi=1$ for threshold functions), and let $n(\eps)$ be the least $n\ge1$ with $M_n(\Hc,\lambda)\le\eps$. Then $a\Psi/(2e\eps)-1\le n(\eps)\le\max\{1,b\Psi/\eps\}$ for $0<\eps<a/(2e)$, and for $0<\eps\le a/(4e)$, in particular for every row and every $\eps\le1/(132e)$, $a\Psi/(4e\eps)\le n(\eps)\le b\Psi/\eps$. Indeed, $M_n\le F_\lambda(\Hc,n+1)/(n+1)\le b\Psi/(n+1)\le\eps$ once $n+1\ge b\Psi/\eps$. If $M_n\le\eps$, then $a\min\{n+1,\Psi\}\le F_\lambda(\Hc,n+1)\le2e\eps(n+1)$ by~\eqref{eq:transfer}; $n+1\le\Psi$ would give $a\le2e\eps$, so $n+1>\Psi$ and $a\Psi\le2e\eps(n+1)$. For $\eps\le a/(4e)$, $a\Psi/(2e\eps)\ge2$ and $b\Psi/\eps\ge1$ since $b\ge1$ and $\Psi\ge1$; every $\Psi$ in Corollary~\ref{cor:natural} is at least $1$, as $K\ge\log_2K$.

\section{The abstention gain}\label{sec:E}\label{app:gain}

This section proves Corollary~\ref{cor:gain}. Throughout, $\Hc$ is nontrivial, $n\ge1$, $m=n+1$, $K=1+\lambda$ and $B_n=B_n(\Hc,\lambda)=K\,M_n(\Hc,0)/M_n(\Hc,\lambda)$. By Lemma~\ref{lem:nonabst}, $K\,M_n(\Hc,0)$ is the minimax risk $M^{\rm na}_n(\Hc,\lambda)$ of learners that never abstain. Both minimax risks are positive: $\Hc$ has a point $x'$ with $\Hc|_{\{x'\}}=\{0,1\}^{\{x'\}}$, a single edge with $\tau_\lambda=f_K(\frac12,\frac12)=\min\{1,K/2\}$, so $F_\lambda(\Hc,m)\ge\min\{1,K/2\}$ (Lemma~\ref{lem:basic}) and $M_n(\Hc,\lambda)>0$ by~\eqref{eq:transfer}.

\subsection{The cutoff and the classes that attain it}

\begin{lemma}\label{lem:gaincut}
$1\le B_n\le\max\{1,K/2\}$, and $B_n=1$ for $\lambda\le1$.
\end{lemma}

\begin{proof}
Lemma~\ref{lem:nonabst} gives $M_n(\Hc,0)\le M_n(\Hc,\lambda)\le KM_n(\Hc,0)$, hence $B_n\ge1$, and $M_n(\Hc,\lambda)=KM_n(\Hc,0)$ for $\lambda\le1$, hence $B_n=1$. Let $\lambda\ge1$. The loss $\ell_\lambda(\nu,y)=\nu(\abst)+(1+\lambda)\nu(1-y)$ is nondecreasing in $\lambda$ for every $\nu$ and $y$, so every risk, and therefore $M_n(\Hc,\cdot)$, is nondecreasing in $\lambda$. With Lemma~\ref{lem:nonabst} at penalty $1$, $M_n(\Hc,\lambda)\ge M_n(\Hc,1)=2M_n(\Hc,0)$, so $B_n\le K/2$.
\end{proof}

\begin{lemma}\label{lem:gainshattered}
Let $\Hc$ have VC dimension $d\ge2$. For every integer $u$ with $2\le u\le d$, $B_n\ge\frac K2(1-\frac1u)^n$. In particular $B_n\ge\frac K2(\frac n{n+1})^n\ge\frac K{2e}$ if $n+1\le d$, and $B_n=K/2$ for every $\lambda\ge1$ if $d=\infty$.
\end{lemma}

\begin{proof}
Let $T$ be a shattered set with $|T|=u$, so $\Hc|_T=\{0,1\}^T$. By Lemma~\ref{lem:cube}, $\tau_0(\{0,1\}^T)=u/2$, and Lemma~\ref{lem:uniform} gives $M_n(\Hc,0)\ge\frac1u(1-\frac1u)^n\frac u2$. The upper bound in~\eqref{eq:transfer} with $F_\lambda(\Hc,m)\le m$ (Lemma~\ref{lem:basic}) gives $M_n(\Hc,\lambda)\le1$. Hence $B_n\ge\frac K2(1-\frac1u)^n$. For $n+1\le d$ take $u=n+1$ and use $(1+\frac1n)^n\le e$. If $d=\infty$, let $u\to\infty$ and combine with Lemma~\ref{lem:gaincut}.
\end{proof}

Call $\Hc$ \emph{affine} if every projection $\Hc|_U$ on a finite $U\subseteq\X$ is an affine subspace $v_0\oplus L_U$ of $\GF^U$, where $L_U=\{v\oplus v':v,v'\in\Hc|_U\}$ is a linear subspace. Full tables and parities $x\mapsto\langle a,x\rangle\bmod2$ on $\{0,1\}^D$ are affine: $\Hc|_U=\GF^U$ for the full table, and for parities $\Hc|_U$ is the image of the linear map $a\mapsto(\langle a,x\rangle\bmod2)_{x\in U}$. For points $x_1,\dots,x_n,x$ with $U=\{x_1,\dots,x_n,x\}$, call $x$ \emph{undetermined} by $x_1,\dots,x_n$ if some $\ell\in L_U$ has $\ell(x_i)=0$ for all $i$ and $\ell(x)=1$, and let $u^*_n(\Hc)=\sup_P\Pr_{(x_1,\dots,x_n,x)\sim P^{n+1}}(x\text{ undetermined by }x_1,\dots,x_n)$.

\begin{lemma}\label{lem:affine}
Let $\Hc$ be affine and $c_K=\min\{1,K/2\}$. Then $M_n(\Hc,\lambda)=c_K\,u^*_n(\Hc)$. If $\Hc$ is nontrivial, then $B_n=\max\{1,K/2\}$. In particular $B_n=K/2$ for every $\lambda\ge1$ and every $n$ for full tables on at least two points and for parities on $\{0,1\}^D$, $D\ge1$.
\end{lemma}

\begin{proof}
Let $h\in\Hc$ be the target and $C=\{v\in\Hc|_U:v(x_i)=h(x_i),\ i\le n\}=h|_U\oplus\{\ell\in L_U:\ell(x_i)=0\ \forall i\}$. The values $\{v(x):v\in C\}$ are $\{h(x)\}$ if $x$ is determined and $\{0,1\}$ otherwise. Let $\nu^*=\delta_\abst$ if $K\ge2$ and $\nu^*=(\frac12,\frac12,0)$ otherwise, so that $\ell_\lambda(\nu^*,0)=\ell_\lambda(\nu^*,1)=c_K$.

\emph{Upper bound.} The learner computes $C$ from the sample and the test point; it predicts the common value $v(x)$ if all $v\in C$ agree at $x$, outputs $\nu^*$ if they do not, and abstains if $C=\emptyset$ (which never happens on samples labeled by a member of $\Hc$). Its loss at $x$ is $0$ if $x$ is determined and $c_K$ otherwise, so its risk under $(P,h)$ is $c_K$ times the probability that the test point is undetermined, and $M_n(\Hc,\lambda)\le c_Ku^*_n(\Hc)$.

\emph{Lower bound.} Let $P$ have finite support $U_P$, draw $v$ uniformly from $\Hc|_{U_P}$ and use a target $h_v\in\Hc$ with $h_v|_{U_P}=v$. Given the sample, the posterior of $v$ is uniform on the consistent patterns, a nonempty affine subspace of $\GF^{U_P}$. At a test point $x$, the map $v'\mapsto v'(x)$ is affine on this subspace, and it is non-constant exactly when $x$ is undetermined, since the restriction of $L_{U_P}$ to $U=\{x_1,\dots,x_n,x\}$ is $L_U$. A non-constant affine functional over $\GF$ takes each value on half of the subspace, so on this event $v(x)$ is a fair coin independent of the action, whose expected loss is at least $f_K(\frac12,\frac12)=c_K$ by Lemma~\ref{lem:fK}(v). Hence every learner has Bayes risk at least $c_K$ times the probability that $x$ is undetermined under $P$. For an arbitrary $P$ with atoms $x^{(1)},x^{(2)},\dots$, the normalized restrictions $P_N$ of $P$ to $\{x^{(1)},\dots,x^{(N)}\}$ satisfy $\Pr_{P_N}(\text{undetermined})\ge\sum_{z\in\{x^{(1)},\dots,x^{(N)}\}^{n+1}}P^{n+1}(z)\one[x\text{ undetermined}]$, which tends to $\Pr_P(\text{undetermined})$. So $M_n(\Hc,\lambda)\ge c_Ku^*_n(\Hc)$.

\emph{Gain.} For nontrivial $\Hc$, $u^*_n(\Hc)>0$ because $M_n(\Hc,\lambda)>0$, and $B_n=Kc_1u^*_n/(c_Ku^*_n)=K/(2c_K)=\max\{1,K/2\}$. Both examples are nontrivial: for the full table take the zero function and $\one_{\{x'\}}$ at points $x\ne x'$; for parities take $a=0$ and $a=e_1$ at the points $0$ and $e_1$.
\end{proof}

\subsection{The gain through cores}

\begin{lemma}\label{lem:gainwindow}
Let $\lambda\ge1$, $d'=d_{n+1}$ and $\kappa=\kappa_\lambda(\Hc,n+1)$, and write $F_\mu=F_\mu(\Hc,n+1)$. Then
\[
\frac{K\,F_0}{2e\,F_\lambda}\ \le\ B_n\ \le\ \frac{2e\,K\,F_0}{F_\lambda}\qquad\text{and}\qquad\frac{K\,d'}{8e\,\kappa}\ \le\ B_n\ \le\ \frac{4e\lambda\,d'}{\kappa}.
\]
In particular $B_n$ lies within the factors $8e$ and $4e$ of $Kd'/\kappa$, and $KF_0/F_\lambda$ is nondecreasing in $\lambda$.
\end{lemma}

\begin{proof}
By~\eqref{eq:transfer} at the penalties $0$ and $\lambda$, $F_\mu/(2e(n+1))\le M_n(\Hc,\mu)\le F_\mu/(n+1)$ for $\mu\in\{0,\lambda\}$, and dividing gives the first pair. By Theorem~\ref{thm:binary}(a) at $\lambda=0$, $d'/2\le F_0\le d'$, and by Theorem~\ref{thm:binary}(b), $\frac K{2\lambda}\kappa\le F_\lambda\le2\kappa$. Inserting these bounds gives $B_n\ge K(d'/2)/(2e\cdot2\kappa)$ and $B_n\le2eKd'\cdot2\lambda/(K\kappa)$, the second pair; as $\lambda<K$, both ends are within the stated factors of $Kd'/\kappa$. Finally $f_K(a,b)/K=\min\{(a+b)/K,a,b\}$ is nonincreasing in $K$, so $\tau_\lambda(V)/K$ and $F_\lambda(\Hc,m)/K$ are nonincreasing in $\lambda$, and $KF_0/F_\lambda=F_0/(F_\lambda/K)$ is nondecreasing.
\end{proof}

\noindent The first pair of bounds is the statement after Corollary~\ref{cor:gain} that $(1+\lambda)F_0/F_\lambda$ determines $B_n$ within the factor $2e$. For the full table on two points, Lemma~\ref{lem:cube} gives $F_\mu(\Hc,m)=\tau_\mu(\{0,1\}^2)=2\min\{1,(1+\mu)/2\}$ for $m\ge2$, so $(1+\lambda)F_0/F_\lambda=\max\{1,K/2\}$. For single intervals $\mathcal I_1$ on a countably infinite $\X\subseteq\R$ and $m\ge2$, $F_0(\mathcal I_1,m)\le d_m=2$ by Theorem~\ref{thm:binary}(a), and every projection on $m$ points contains the star formed by the empty interval and the $m$ singletons, so $F_\lambda(\mathcal I_1,m)\ge mK/(m+\lambda)$ for $\lambda\ge1$ by Lemmas~\ref{lem:Dinv}(ii) and~\ref{lem:star}. Hence $(1+\lambda)F_0/F_\lambda\le2(1+\lambda/m)$ for $\lambda\ge1$, and $(1+\lambda)F_0/F_\lambda=1$ for $\lambda\le1$; at $m=128$ this is at most $3$ for every $\lambda\le64$.

\subsection{The three regimes}

\restatemain{cor:gain}

\begin{proof}[Proof of Corollary~\ref{cor:gain}]
The bounds $1\le B_n\le\max\{1,K/2\}$ and $B_n=1$ for $\lambda\le1$ are Lemma~\ref{lem:gaincut}; equality for infinite VC dimension is Lemma~\ref{lem:gainshattered}, and for full tables and parities Lemma~\ref{lem:affine}. The relation $B_n\asymp Kd_{n+1}/\kappa_\lambda(\Hc,n+1)$ for $\lambda\ge1$ is Lemma~\ref{lem:gainwindow}. Let $\lambda\ge1$ for the rest of the proof. For a class with $a\Psi_\lambda(m)\le F_\lambda(\Hc,m)\le b\Psi_\lambda(m)$ (Propositions~\ref{prop:Dthresholds}--\ref{prop:Dconjunctions}), Lemma~\ref{lem:gainwindow} and $d'/2\le F_0\le d'$ give
\begin{equation}\label{eq:gainrows}
\frac{K\,d'}{4e\,b\,\Psi_\lambda(m)}\ \le\ B_n\ \le\ \frac{2e\,K\,d'}{a\,\Psi_\lambda(m)},\qquad d'=d_m .
\end{equation}

\emph{(i) Order $K$.} For threshold functions, $d'=1$ and $F_\lambda(\Hc,m)\le2$ (Proposition~\ref{prop:Dthresholds}), so the first pair of Lemma~\ref{lem:gainwindow} with $F_0\ge\frac12$ gives $B_n\ge K/(8e)$, and $B_n\le K/2$. Parities and full tables have $B_n=K/2$ (Lemma~\ref{lem:affine}). Every class with $n+1\le d$ has $B_n\ge K/(2e)$ (Lemma~\ref{lem:gainshattered}).

\emph{(ii) Order $\log K$.} Dictators on $p\ge2$ features have VC dimension $d_p=\lfloor\log_2(p+1)\rfloor$ (Proposition~\ref{prop:Ddictators}), and $\frac23\log_2p\le d_p\le\log_23\cdot\log_2p$: from $p+1<2^{d_p+1}$, $\log_2p\le1=d_p$ if $d_p=1$ and $\log_2p<d_p+1\le\frac32d_p$ if $d_p\ge2$; and $p+1\le p^{\log_23}$ for $p\ge2$. If $Q:=K\log_2p/\log_2K\le\min\{p,m\}$, then $\Psi_\lambda(m)=Q$, and $m\ge Q\ge\log_23\cdot\log_2p\ge d_p$ because $K/\log_2K\ge e\ln2>\log_23$ for $K\ge2$; so $d'=d_p$ and $Kd'/\Psi_\lambda(m)=(d_p/\log_2p)\log_2K$. With the constants $a=\frac1{33}$, $b=2$ of Proposition~\ref{prop:Ddictators}(ii), \eqref{eq:gainrows} gives $\log_2K/(12e)\le B_n\le66e\log_23\cdot\log_2K\le105e\log_2K$. Monotone conjunctions and disjunctions on $D\ge2$ attributes have VC dimension $D$ and the same values $F_\lambda$ (Proposition~\ref{prop:Dconjunctions}); if $K\le2^{D/2}$ and $m\ge KD/\log_2K$, then $m\ge D$ (as $K\ge2$), $d'=D$, and Proposition~\ref{prop:Dconjunctions} gives $\frac1{12}\cdot\frac{KD}{\log_2K}\le F_\lambda(\Hc,m)\le\frac{KD}{\log_2K}$, since $m/6\ge KD/(6\log_2K)$; so \eqref{eq:gainrows} with $\Psi_\lambda(m)=KD/\log_2K$, $a=\frac1{12}$ and $b=1$ gives $\log_2K/(4e)\le B_n\le24e\log_2K$.

\emph{(iii) Constant order.} For unions of at most $k$ intervals (VC dimension $2k$, Proposition~\ref{prop:Dintervals}) with $m\ge kK$, affine halfspaces in $\Q^d$ (VC dimension at most $d+1$, Proposition~\ref{prop:Dhalfspaces}) with $m\ge(d+1)K$, homogeneous halfspaces (VC dimension at most $d$, Proposition~\ref{prop:Dhomogeneous}) with $m\ge dK$, and boxes (VC dimension at most $2d$, Proposition~\ref{prop:Dboxes}) with $m\ge2dK$, the second argument of $\Psi_\lambda(m)$ is the minimum, and $d'\le$ the VC dimension, so $Kd'/\Psi_\lambda(m)$ is at most $2$, $1$, $1$ and $1$; with the lower constants $a=\frac12,\frac18,\frac1{10},\frac1{12}$ of these propositions, \eqref{eq:gainrows} gives $B_n\le8e$, $16e$, $20e$ and $24e$. For $k$-sparse sets the next lemma gives $B_n<2$.
\end{proof}

\begin{lemma}\label{lem:gainsparse}
Let $\lambda\ge1$, $k\ge1$, let $\X$ be a set of $N\in\{2,3,\dots\}\cup\{\infty\}$ keys, $\Sc_k=\{\one_A:A\subseteq\X,|A|\le k\}$, and $t=\lfloor\min\{N-1,n+1\}/k\rfloor\ge1$. Then $1\le B_n(\Sc_k,\lambda)\le1+\lambda/t$. In particular $B_n(\Sc_k,\lambda)<2$ when $kK\le\min\{N-1,n+1\}$.
\end{lemma}

\begin{proof}
The class is nontrivial: for keys $x\ne y$, $\one_\emptyset$ and $\one_{\{x\}}$ agree at $y$ and differ at $x$. Write $\varphi_n=\frac1{n+1}(\frac n{n+1})^n=\max_{q\in[0,1]}q(1-q)^n$.

\emph{Penalty $0$.} The learner that predicts $1$ at $x$ if $(x,1)$ is in the sample and $0$ otherwise never abstains, and under a target $\one_A$ it errs only at the points of $A$ missing from the sample, so its risk is $\sum_{x\in A}P(\{x\})(1-P(\{x\}))^n\le k\varphi_n$. Hence $M_n(\Sc_k,0)\le k\varphi_n$.

\emph{Penalty $\lambda$.} Let $u=\min\{N-1,n+1\}\ge kt$. If $N\ge n+2$, take $U\subseteq\X$ with $|U|=n+1$; then $\Sc_k|_U$ is the family of all indicators of subsets of $U$ of size at most $k$. If $N\le n+1$, take a key $x_0$ and $U=\X\setminus\{x_0\}$, so $|U|=N-1\le n$; the members of $\Sc_k$ that vanish at $x_0$ restrict to the same family on $U$. Partition $U$ into $k$ blocks $T_1,\dots,T_k$ of sizes at least $t$. Every partial transversal of the blocks has at most $k$ elements, so the family star-shatters $(0;T_1,\dots,T_k)$, and Lemma~\ref{lem:starprod}(i), applied to the subclass of members of $\Sc_k$ that vanish at $x_0$ in the second case, with the star values of Lemma~\ref{lem:star} gives $\tau_\lambda\ge\sum_j\sigma_\lambda(|T_j|)\ge k\sigma_\lambda(t)$, with $\sigma_\lambda(t)=tK/(t+\lambda)$ nondecreasing in $t$. In the first case Lemma~\ref{lem:uniform} with $|U|=n+1$, and in the second Lemma~\ref{lem:anchor} with $U'=U$ and $b=0$, give $M_n(\Sc_k,\lambda)\ge\varphi_n\,k\sigma_\lambda(t)$.

\emph{Ratio.} $B_n=KM_n(\Sc_k,0)/M_n(\Sc_k,\lambda)\le Kk\varphi_n/(\varphi_nk\sigma_\lambda(t))=(t+\lambda)/t$. If $kK\le\min\{N-1,n+1\}$, then $t\ge\lfloor K\rfloor>K-1=\lambda$, so $\lambda/t<1$.
\end{proof}

\section{Full tables and the missing mass}\label{sec:F}\label{app:exactcosts}

This section proves Proposition~\ref{prop:exact} and the identity with the Good--Turing quantity stated after it.

\restatemain{prop:exact}

\noindent Throughout, $\X$ is countable, $\Hc=\{0,1\}^{\X}$, $n\ge1$, $\lambda\in[0,\infty)$ and $K=1+\lambda$. For a distribution $P$ on $\X$ we write $p_x=P(\{x\})$, and for a learner $g$ and a target $h\in\Hc$ we write $\Risk^\lambda_n(g;P,h)=\E_{x_{1:n}\sim P^n}R_\lambda(g_S;P,h)$ with $S=((x_i,h(x_i)))_{i\le n}$. Since $\ell_\lambda$ is affine in the action, a randomized learner has the same risks as the learner that outputs its expected action, so we treat learners as maps $S\mapsto g_S$. A learner never abstains if $g_S(x)(\abst)=0$ for all $S$ and $x$. By Tonelli's theorem,
\[
\mm_n(P)=\E_{x_{1:n}\sim P^n}P\big(\X\setminus\{x_1,\dots,x_n\}\big)=\sum_{x\in\X}p_x(1-p_x)^n .
\]

\paragraph{Memorizing learners.} Let $\nu_{\rm c}\in\Pc(\{0,1,\abst\})$ be the uniform guess, $\nu_{\rm c}(0)=\nu_{\rm c}(1)=\frac12$, and let $\nu_K:=\delta_\abst$ if $K\ge2$ and $\nu_K:=\nu_{\rm c}$ if $K<2$. For every $\nu\in\Pc(\{0,1,\abst\})$,
\begin{equation}\label{eq:F-avg}
\tfrac12\big(\ell_\lambda(\nu,0)+\ell_\lambda(\nu,1)\big)=\nu(\abst)\cdot1+\big(1-\nu(\abst)\big)\cdot\tfrac K2\ \ge\ \min\{1,K/2\},
\end{equation}
and the middle expression equals $K/2$ if $\nu(\abst)=0$. Moreover $\ell_\lambda(\nu_K,y)=\min\{1,K/2\}$ and $\ell_\lambda(\nu_{\rm c},y)=K/2$ for $y\in\{0,1\}$. The learners $g^{\rm mem}$ and $g^{\rm c}$ output $\delta_{y_i}$ at a sample point $x=x_i$, with $i$ the smallest such index, and output $\nu_K$, respectively $\nu_{\rm c}$, at every other point. The learner $g^{\rm c}$ never abstains.

\begin{lemma}[A fixed distribution]\label{lem:F-fixedP}
Let $P$ be a distribution on $\X$.
\begin{enumerate}
\item[(i)] For every $h\in\Hc$, $\Risk^\lambda_n(g^{\rm mem};P,h)=\min\{1,K/2\}\,\mm_n(P)$ and $\Risk^\lambda_n(g^{\rm c};P,h)=\frac K2\,\mm_n(P)$.
\item[(ii)] For every learner $g$, every finite $F\subseteq\X$ and every $h_0\in\Hc$, the average of $\Risk^\lambda_n(g;P,h)$ over the $2^{|F|}$ targets $h\in\Hc$ that agree with $h_0$ outside $F$ is at least $\min\{1,K/2\}\sum_{x\in F}p_x(1-p_x)^n$, and at least $\frac K2\sum_{x\in F}p_x(1-p_x)^n$ if $g$ never abstains.
\item[(iii)] Consequently $\inf_g\sup_{h\in\Hc}\Risk^\lambda_n(g;P,h)=\min\{1,K/2\}\,\mm_n(P)$, and the same infimum over learners that never abstain equals $\frac K2\,\mm_n(P)$. The learners $g^{\rm mem}$ and $g^{\rm c}$ attain these values for every $P$.
\end{enumerate}
\end{lemma}

\begin{proof}
(i) Let $S=((x_i,h(x_i)))_{i\le n}$. At a sample point $x=x_i$ the learner $g^{\rm mem}$ outputs $\delta_{h(x)}$, which has loss $0$. At a point $x\notin\{x_1,\dots,x_n\}$ its loss is $\ell_\lambda(\nu_K,h(x))=\min\{1,K/2\}$. Hence $\Risk^\lambda_n(g^{\rm mem};P,h)=\min\{1,K/2\}\,\E\,P(\X\setminus\{x_1,\dots,x_n\})=\min\{1,K/2\}\,\mm_n(P)$. The same computation with $\nu_{\rm c}$ gives the value of $g^{\rm c}$.

(ii) Let $H_F$ be the set of the $2^{|F|}$ targets in question and let $h$ be uniform on $H_F$. Fix $x_{1:n}$ and a point $x\in F\setminus\{x_1,\dots,x_n\}$. The sample depends on $h$ only through $(h(x_i))_{i\le n}$, and $h(x)$ is uniform on $\{0,1\}$ and independent of $(h(x_i))_{i\le n}$, because $x\in F$ differs from every $x_i$. So, given $(h(x_i))_{i\le n}$, the action $\nu=g_S(x)$ is fixed and the expected loss at $x$ is $\frac12(\ell_\lambda(\nu,0)+\ell_\lambda(\nu,1))$, which is at least $\min\{1,K/2\}$ by~\eqref{eq:F-avg}, and equals $K/2$ if $g$ never abstains. The risk $\Risk^\lambda_n(g;P,h)$ is a countable sum of the nonnegative terms $P^n(\{x_{1:n}\})\,p_x\,\ell_\lambda(g_S(x),h(x))$, so the average over $H_F$ can be taken term by term. Keeping only the terms with $x\in F\setminus\{x_1,\dots,x_n\}$, and using $\Pr[x\notin\{x_1,\dots,x_n\}]=(1-p_x)^n$, gives
\[
\frac1{|H_F|}\sum_{h\in H_F}\Risk^\lambda_n(g;P,h)\ \ge\ \min\{1,K/2\}\sum_{x\in F}p_x(1-p_x)^n,
\]
and the same bound with $K/2$ for a learner that never abstains.

(iii) By (ii), $\sup_h\Risk^\lambda_n(g;P,h)\ge\min\{1,K/2\}\sum_{x\in F}p_x(1-p_x)^n$ for every finite $F\subseteq\X$, and the supremum of the right side over finite $F$ is $\min\{1,K/2\}\,\mm_n(P)$. By (i), $g^{\rm mem}$ attains this value at every $h$. The case of learners that never abstain is the same, with $K/2$ and $g^{\rm c}$.
\end{proof}

\begin{proof}[Proof of Proposition~\ref{prop:exact}]
Let $|\X|=N\in\{2,3,\dots\}\cup\{\infty\}$ and $\mm^*_n(N)=\sup_P\mm_n(P)$. For every learner $g$, Lemma~\ref{lem:F-fixedP}(iii) gives
\[
\sup_{P,\,h}\Risk^\lambda_n(g;P,h)\ \ge\ \sup_P\,\min\{1,K/2\}\,\mm_n(P)=\min\{1,K/2\}\,\mm^*_n(N),
\]
and by Lemma~\ref{lem:F-fixedP}(i) the learner $g^{\rm mem}$ has worst-case risk exactly $\min\{1,K/2\}\,\mm^*_n(N)$. Hence $M_n(\Hc,\lambda)=\min\{1,K/2\}\,\mm^*_n(N)$. At $\lambda=0$ this reads $M_n(\Hc,0)=\frac12\mm^*_n(N)$, so $(1+\lambda)M_n(\Hc,0)=\frac K2\mm^*_n(N)$. The same argument over learners that never abstain, with $g^{\rm c}$, shows that their minimax risk is $\frac K2\mm^*_n(N)$, in agreement with Lemma~\ref{lem:nonabst}(i). The class is nontrivial, since for points $x\ne x'$ the zero function and $\one_{\{x\}}$ agree at $x'$ and differ at $x$, and $\mm^*_n(N)\ge2^{-n}>0$ by the uniform distribution on two points. Therefore $B_n(\Hc,\lambda)=\frac K2\mm^*_n(N)/(\min\{1,K/2\}\,\mm^*_n(N))=\max\{1,K/2\}$. The statements for a fixed $P$ are Lemma~\ref{lem:F-fixedP}(iii).
\end{proof}

For a finite sequence $z_{1:j}$ in $\X$, let $N_1(z_{1:j})$ be the number of points that occur exactly once in it.

\begin{corollary}[Missing mass and singletons]\label{cor:goodturing}
Let $P$ be a distribution on $\X$ and $n\ge1$.
\begin{enumerate}
\item[(i)] $\mm_n(P)=\E_{x_{1:n+1}\sim P^{n+1}}N_1(x_{1:n+1})/(n+1)$.
\item[(ii)] $\E\,N_1(x_{1:n})/n-\frac1n(1-\frac1n)^{n-1}\le\mm_n(P)\le\E\,N_1(x_{1:n})/n$ for $x_{1:n}\sim P^n$, with $0^0=1$.
\item[(iii)] Let $|\X|=N\in\{2,3,\dots\}\cup\{\infty\}$, $\Hc=\{0,1\}^{\X}$ and $t\in[\frac12,1)$. The minimax expected score shortfall $\inf_g\sup_{P,h}\E_{x_{1:n}\sim P^n}\big(1-\Score_t(g_S;P,h)\big)$ equals
\[
\mm^*_n(N)=\sup_P\frac{\E\,N_1(x_{1:n+1})}{n+1}
\]
for every such $t$. The learner $g^{\rm mem}$ attains it, never answers wrongly, and has shortfall $\mm_n(P)$ at every $(P,h)$. Among learners that never abstain the minimax shortfall is $\mm^*_n(N)/(2(1-t))$.
\end{enumerate}
\end{corollary}

\begin{proof}
(i) A point $x$ occurs exactly once in $x_{1:n+1}$ with probability $(n+1)p_x(1-p_x)^n$; summing over $x$ by Tonelli's theorem gives $\E\,N_1(x_{1:n+1})=(n+1)\,\mm_n(P)$.
(ii) In the same way $\E\,N_1(x_{1:n})=\sum_xnp_x(1-p_x)^{n-1}$, so $\E\,N_1(x_{1:n})/n-\mm_n(P)=\sum_xp_x\cdot p_x(1-p_x)^{n-1}$. As $\sum_xp_x=1$, this lies between $0$ and $\max_{q\in[0,1]}q(1-q)^{n-1}=\frac1n(1-\frac1n)^{n-1}$, attained at $q=1/n$.
(iii) By the identity $1-\Score_t=R_{\lambda(t)}$ of Section~\ref{sec:loss}, the expected shortfall of a learner is its risk at $\lambda(t)=t/(1-t)$, where $K=1/(1-t)\ge2$ and $\min\{1,K/2\}=1$. So Proposition~\ref{prop:exact} gives the value $\mm^*_n(N)$, its proof gives the value $\frac K2\mm^*_n(N)=\mm^*_n(N)/(2(1-t))$ for learners that never abstain, and (i) gives the expression through singletons. For $K\ge2$ the learner $g^{\rm mem}$ repeats observed labels and abstains elsewhere, so it never answers wrongly, and Lemma~\ref{lem:F-fixedP}(i) gives its shortfall.
\end{proof}

\begin{remark}[Calibrated language models]\label{rem:F-KV}
For arbitrary facts and under their assumptions, including sparsity and regularity, \citet{KV24} show that, with high probability over the random world and the training sample, a calibrated language model generates false statements with probability at least the fraction of facts that occur exactly once in its training data, up to the miscalibration and correction terms of their bound; this fraction is the Good--Turing estimate of the missing mass. Corollary~\ref{cor:goodturing} gives the corresponding exact minimax identities for binary questions with a reject option. The keys of the full table play the role of arbitrary facts, since the label of a key is not determined by the labels of the other keys. A learner that must answer has minimax probability of a wrong answer equal to half the supremum over $P$ of the expected fraction of singletons in a sample of size $n+1$ (Proposition~\ref{prop:exact} at $\lambda=0$), and by Lemma~\ref{lem:F-fixedP} it equals half the expected fraction at each fixed $P$. When abstention is scored at a threshold $t\ge1/2$, the minimax shortfall equals this supremum itself, for every such $t$, and it is attained by a learner that never answers wrongly.
\end{remark}

\section{Three examples}\label{sec:G}\label{app:panelexamples}

This section proves the statements of the main text on three examples: two classes with equal VC dimension and star number (Section~\ref{sec:beyond}), group-sparse supports (Sections~\ref{sec:panels} and~\ref{sec:ov-panels}), and stars and cubes (Section~\ref{sec:gain}). Let $1\le\lambda<\infty$ and $K=1+\lambda$ unless stated otherwise. For $S\subseteq U$ we write $\one_S\in\{0,1\}^U$ for its indicator and $e_x=\one_{\{x\}}$.

\paragraph{Tools.} As in Section~\ref{sec:A}, a rule $q$ on a finite nonempty $V\subseteq\{0,1\}^U$ assigns an action $q_e\in\Pc(\{0,1,\abst\})$ to each edge $e$ of $G(V)$. For an endpoint $v$ of an edge $e$ in direction $x$, the cost of $q_e$ to $v$ is $\ell_\lambda(q_e,v(x))$, and the leave-one-out cost $L_q(v)$ of $v$ is the sum of these costs over the edges at $v$. So $\delta_y$ costs $0$ to the endpoint with label $y$ at $x$ and $K$ to the other endpoint, and $\delta_\abst$ costs $1$ to both. Lemma~\ref{lem:lp}(ii) gives
\begin{equation}\label{eq:G-rule}
\tau_\lambda(V)\le\max_{v\in V}L_q(v)\qquad\text{for every rule }q\text{ on }V .
\end{equation}
For lower bounds we use the inequality $F_\lambda(\Hc,m)\ge\frac K{2\lambda}\,\pi_\lambda(\Hc,m)$ of Theorem~\ref{thm:binary}(b). We also use $\tau_\lambda(V)\le\Delta(V)\le|U|$ and the monotonicity of $F_\lambda(\Hc,m)$ in $m$ (Lemma~\ref{lem:basic}).

\subsection{Equal VC dimension and star number}\label{sec:G-FL}

For integers $c\ge0$ and $k,N\ge1$, let $\X=C\sqcup T$ with $|C|=c$ and $|T|=N$, and let $\mathrm{FL}(c,k,N)$ be the class of all $h\in\{0,1\}^{\X}$ with at most $k$ ones on $T$. For every $U\subseteq\X$,
\begin{equation}\label{eq:G-FLproj}
\mathrm{FL}(c,k,N)|_U=\big\{v\in\{0,1\}^U:\ v\text{ has at most }k\text{ ones on }U\cap T\big\},
\end{equation}
since a restriction has at most as many ones on $U\cap T$ as the function has on $T$, and a pattern with at most $k$ ones on $U\cap T$ extends by zeros.

\begin{proposition}\label{prop:FL}
Let $c\ge0$, $k,N,d\ge1$, $\Hc_1=\mathrm{FL}(d-1,1,N)$ and $\Hc_2=\mathrm{FL}(0,d,N+d-1)$.
\begin{enumerate}
\item[(a)] $\mathrm{FL}(c,k,N)$ has VC dimension $c+\min\{k,N\}$ and star number $c+N$. In particular $\Hc_1$ and $\Hc_2$ both have VC dimension $d$ and star number $N+d-1$.
\item[(b)] $F_\lambda(\mathrm{FL}(c,k,N),m)\le\min\{m,\,c+\min\{N,kK\}\}$ for every $m\ge1$. In particular $F_\lambda(\Hc_1,m)\le d-1+\min\{N,K\}$.
\item[(c)] The $(d-1)$-subsets and the $d$-subsets of the $N+d-1$ points of the domain of $\Hc_2$ form a panel system with $a=N$ and $b=d$, which $\Hc_2$ shatters with center $0$. Hence $F_\lambda(\Hc_2,m)\ge\frac K{2\lambda}\min\{N,\lambda d\}$ for $m\ge N+d-1$.
\item[(d)] If $K\le N/d$ and $m\ge N+d-1$, then $F_\lambda(\Hc_2,m)\ge Kd/2$ and
\[
F_\lambda(\Hc_2,m)\ \ge\ \frac{dK}{2(d-1+K)}\,F_\lambda(\Hc_1,m)\ \ge\ \frac14\min\{d,K\}\,F_\lambda(\Hc_1,m).
\]
\end{enumerate}
\end{proposition}

\begin{proof}
(a) By~\eqref{eq:G-FLproj}, a set $U$ is shattered if and only if the all-ones pattern on $U$ is realized, that is, if and only if $|U\cap T|\le k$; the largest such sets have $c+\min\{k,N\}$ points. In $G(\mathrm{FL}(c,k,N)|_U)$ the zero pattern is adjacent to $e_x$ for every $x\in U$, since $k\ge1$, and no degree exceeds $|U|$. So the largest degree in the one-inclusion graphs of the projections is $|\X|=c+N$, which is the star number. For $\Hc_1$ this gives $(d-1)+\min\{1,N\}=d$ and $(d-1)+N$, and for $\Hc_2$ it gives $\min\{d,N+d-1\}=d$ and $N+d-1$.

(b) Fix $U$ with $|U|\le m$ and let $V=\mathrm{FL}(c,k,N)|_U$. By Lemma~\ref{lem:basic}, $\tau_\lambda(V)\le|U|\le\min\{m,c+N\}$. Let $q$ abstain on every edge in a direction $x\in C$ and answer $0$ on every edge in a direction $x\in T$. A pattern $v\in V$ has at most $c$ edges in directions of $C$, each of cost $1$. An edge at $v$ in a direction $x\in T$ costs $K$ if $v(x)=1$ and $0$ otherwise, and $v$ has at most $k$ ones on $U\cap T$. So $L_q(v)\le c+kK$, and~\eqref{eq:G-rule} gives $\tau_\lambda(V)\le c+kK$. Hence $\tau_\lambda(V)\le\min\{m,c+\min\{N,kK\}\}$; maximize over $U$. For $\Hc_1$, $c+\min\{N,kK\}=d-1+\min\{N,K\}$.

(c) Let $T$ be the domain of $\Hc_2$, with $|T|=N+d-1\ge d$. Each $(d-1)$-subset of $T$ lies in exactly $|T|-(d-1)=N$ of the $d$-subsets, and each $d$-subset contains exactly $d$ of the $(d-1)$-subsets; so the two families form a panel system of rank $d$ with $a=N$ and $b=d$. Its sets have at most $d$ points, so by~\eqref{eq:G-FLproj} the patterns $0\oplus\one_S=\one_S$ lie in $\Hc_2|_T$, and $\Hc_2$ shatters it with center $0$. Thus $\pi_\lambda(\Hc_2,m)\ge\min\{N,\lambda d\}$ for $m\ge N+d-1$, and Theorem~\ref{thm:binary}(b) gives the bound on $F_\lambda(\Hc_2,m)$.

(d) From $K\le N/d$ we get $\lambda d<Kd\le N$, so (c) gives $F_\lambda(\Hc_2,m)\ge\frac K{2\lambda}\lambda d=Kd/2$. By (b), $F_\lambda(\Hc_1,m)\le d-1+K$, which gives the first ratio. Since $d-1+K<2\max\{d,K\}$, we have $dK/(2(d-1+K))\ge dK/(4\max\{d,K\})=\frac14\min\{d,K\}$.
\end{proof}

\subsection{Group-sparse supports}\label{sec:G-group}

Let $L,k\ge1$ be integers, and let the countable domain $\X$, with $|\X|=N\in\{kL,(k+1)L,\dots\}\cup\{\infty\}$, be partitioned into groups of $L$ keys each, so that there are $N/L\ge k$ groups. For $S\subseteq\X$ let $\gamma(S)$ be the number of groups that $S$ meets, let
\[
\Gs=\{\one_S:\ S\subseteq\X,\ \gamma(S)\le k\},
\]
and let $\Psi(m)=\min\{m,N,k\max\{L,K\}\}$. For $L=1$, $\Gs$ is the class of $k$-sparse sets on $N$ keys.

\paragraph{Basic facts.} For every finite $U\subseteq\X$, $\Gs|_U=\{\one_S:S\subseteq U,\ \gamma(S)\le k\}$, since the restriction of $\one_S$ is $\one_{S\cap U}$ and $\gamma(S\cap U)\le\gamma(S)$. Hence a finite set $U$ is shattered if and only if $\gamma(U)\le k$, and $\Gs$ has VC dimension $d=kL$, attained by the union of $k$ groups. The zero pattern of $\Gs|_U$ is adjacent to $e_x$ for every $x\in U$, so $\max_{|U|\le m}\Delta(\Gs|_U)=\min\{m,N\}$ and the star number is $s=N$. If $N\ge2$, $\Gs$ is nontrivial: for keys $x\ne x'$, the zero function and $\one_{\{x\}}$ agree at $x'$ and differ at $x$.

\paragraph{Layers.} Fix a finite $U\subseteq\X$, let $V=\Gs|_U$ and $V_j=\{\one_S\in V:\gamma(S)\le j\}$ for $0\le j\le k$, so that $V_0=\{0\}$ and $V_k=V$. Let $\one_S\in V$ with $\gamma(S)=j\ge1$. Its neighbors are the patterns $\one_{S\setminus\{x\}}$ with $x\in S$ and $\one_{S\cup\{x\}}$ with $x\in U\setminus S$ and $\gamma(S\cup\{x\})\le k$. Adding a point does not decrease $\gamma$, and $\gamma(S\cup\{x\})=j$ exactly when $x$ lies in one of the $j$ groups that $S$ meets. So the neighbors of $\one_S$ in $V_j$ are the patterns $\one_{S\triangle\{x\}}$ with $x\in U$ in one of these groups, and a neighbor in $V_{j-1}$ is a pattern $\one_{S\setminus\{x\}}$ with $x$ the only element of $S$ in its group. There is at most one such $x$ in each group that $S$ meets, so
\begin{equation}\label{eq:G-layers}
|N_V(\one_S)\cap V_j|\le jL\le kL\qquad\text{and}\qquad|N_V(\one_S)\cap V_{j-1}|\le j\le k ,
\end{equation}
where $N_V(v)$ is the set of neighbors of $v$ in $G(V)$.

\begin{proposition}\label{prop:groupsparse}
Let $1\le\lambda<\infty$ and $K=1+\lambda$.
\begin{enumerate}
\item[(a)] For every finite $U\subseteq\X$, every two-sided core $(A,W)$ of $G(\Gs|_U)$ has $\delta_A\le kL$ or $\delta_W\le k$, and every panel system that $\Gs$ shatters with some center has $a\le kL$ or $b\le k$. Hence $\pi_\lambda(\Gs,m)\le\kappa_\lambda(\Gs,m)\le\min\{m,k\max\{L,\lambda\}\}$ for every $m\ge1$.
\item[(b)] For every finite $U\subseteq\X$, let $q$ be the rule on $G(\Gs|_U)$ that, on the edge between $\one_S$ and $\one_{S\cup\{x\}}$, answers $0$ if $x$ lies in a group that $S$ does not meet and abstains otherwise. In the leave-one-out game, $q$ abstains at a hidden key whose group contains another positive key, and it answers $0$ otherwise. Then $L_q(v)\le k(L+\lambda)$ for every $v\in\Gs|_U$, and for every $m\ge1$
\[
\tfrac12\Psi(m)\ \le\ F_\lambda(\Gs,m)\ \le\ \min\{m,N,k(L+\lambda)\}\ \le\ 2\Psi(m).
\]
\item[(c)] If $N\ge2$, then $\Psi(n+1)/(4e(n+1))\le M_n(\Gs,\lambda)\le2\Psi(n+1)/(n+1)$ for every $n\ge1$.
\item[(d)] If $N\ge k\max\{L,K\}$ and $n+1\ge k\max\{L,K\}$, then $\min\{K,L\}/(8e)\le B_n(\Gs,\lambda)\le4e\min\{K,L\}$.
\item[(e)] Let $m\ge kL$ and $F=F_\lambda(\Gs,m)$. The right sides of the bounds $F\le Kd_m$, $F\le s$ and of the entropy bound $F\le\max_{|U|\le m}\log_2|\Gs|_U|/h(1/K)$ of Theorem~\ref{thm:entropy}, with $h$ the binary entropy, satisfy
\[
\frac{Kd_m}F\ge\frac12\min\{K,L\},\qquad\frac sF\ge\frac{\min\{m,N\}}{k(L+\lambda)},\qquad\max_{|U|\le m}\frac{\log_2|\Gs|_U|}{h(1/K)\,F}\ge\frac{\min\{K,L\}}{2\log_2(eK)} .
\]
If $K,L\to\infty$ with $\min\{K,L\}/\log_2K\to\infty$ and $m,N\ge k\max\{L,K\}\log_2K$, then all three ratios tend to infinity.
\end{enumerate}
\end{proposition}

\begin{proof}
(a) Let $(A,W)$ be a two-sided core of $G(V)$, $V=\Gs|_U$, and suppose that $\delta_A>kL$ and $\delta_W>k$. Let $j$ be the largest value of $\gamma(S)$ over the patterns $\one_S\in A\cup W$, so that $A\cup W\subseteq V_j$, and $j\ge1$ because $V_0$ has only one element. A pattern of $A$ with $\gamma(S)=j$ has more than $kL$ neighbors in $W\subseteq V_j$, which contradicts~\eqref{eq:G-layers}; so $A\subseteq V_{j-1}$. A pattern of $W$ with $\gamma(S)=j$ then has more than $k$ neighbors in $A\subseteq V_{j-1}$, which again contradicts~\eqref{eq:G-layers}; so $W\subseteq V_{j-1}$ as well, contrary to the choice of $j$. Hence $\delta_A\le kL$ or $\delta_W\le k$. A panel system that $\Gs$ shatters with center $c$ lives on a finite set $U$, and by Lemma~\ref{lem:panelcore}(i) its patterns $c\oplus\one_S$ form a two-sided core of $G(\Gs|_U)$ with $\delta_A\ge a$ and $\delta_W\ge b$; so $a\le kL$ or $b\le k$. Every two-sided core has $\min\{\delta_A,\lambda\delta_W\}\le\max\{kL,\lambda k\}$, so $\kappa_\lambda(\Gs|_U)\le k\max\{L,\lambda\}$. Lemma~\ref{lem:panelcore}(i) and Lemma~\ref{lem:basic}(i) give $\pi_\lambda(\Gs|_U)\le\kappa_\lambda(\Gs|_U)\le\Delta(\Gs|_U)\le|U|$. Maximizing over $|U|\le m$ proves (a).

(b) Every edge of $G(V)$ joins some $\one_S$ and $\one_{S\cup\{x\}}$ with $x\in U\setminus S$, so $q$ is defined on all edges. Let $v=\one_S$ with $\gamma(S)=j$. An edge from $v$ to $\one_{S\cup\{x\}}$ with $x$ in a group that $S$ does not meet costs $0$, since $q$ answers $0=v(x)$. An edge from $v$ to $\one_{S\cup\{x\}}$ with $x$ in a group that $S$ meets costs $1$. An edge from $v$ to $\one_{S\setminus\{x\}}$ costs $K$ if $x$ is the only element of $S$ in its group, since $q$ then answers $0\ne v(x)$, and costs $1$ otherwise. By the description of the neighbors before~\eqref{eq:G-layers}, this gives
\[
L_q(v)=|N_V(v)\cap V_j|+\lambda\,|N_V(v)\cap V_{j-1}|\le jL+\lambda j\le k(L+\lambda)
\]
for $j\ge1$, while $L_q(0)=0$. By~\eqref{eq:G-rule} and Lemma~\ref{lem:basic}, $\tau_\lambda(V)\le\min\{|U|,N,k(L+\lambda)\}$, and maximizing over $|U|\le m$ gives the upper bound on $F_\lambda(\Gs,m)$. Since $L+\lambda<L+K\le2\max\{L,K\}$, this upper bound is at most $2\Psi(m)$.

For the lower bound, a subset of the union of $k$ groups with $d_m=\min\{m,kL\}$ points is shattered, so by Lemma~\ref{lem:cube} $F_\lambda(\Gs,m)\ge d_m$. Next let $u=\min\{m,N\}$, let $U\subseteq\X$ have $|U|=u$, and let $j=\min\{k,\lceil u/K\rceil\}$, so that $1\le j\le\min\{k,u\}$. The $(j-1)$-subsets and the $j$-subsets of $U$ form a panel system with $a=u-j+1$ and $b=j$, and its sets meet at most $j\le k$ groups, so $\Gs$ shatters it with center $0$. Its value is at least $\frac\lambda K\min\{u,kK\}$. Indeed, if $j=\lceil u/K\rceil$, then $u-j+1>u-u/K=\lambda u/K$ and $\lambda j\ge\lambda u/K$. Otherwise $j=k<u/K$, so $u-k+1>kK-k=\lambda k$, and the value is $\lambda k=\frac\lambda K\,kK$. Theorem~\ref{thm:binary}(b) gives $F_\lambda(\Gs,m)\ge\frac K{2\lambda}\cdot\frac\lambda K\min\{u,kK\}=\frac12\min\{u,kK\}$. If $L\ge K$, then $\Psi(m)=\min\{m,kL\}=d_m$, because $kL\le N$. If $L<K$, then $\Psi(m)=\min\{u,kK\}$. In both cases $F_\lambda(\Gs,m)\ge\frac12\Psi(m)$.

(c) By the basic facts $\Gs$ is nontrivial, so~\eqref{eq:transfer} and (b) give $M_n(\Gs,\lambda)\ge F_\lambda(\Gs,n+1)/(2e(n+1))\ge\Psi(n+1)/(4e(n+1))$ and $M_n(\Gs,\lambda)\le F_\lambda(\Gs,n+1)/(n+1)\le2\Psi(n+1)/(n+1)$. The bound of (a) alone already gives the order of the rate: with Theorem~\ref{thm:binary}(b) and Lemma~\ref{lem:basic}, $F_\lambda(\Gs,m)\le\min\{N,4\pi_\lambda(\Gs,m)\}\le4\Psi(m)$.

(d) Let $m=n+1$. Then $\Gs$ is nontrivial because $N\ge K\ge2$, $d_m=kL$ because $m\ge kL$, and $\Psi(m)=k\max\{L,K\}$. By Theorem~\ref{thm:binary}(a) at $\lambda=0$, $kL/2\le F_0(\Gs,m)\le kL$, and by (b), $\frac12k\max\{L,K\}\le F_\lambda(\Gs,m)\le k(L+\lambda)$. Lemma~\ref{lem:gainwindow} gives
\[
\frac{K\cdot kL/2}{2e\cdot k(L+\lambda)}\ \le\ B_n(\Gs,\lambda)\ \le\ \frac{2eK\cdot kL}{\frac12k\max\{L,K\}}=4e\min\{K,L\},
\]
where the equality uses $KL/\max\{L,K\}=\min\{K,L\}$. Since $L+\lambda<2\max\{L,K\}$, the left side is at least $KL/(8e\max\{L,K\})=\min\{K,L\}/(8e)$.

(e) Since $m\ge kL$, $d_m=kL$, and $F\le k(L+\lambda)\le2k\max\{L,K\}$ by (b). First, $Kd_m/F\ge KkL/(k(L+\lambda))\ge KL/(2\max\{L,K\})=\frac12\min\{K,L\}$. Second, $s=N\ge\min\{m,N\}$ and $F\le k(L+\lambda)$. Third, the union $U$ of $k$ groups has $kL\le m$ points and $\Gs|_U=\{0,1\}^U$, so $\log_2|\Gs|_U|=kL$. For $q\in(0,1)$, $-\ln(1-q)\le q/(1-q)$ gives $(1-q)\log_2\frac1{1-q}\le q\log_2e$, hence $h(q)\le q\log_2(e/q)$ and $h(1/K)\le\log_2(eK)/K$. So $kL/h(1/K)\ge kLK/\log_2(eK)$, and dividing by $F\le2k\max\{L,K\}$ gives the third ratio. Under the limit conditions, $\min\{K,L\}\to\infty$; $\log_2(eK)\le2\log_2K$ once $K\ge e$, so the third ratio is at least $\min\{K,L\}/(4\log_2K)\to\infty$; and $\min\{m,N\}/(k(L+\lambda))\ge k\max\{L,K\}\log_2K/(2k\max\{L,K\})=\frac12\log_2K\to\infty$.
\end{proof}

\noindent For $L=1$, part~(b) reads $\frac12\min\{m,N,kK\}\le F_\lambda(\Gs,m)\le\min\{m,N,kK\}$, the row of $k$-sparse sets in Corollary~\ref{cor:natural} for $k\le N$.

\subsection{Stars and cubes}\label{sec:G-starcube}

Let $Q_d=\{0,1\}^{[d]}$ for $d\ge1$, and let $S_m=\{0,e_1,\dots,e_m\}\subseteq\{0,1\}^{[m]}$ for $m\ge1$, whose edges are $\{0,e_i\}$, $i\in[m]$. In this subsection $\lambda\in[0,\infty)$ and $t=\lambda/K$, so that $1-t=1/K$. By Lemma~\ref{lem:nonabst}(iv), the leave-one-out value of $V$ against rules that never abstain is $K\tau_0(V)$, and we compare it with $\tau_\lambda(V)$.

\paragraph{Posteriors.} Let $\mu\in\Pc(V)$ and let $e=\{u,w\}$ be an edge of $G(V)$ in direction $x$ with $\mu_u+\mu_w>0$. If the target pattern is drawn from $\mu$ and its coordinate $x$ is hidden, then, when the visible coordinates are those of $u$, the target is $u$ or $w$, and the posterior probability of $w$ is $p=\mu_w/(\mu_u+\mu_w)$. Answering $u(x)$ then has expected loss $Kp$, abstaining has loss $1$, and answering $w(x)$ has expected loss $K(1-p)$. So answering $u(x)$ is at least as good as abstaining if and only if $p\le1-t$, with equality exactly when $\mu_u=\lambda\mu_w$. Multiplying by $\mu_u+\mu_w$ gives the Bayes costs $K\mu_w$, $\mu_u+\mu_w$ and $K\mu_u$ on $e$, whose minimum is $f_K(\mu_u,\mu_w)$; a rule is a Bayes rule for $\mu$ if it takes a minimizing action on every edge, and by Lemma~\ref{lem:lp}(i) its Bayes cost is $\Phi_\lambda(\mu;V)=\sum_{\{u,w\}\in E(V)}f_K(\mu_u,\mu_w)$.

\begin{proposition}\label{prop:starcube}
Let $\lambda\in[0,\infty)$, $K=1+\lambda$ and $t=\lambda/K$.
\begin{enumerate}
\item[(a)] (Cube.) $\tau_0(Q_d)=d/2$ and $\tau_\lambda(Q_d)=\min\{1,K/2\}\,d$, so $K\tau_0(Q_d)/\tau_\lambda(Q_d)=K/2$ for $\lambda\ge1$. The uniform prior attains $\tau_\lambda(Q_d)$ for every $\lambda$, and under it the posterior on every edge is $\frac12$, so abstaining has loss $1$ and answering has expected loss $K/2$. For $\lambda>1$ abstaining is the only Bayes action on every edge.
\item[(b)] (Star.) $\tau_0(S_m)=m/(m+1)$, and $\tau_\lambda(S_m)=Km/(m+\lambda)$ for $\lambda\ge1$, so $K\tau_0(S_m)/\tau_\lambda(S_m)=(m+\lambda)/(m+1)$, which lies in $[1,2)$ when $m\ge\lambda\ge1$. For $\lambda\ge1$, the prior $\mu^*$ with $\mu^*_0=\lambda/(m+\lambda)$ and $\mu^*_{e_i}=1/(m+\lambda)$ attains $\tau_\lambda(S_m)$. Under $\mu^*$ the posterior of the exception $e_i$ on the edge $\{0,e_i\}$ is $1-t$, answering the default label $0$ and abstaining have the same Bayes cost $K/(m+\lambda)$ on every edge, and answering $0$ on every edge is a Bayes rule.
\end{enumerate}
\end{proposition}

\begin{proof}
(a) The values are Lemma~\ref{lem:cube}. The cube has $d2^{d-1}$ edges, and under the uniform prior each contributes $f_K(2^{-d},2^{-d})=\min\{2,K\}2^{-d}$, in total $\min\{1,K/2\}\,d=\tau_\lambda(Q_d)$. Both endpoints of every edge have mass $2^{-d}$, so the posterior is $\frac12$, and the Bayes cost of an action $\nu$ on an edge is $2^{-d}\big(2\nu(\abst)+K(\nu(0)+\nu(1))\big)$, which for $K>2$ is minimized only by $\nu=\delta_\abst$. For $\lambda\ge1$, $K\tau_0(Q_d)/\tau_\lambda(Q_d)=K(d/2)/d=K/2$.

(b) Write $\alpha=\mu_0$ and $\beta_i=\mu_{e_i}$. Then $\Phi_0(\mu;S_m)=\sum_i\min\{\alpha,\beta_i\}\le\min\{m\alpha,1-\alpha\}\le m/(m+1)$, since the two terms meet at $\alpha=1/(m+1)$, and the uniform prior attains $m/(m+1)$. For $\lambda\ge1$, $\tau_\lambda(S_m)=Km/(m+\lambda)$ by Lemma~\ref{lem:star}. Under $\mu^*$, $\alpha=\lambda\beta$ with $\beta=1/(m+\lambda)$ and $\alpha\ge\beta$, so $f_K(\alpha,\beta)=\min\{\alpha+\beta,K\alpha,K\beta\}=K\beta$ on each of the $m$ edges, and $\Phi_\lambda(\mu^*;S_m)=Km/(m+\lambda)=\tau_\lambda(S_m)$. The posterior of $e_i$ is $\beta/(\alpha+\beta)=1/K=1-t$. Answering $0$ costs $K\beta$, abstaining costs $\alpha+\beta=K\beta$, and answering $1$ costs $K\alpha\ge K\beta$, so answering $0$ attains the minimum $f_K(\alpha,\beta)$ on every edge. Finally $K\frac m{m+1}\cdot\frac{m+\lambda}{Km}=\frac{m+\lambda}{m+1}=1+\frac{\lambda-1}{m+1}$, which is at least $1$ for $\lambda\ge1$ and less than $2$ when $m\ge\lambda$.
\end{proof}

A cube $Q_u$ has the two-sided core formed by its patterns with an even and with an odd number of ones, with $\delta_A=\delta_W=u$, and a star $S_m$ has the core $(\{0\},\{e_1,\dots,e_m\})$ with $\delta_A=m$ and $\delta_W=1$. For a class, the following criterion relates the two-sided cores of its projections on at most $n+1$ points to the abstention gain.

\begin{corollary}\label{cor:starcube}
Let $1\le\lambda<\infty$, $K=1+\lambda$, $n\ge1$ and $g\ge1$, let $\Hc$ be nontrivial and $d'=d_{n+1}$.
\begin{enumerate}
\item[(a)] If every two-sided core of every $G(\Hc|_U)$ with $|U|\le n+1$ has $\min\{\delta_A,\lambda\delta_W\}\le g\,d'$, then $B_n(\Hc,\lambda)\ge\max\{1,K/(8eg)\}$.
\item[(b)] If $B_n(\Hc,\lambda)\le g$, then some $G(\Hc|_U)$ with $|U|\le n+1$ has a two-sided core with $\delta_A\ge Kd'/(8eg)$ and $\delta_W\ge d'/(8eg)$.
\item[(c)] If some $G(\Hc|_U)$ with $|U|\le n+1$ has a two-sided core with $\delta_A\ge Kd'/g$ and $\delta_W\ge d'/g$, then $B_n(\Hc,\lambda)\le4eg$.
\end{enumerate}
\end{corollary}

\begin{proof}
Let $\kappa=\kappa_\lambda(\Hc,n+1)$. A nontrivial class has $d\ge1$, so $d'\ge1$, and $\kappa\ge d'$ by Lemma~\ref{lem:panelcore}(iii). By Lemma~\ref{lem:gainwindow}, $Kd'/(8e\kappa)\le B_n\le4e\lambda d'/\kappa$, and $B_n\ge1$ by Lemma~\ref{lem:gaincut}. (a) The hypothesis gives $\kappa\le gd'$, so $B_n\ge K/(8eg)$. (b) Here $\kappa\ge Kd'/(8eB_n)\ge Kd'/(8eg)$. As $\kappa>0$, it is the value $\min\{\delta_A,\lambda\delta_W\}$ of a two-sided core of some $G(\Hc|_U)$ with $|U|\le n+1$, and this core has $\delta_A\ge\kappa\ge Kd'/(8eg)$ and $\delta_W\ge\kappa/\lambda\ge d'/(8eg)$, since $K>\lambda$. (c) The given core has $\min\{\delta_A,\lambda\delta_W\}\ge\min\{K,\lambda\}d'/g=\lambda d'/g$, so $\kappa\ge\lambda d'/g$ and $B_n\le4eg$.
\end{proof}

\noindent Here $B_n(\Hc,\lambda)=(1+\lambda)M_n(\Hc,0)/M_n(\Hc,\lambda)$ is the ratio of minimax risks of Section~\ref{sec:A}, while Proposition~\ref{prop:starcube} concerns the leave-one-out ratio $K\tau_0(V)/\tau_\lambda(V)$ of a fixed pattern set. For classes, the statements of the main text on stars and cubes rest on the following results: full tables on at least two points have $B_n(\Hc,\lambda)=K/2$ for every $\lambda\ge1$ and every $n$ (Proposition~\ref{prop:exact}, proved in Section~\ref{sec:F}, and Lemma~\ref{lem:affine}), and $k$-sparse sets on $N$ keys have $B_n(\Sc_k,\lambda)<2$ when $kK\le\min\{N-1,n+1\}$ (Lemma~\ref{lem:gainsparse}), although they contain a cube on any $k$ keys.

\section{High-probability bounds}\label{sec:H}\label{app:highprob}

This section proves Proposition~\ref{prop:highprob}.

\restatemain{prop:highprob}

\noindent Part~(a) follows from Theorems~\ref{thm:H-prefix} and~\ref{thm:H-dis}. Parts~(b) and~(c) are proved at the end of the section from Theorem~\ref{thm:H-prefix}, a bound for the disagreement-based learner (Theorem~\ref{thm:H-dis}) and lower bounds (Theorem~\ref{thm:H-lower}).

\paragraph{Quantiles.} Throughout, $\lambda\in[0,\infty)$ unless stated otherwise and $K=1+\lambda$. Here a learner at sample size $n$ is a map $S\mapsto g_S$ from $(\X\times\{0,1\})^n$ to rules, which may depend on the order of the sample; a rule may randomize its action at each point, and its risk averages over the action. Learners that also use a random seed independent of the sample are treated after Theorem~\ref{thm:H-lower}. For $h\in\Hc$ and $w\in\X^k$ let $S_h(w):=((w_i,h(w_i)))_{i\le k}$, and let $S_{\le t}$ be the first $t$ pairs of a sample $S$. For a learner $g$, a distribution $P$ on $\X$, $h\in\Hc$ and $\delta\in(0,1)$, let $R(S):=R_\lambda(g_S;P,h)$ with $S=S_h(x_{1:n})$, $x_{1:n}\sim P^n$, and
\[
Q_\delta(g;P,h):=\inf\{\eps\ge0:\Pr(R(S)>\eps)\le\delta\},\quad \eps_n(\Hc,\lambda,\delta):=\inf_g\ \sup_{P,\,h\in\Hc}Q_\delta(g;P,h),
\]
the infimum over all learners at sample size $n$, which may depend on $\delta$. As $\eps\mapsto\Pr(R(S)>\eps)$ is right-continuous and nonincreasing, $\Pr(R(S)>Q_\delta(g;P,h))\le\delta$; $Q_\delta$ is nonincreasing in $\delta$, and $\eps_n(\Hc,\lambda,\delta)\le1$ since the learner that always abstains has risk $1$. The sample space $\X^n$ is countable.

\paragraph{Star sets and disagreement.} For $m\ge1$, $\Delta_m(\Hc)=\max_{U\subseteq\X,\,|U|\le m}\Delta(\Hc|_U)$ as in Section~\ref{sec:D}, and $\Delta_m(\Hc)=F_\infty(\Hc,m)=\min\{s,m\}$ by Lemma~\ref{lem:Finfty}; it is nondecreasing in $m$.
A star set of size $m$ \citep{HY15} consists of distinct points $x_1,\dots,x_m\in\X$ and $h_0,\dots,h_m\in\Hc$ with $h_i(x_i)\ne h_0(x_i)$ and $h_i(x_j)=h_0(x_j)$ for all $i\ne j$ in $[m]$. If $v$ has degree $m\ge1$ in $G(\Hc|_U)$, with edges in the directions $x_1,\dots,x_m$, then any $h_0,h_i\in\Hc$ with $h_0|_U=v$ and $h_i|_U=v^{(x_i)}$ ($v$ with coordinate $x_i$ flipped) form a star set of size $m$; so $\Hc$ has a star set of size $\Delta_m(\Hc)$ if $\Delta_m(\Hc)\ge1$. For a sample $S=((x_i,y_i))_{i\le n}$ let $V(S):=\{h\in\Hc:h(x_i)=y_i\ \forall i\}$, and let $\DIS(V):=\{x:h(x)\ne h'(x)\text{ for some }h,h'\in V\}$ for $V\subseteq\Hc$.

\paragraph{The loss and the two learners.} As $\ell_\lambda(\nu,y)$ is affine in $\nu$, the pointwise average of rules $g_1,\dots,g_N$ has risk $\frac1N\sum_tR_\lambda(g_t;P,h)$. As $K\ge1$, $\ell_\lambda(\nu,y)\ge1-\nu(y)$; and $0\le\ell_\lambda\le K$. We use the learner $g^\lambda$ of the upper bound in~\eqref{eq:transfer} and the disagreement-based learner $g^\infty$, defined before Lemma~\ref{lem:A-loo}; both depend on $S$ only through the set of its labeled pairs. On a sample $S=((x_i,y_i))_{i\le n}$ labeled by $h^\star\in\Hc$, the consistent set used there at a point $x$ is $V(S)|_U\ni h^\star|_U$ with $U=\{x_1,\dots,x_n,x\}$, so both learners predict $h^\star(x)$ at $x\notin\DIS(V(S))$, and $g^\infty$ abstains on $\DIS(V(S))$. Hence $g^\infty_S$ never errs and $R_\lambda(g^\infty_S;P,h^\star)=P(\DIS(V(S)))$ for every $\lambda\in[0,\infty]$, including the risk $R_\infty$ of Section~\ref{sec:loss}. Let $h^\star\in\Hc$, $m\ge2$, $z\in\X^m$, and let $z^{-j}$ be $z$ with entry $j$ removed. As the set of entries of $z$ has at most $m$ elements, Lemma~\ref{lem:A-loo} gives
\begin{equation}\label{eq:H-loo}
\sum_{j=1}^m\ell_\lambda\big(g^\lambda_{S_{h^\star}(z^{-j})}(z_j),h^\star(z_j)\big)\le F_\lambda(\Hc,m),\qquad\sum_{j=1}^m\one\big[g^\infty_{S_{h^\star}(z^{-j})}(z_j)(\abst)=1\big]\le \Delta_m(\Hc).
\end{equation}

\begin{proposition}\label{prop:H-prefix}
Let $h^\star\in\Hc$, $n\ge2$, $t_0=\lfloor n/2\rfloor$ and $N=n-t_0$. Let either $G=g^\lambda$, $B=K$ and $M_m=F_\lambda(\Hc,m)$, or $G=g^\infty$ at $\lambda=0$, $B=1$ and $M_m=\Delta_m(\Hc)$, and let $R_t(w):=R_\lambda(G_{S_{h^\star}(w_{1:t})};P,h^\star)$ for $w\in\X^n$. For every $P$ and $\delta\in(0,1)$, with probability at least $1-\delta$ over $w\sim P^n$, $N^{-1}\sum_{t=t_0}^{n-1}R_t(w)<(2/n)\big(e\ln3\,M_n+(2e/(e-1))B\ln(2/\delta)\big)$.
\end{proposition}

\begin{proof}
We follow the argument of \citet[Theorem~2.1]{ACSZ23}. In both cases $G_S$ depends on $S$ only through its set of labeled pairs, and on samples labeled by $h^\star$ its loss is at most $B$ (for $g^\infty$ at $\lambda=0$ it is the indicator of $\DIS$). By~\eqref{eq:H-loo}, $\sum_{j=1}^m\ell_\lambda(G_{S_{h^\star}(z^{-j})}(z_j),h^\star(z_j))\le M_m$ for all $m\ge2$ and $z\in\X^m$, and $M_m\ge0$ is nondecreasing in $m$ (Lemmas~\ref{lem:basic} and~\ref{lem:Finfty}). Let $\phi_t(w):=\ell_\lambda\big(G_{S_{h^\star}(w_{1:t})}(w_{t+1}),h^\star(w_{t+1})\big)\in[0,B]$ and $\Lambda(w):=\sum_{t=t_0}^{n-1}\phi_t(w)$.

\emph{Exponential inequalities.} For $W_1,\dots,W_N\in[0,1]$ adapted to a filtration $(\G_k)_{k=0}^N$ with $\bar W_k:=\E[W_k\mid\G_{k-1}]$, the bounds $e^{w}\le1+(e-1)w$ and $e^{-w}\le1-(1-e^{-1})w$ on $[0,1]$ and $1+u\le e^u$ make $\exp\sum_{i\le k}(W_i-(e-1)\bar W_i)$ and $\exp\sum_{i\le k}((1-e^{-1})\bar W_i-W_i)$ supermartingales starting at $1$. By Markov's inequality, each of the events (E1) $\sum_kW_k\ge(e-1)\sum_k\bar W_k+\ln(1/\delta')$ and (E2) $(1-e^{-1})\sum_k\bar W_k\ge\sum_kW_k+\ln(1/\delta')$ has probability at most $\delta'\in(0,1)$, as in \citet{ACSZ23} with both parameters equal to $1$.

\emph{Online losses.} Fix $z\in\X^n$, let $\sigma$ be a uniformly random permutation of $[n]$ and $z\circ\sigma:=(z_{\sigma(1)},\dots,z_{\sigma(n)})$. For $s=1,\dots,N$ let $W_s:=\phi_{n-s}(z\circ\sigma)/B$, and let $\G_s$ be generated by $\sigma(n-s+1),\dots,\sigma(n)$, with $\G_0$ trivial. $W_s$ is the normalized loss at $z_{\sigma(n-s+1)}$ of $G$ trained on the pairs indexed by $[n]\setminus\{\sigma(n-s+1),\dots,\sigma(n)\}$, so it is $\G_s$-measurable. Given $\G_{s-1}$, the set $I:=[n]\setminus\{\sigma(n-s+2),\dots,\sigma(n)\}$ of size $m:=n-s+1\ge2$ is fixed and $\sigma(n-s+1)$ is uniform on $I$. As $G$ trained on the pairs indexed by $I\setminus\{j\}$ is $G$ trained on $S_{h^\star}((z^I)^{-j'})$, where $z^I$ lists $(z_i)_{i\in I}$ in increasing order of $i$ and $j'$ is the position of $j$, $\E[W_s\mid\G_{s-1}]\le M_m/(Bm)\le M_n/(B(n-s+1))$. As $\sum_{s=1}^N1/(n-s+1)=\sum_{k=t_0+1}^n1/k\le\ln(n/t_0)$, (E1) with $\delta'=\delta/2$ gives $\Pr_\sigma(\Lambda(z\circ\sigma)\ge T)\le\delta/2$ for every $z$, where $T:=(e-1)\ln(n/t_0)M_n+B\ln(2/\delta)$. For $w\sim P^n$ independent of $\sigma$, $w\circ\sigma\sim P^n$, so $\Pr_w(\Lambda(w)\ge T)=\E_w[\Pr_\sigma(\Lambda(w\circ\sigma)\ge T)]\le\delta/2$.

\emph{Risks.} Let $\G'_k$ be generated by $w_1,\dots,w_{t_0+k}$ and $W'_k:=\phi_{t_0+k-1}(w)/B$. As $w_{t_0+k}\sim P$ is independent of $\G'_{k-1}$, $\E[W'_k\mid\G'_{k-1}]=R_{t_0+k-1}(w)/B$, and (E2) with $\delta'=\delta/2$ gives $\Pr\big((1-e^{-1})\sum_{t=t_0}^{n-1}R_t(w)\ge\Lambda(w)+B\ln(2/\delta)\big)\le\delta/2$.

\emph{Combination.} Outside an event of probability at most $\delta$, $(1-e^{-1})\sum_tR_t(w)<(e-1)\ln(n/t_0)M_n+2B\ln(2/\delta)$, that is, $\sum_tR_t(w)<e\ln(n/t_0)M_n+(2e/(e-1))B\ln(2/\delta)$. Finally $N\ge n/2$, and $n/t_0\le3$, as $n/t_0=2$ for even $n$ and $n/t_0=2n/(n-1)\le3$ for odd $n\ge3$.
\end{proof}

\begin{theorem}\label{thm:H-prefix}
Let $\Hc$ be nonempty, $n\ge2$, $t_0=\lfloor n/2\rfloor$, $N=n-t_0$ and $\bar g_S(x):=\frac1N\sum_{t=t_0}^{n-1}g^\lambda_{S_{\le t}}(x)$. For every $P$, $h^\star\in\Hc$ and $\delta\in(0,1)$, with probability at least $1-\delta$ over $x_{1:n}\sim P^n$, the sample $S=S_{h^\star}(x_{1:n})$ satisfies
\[
R_\lambda(\bar g_S;P,h^\star)\ \le\ \frac2n\Big(e\ln3\,F_\lambda(\Hc,n)+\frac{2e}{e-1}\,K\ln\frac2\delta\Big)\ \le\ \frac{6.33}{n}\Big(F_\lambda(\Hc,n)+K\ln\frac2\delta\Big).
\]
The learner $\bar g$ does not depend on $\delta$.
\end{theorem}

\begin{proof}
As $\ell_\lambda$ is affine in the action, $R_\lambda(\bar g_S;P,h^\star)=\frac1N\sum_{t=t_0}^{n-1}R_t(x_{1:n})$ with $G=g^\lambda$, so the average needs no majority vote, and Proposition~\ref{prop:H-prefix} gives the first inequality. The second follows from $2e\ln3<5.98$ and $4e/(e-1)<6.33$.
\end{proof}

As $f_K(a,b)\le K\min\{a,b\}$, $F_\lambda(\Hc,n)\le KF_0(\Hc,n)\le Kd_n$ by Theorem~\ref{thm:binary}(a), so the bound is at most $(6.33K/n)(d+\ln(2/\delta))$.

\begin{theorem}\label{thm:H-dis}
Let $\Hc$ be nonempty and $n\ge2$. For every $P$, $h^\star\in\Hc$ and $\delta\in(0,1)$, with probability at least $1-\delta$ over $x_{1:n}\sim P^n$, the sample $S=S_{h^\star}(x_{1:n})$ satisfies
\[
P\big(\DIS(V(S))\big)\ \le\ \frac2n\Big(e\ln3\,\Delta_n(\Hc)+\frac{2e}{e-1}\ln\frac2\delta\Big)\ \le\ \frac{6.33}{n}\Big(\Delta_n(\Hc)+\ln\frac2\delta\Big),
\]
where $\Delta_n(\Hc)=F_\infty(\Hc,n)=\min\{s,n\}$. Hence for every $\lambda\in[0,\infty]$, $g^\infty$ never errs and $R_\lambda(g^\infty_S;P,h^\star)$ satisfies the same bound.
\end{theorem}

\begin{proof}
Proposition~\ref{prop:H-prefix} with $G=g^\infty$ bounds $\frac1N\sum_{t=t_0}^{n-1}P(\DIS(V(S_{\le t})))$. As $V(S)\subseteq V(S_{\le t})$, $\DIS(V(S))\subseteq\DIS(V(S_{\le t}))$ for every $t$, so $P(\DIS(V(S)))$ is at most this average. The constants are those of Theorem~\ref{thm:H-prefix}, and $R_\lambda(g^\infty_S;P,h^\star)=P(\DIS(V(S)))$.
\end{proof}

The learner $g^\infty$ uses the whole sample; the averaging enters only the analysis.

\begin{theorem}\label{thm:H-lower}
Let $\lambda\in[0,\infty)$ and $n\ge1$.
\begin{itemize}
\item[(a)] If $\Hc$ is nontrivial, then $\eps_n(\Hc,\lambda,\delta)\ge\frac12\min\{1,K/2\}\min\{1,\ln(1/(4\delta))/n\}$ for every $\delta\in(0,1/4)$.
\item[(b)] Let $n\ge2$ and let $\Hc$ have a star set of size $m$ with $2\le m\le n+1$, for instance $m=\Delta_{n+1}(\Hc)$ if $\Delta_{n+1}(\Hc)\ge2$. Then every learner $g$ at sample size $n$ and every $\delta\in(0,1/32]$ satisfy $\sup_{P,h}Q_{1/32}(g;P,h)\ge5(m-1)/(64n)$ or $\sup_{P,h}Q_\delta(g;P,h)\ge(K/4)\min\{1,\ln(1/(8\delta))/n\}$. In particular $\eps_n(\Hc,\lambda,\delta)\ge\min\{5(m-1)/(64n),(K/4)\min\{1,\ln(1/(8\delta))/n\}\}$ for every $\delta\in(0,1/32]$.
\item[(c)] If $\Hc$ is nontrivial, then $\eps_n(\Hc,\lambda,\delta)\ge(1/e-2\delta)F_\lambda(\Hc,n+1)/(8(n+1))$ for every $\delta\in(0,1/(2e))$, and in particular $\eps_n(\Hc,\lambda,\delta)\ge F_\lambda(\Hc,n+1)/(33(n+1))$ for every $\delta\in(0,1/16]$.
\end{itemize}
\end{theorem}

\begin{proof}[Proof of (a)]
By nontriviality there are $h,h'\in\Hc$ and $x_0\ne x'$ with $h(x_0)=h'(x_0)=:b$ and $h(x')\ne h'(x')$. Let $u=\ln(1/(4\delta))>0$ and $p=1-e^{-u/n}$, so $(1-p)^n=4\delta$, and let $P$ put mass $p$ on $x'$ and $1-p$ on $x_0$. On the event $E$ that $x'$ is not sampled, of probability $4\delta$, the labeled sample is $((x_0,b))_{i\le n}$ under both targets. If $\nu$ is the action of $g$ at $x'$ on this sample, $\ell_\lambda(\nu,0)+\ell_\lambda(\nu,1)=2\nu(\abst)+K(\nu(0)+\nu(1))\ge2\min\{1,K/2\}$, so for one target $h^\dagger\in\{h,h'\}$ the risk on $E$ is at least $p\min\{1,K/2\}$. As $4\delta>\delta$, $Q_\delta(g;P,h^\dagger)\ge p\min\{1,K/2\}$, and $p\ge u/(n+u)\ge\frac12\min\{1,u/n\}$.
\end{proof}

\begin{proof}[Proof of (b)]
Fix $g$, $\delta\in(0,1/32]$ and a star set $x_1,\dots,x_m$, $h_0,\dots,h_m$; let $\eps_1:=\sup_{P,h}Q_{1/32}(g;P,h)$ and $\eps_2:=\sup_{P,h}Q_\delta(g;P,h)$. Let $u=\ln(1/(8\delta))\ge\ln4$ and $p=1-e^{-u/n}$, so $(1-p)^n=8\delta$ and $pK/2\ge(K/4)\min\{1,u/n\}$. We assume $\eps_2<pK/2$ and show $\eps_1\ge5(m-1)/(64n)$; the last sentence of (b) follows as $Q_\delta\ge Q_{1/32}$ for $\delta\le1/32$.

\emph{Scenarios.} In scenario A the target is $h_0$ and $P_A$ puts mass $1/n$ on each of $x_2,\dots,x_m$ and $1-(m-1)/n\ge0$ on $x_1$. For $2\le i\le m$, let $P^{-i}$ be $P_A$ conditioned on $\X\setminus\{x_i\}$ (of mass $1-1/n>0$); in scenario $B_i$ the target is $h_i$ and $P_i$ puts mass $p$ on $x_i$ and distributes $1-p$ according to $P^{-i}$. Let $E_i$ be the event that $x_i$ is not sampled. Then $P_A^n(E_i)=(1-1/n)^n\ge1/4$, as $n\ge2$, and $P_i^n(E_i)=8\delta$. Given $E_i$, in both scenarios the sample points are i.i.d.\ with law $P^{-i}$, supported where $h_0$ and $h_i$ agree, so the labeled sample has the same conditional law $\Pi_i$ given $E_i$ in A and $B_i$.

\emph{Scenario $B_i$.} Let $a_j(S):=g_S(x_j)(h_0(x_j))$. As $h_i(x_i)=1-h_0(x_i)$, $R_\lambda(g_S;P_i,h_i)\ge pK\,a_i(S)$. Let $\eta:=\eps_2/(pK)<1/2$ and $\Gamma_i:=\{S:a_i(S)>\eta\}$. On $E_i\cap\Gamma_i$ this risk exceeds $\eps_2\ge Q_\delta(g;P_i,h_i)$, so $8\delta\,\Pi_i(\Gamma_i)\le\delta$ and $\Pi_i(\Gamma_i)\le1/8$.

\emph{Scenario A.} As $\ell_\lambda(\nu,y)\ge1-\nu(y)$, $R_\lambda(g_S;P_A,h_0)\ge Y(S):=\frac1n\sum_{j=2}^m\one[E_j](1-a_j(S))$, with $0\le Y\le(m-1)/n$ and $1-a_j>1/2$ on $E_j\setminus\Gamma_j$. Hence
\[
\E_A[Y]\ \ge\ \frac1{2n}\sum_{j=2}^mP_A^n(E_j)\big(1-\Pi_j(\Gamma_j)\big)\ \ge\ \frac{m-1}{2n}\cdot\frac14\cdot\frac78=\frac{7(m-1)}{64\,n}.
\]
Also $\E_A[Y]\le\eps_1+((m-1)/n)\Pr_A(Y>\eps_1)$ and $\Pr_A(Y>\eps_1)\le\Pr_A(R_\lambda(g_S;P_A,h_0)>\eps_1)\le1/32$, as $\eps_1\ge Q_{1/32}(g;P_A,h_0)$. Hence $7(m-1)/(64n)\le\eps_1+2(m-1)/(64n)$.
\end{proof}

\noindent The proof of (c) compares the two targets at the ends of an edge whose direction is not sampled. For a finite nonempty $V\subseteq\{0,1\}^U$ let $\tilde\rho_K(V)$ be the maximum of $|E_V(A,W)|/(|A|+|W|/K)$ over disjoint $A,W\subseteq V$, not both empty.

\begin{lemma}\label{lem:H-density}
For every $\lambda\in[0,\infty)$ and every finite nonempty $V\subseteq\{0,1\}^U$, $\tilde\rho_K(V)\ge\tau_\lambda(V)/2$.
\end{lemma}

\begin{proof}
If $E(V)=\emptyset$, both sides are $0$. For $\lambda\ge1$, $|A|+|W|/K\le|A|+|W|/\lambda$ gives $\tilde\rho_K(V)\ge\rho_\lambda(V)\ge\tau_\lambda(V)/2$ by Lemma~\ref{lem:layer}. For $\lambda<1$, Lemma~\ref{lem:smalllambda} gives $\tau_\lambda(V)=K\tau_0(V)$ and a set $B$ with $\tau_0(V)=|E_V(B)|/|B|$. Splitting $B$ into its patterns $A$ with an even and $W$ with an odd number of ones, every edge of $E_V(B)$ joins $A$ and $W$, and $|A|+|W|/K\le|B|$, so $\tilde\rho_K(V)\ge\tau_0(V)=\tau_\lambda(V)/K>\tau_\lambda(V)/2$.
\end{proof}

\begin{lemma}\label{lem:H-charge}
Let $n\ge1$, $x_0\in\X$, $b\in\{0,1\}$, let $U'\subseteq\X\setminus\{x_0\}$ be finite and nonempty, $p\in(0,1/|U'|]$ and $q=(1-p)^n$. Let $V=\Hc^b_{x_0}|_{U'}$ be nonempty, let $A,W\subseteq V$ be disjoint with $E_V(A,W)\ne\emptyset$, and for $v\in V$ fix $h_v\in\Hc^b_{x_0}$ with $h_v|_{U'}=v$. Let $P$ put mass $p$ on each point of $U'$ and $1-p|U'|$ on $x_0$. Then every learner $g$ and every $\delta\in(0,1)$ satisfy
\[
\max_{v\in A\cup W}Q_\delta(g;P,h_v)\ \ge\ \frac{(q-2\delta)\,p\,|E_V(A,W)|}{2(|A|+|W|/K)}.
\]
\end{lemma}

\begin{proof}
Assume $q>2\delta$. Let $E=E_V(A,W)$, let $\deg(v)$ be the number of edges of $E$ at $v$, so $\sum_{v\in A\cup W}\deg(v)=2|E|$, let $\eps=\max_{v\in A\cup W}Q_\delta(g;P,h_v)$, draw $Z\sim P^n$ and set $R_v=R_\lambda(g_{S_{h_v}(Z)};P,h_v)$; then $\Pr(R_v>\eps)\le\delta$ for every $v$. Fix $Z$ and let $Y\subseteq U'$ be the set of points of $U'$ that do not occur in $Z$. For $e=\{a,w\}\in E$ with $a\in A$, $w\in W$ and direction $x_e\in Y$, the targets $h_a$ and $h_w$ agree at $x_0$ and on $U'\setminus\{x_e\}$, so they give the same labeled sample and the learner takes the same action $\nu_e$ at $x_e$. Charge $e$ to $a$ if $\nu_e(a(x_e))<1/2$, and to $w$ otherwise. As $\ell_\lambda(\nu,y)\ge1-\nu(y)$ and $\ell_\lambda(\nu,1-y)\ge K\nu(y)$, the charged end pays more than $1/2$ at $x_e$ if it is $a$, and at least $K/2$ if it is $w$. Let $C_v$ count the edges charged to $v$. The edges at $v$ have distinct directions, so $R_v\ge pC_v/2$ for $v\in A$ and $R_v\ge pKC_v/2$ for $v\in W$. With $C_v\le\deg(v)$ this gives $C_v\le2\eps/p+\deg(v)\one[R_v>\eps]$ for $v\in A$ and $C_v\le2\eps/(pK)+\deg(v)\one[R_v>\eps]$ for $v\in W$. Every edge with direction in $Y$ is charged once, and each direction is unsampled with probability $q$, so taking expectations,
\[
q|E|=\E\sum_{v\in A\cup W}C_v\ \le\ \frac{2\eps}p\Big(|A|+\frac{|W|}K\Big)+2\delta|E|.\qedhere
\]
\end{proof}

\begin{proof}[Proof of (c)]
Fix $g$ and take $p=1/(n+1)$, so $q=(1+1/n)^{-n}\ge1/e$. By nontriviality there are $h,h'\in\Hc$ and $x_0\ne x'$ with $h(x_0)=h'(x_0)=:b$ and $h(x')\ne h'(x')$; with $U'=\{x'\}$ and $A,W$ the two patterns, the ratio in Lemma~\ref{lem:H-charge} is $K/(K+1)\ge1/2$. Let $F=F_\lambda(\Hc,n+1)$. If $F<2$, this configuration has ratio at least $1/2>F/4$. If $F\ge2$, by Lemma~\ref{lem:basic} there is $U$ with $2\le|U|\le n+1$ and $\tau_\lambda(\Hc|_U)=F$; for $x_0\in U$ and $U'=U\setminus\{x_0\}$, Lemma~\ref{lem:A-split} gives $b$ with $\tau_\lambda(\Hc^b_{x_0}|_{U'})\ge F-c_K\ge F-1$, and Lemma~\ref{lem:H-density} gives $A,W$ with ratio at least $(F-1)/2\ge F/4$. In all cases one of the two configurations has ratio at least $F/4$, and Lemma~\ref{lem:H-charge} gives $\sup_{P,h}Q_\delta(g;P,h)\ge(1/e-2\delta)F/(8(n+1))$. For $\delta\le1/16$, $(1/e-1/8)/8>1/33$.
\end{proof}

\paragraph{Learners with a seed.} Let a learner draw a seed $\xi$ independent of the sample and output the rule $g_{S,\xi}$, where $\xi\mapsto g_{S,\xi}(x)(a)$ is measurable for every sample $S$, point $x$ and action $a$, and let $Q_\delta$ be defined with the probability taken jointly over $(S,\xi)$. Theorems~\ref{thm:H-prefix} and~\ref{thm:H-dis} concern learners without a seed, and Theorem~\ref{thm:H-lower} holds for learners with a seed by the same proofs. In (a), for every $\xi$ the two targets have risks on $E$ whose sum is at least $2p\min\{1,K/2\}$, so one target $h^\dagger$ has risk at least $p\min\{1,K/2\}$ on $E$ for a set of seeds of probability at least $\frac12$; as $E$ is independent of $\xi$, this event has probability at least $2\delta>\delta$. In (b), replace $S$ by $(S,\xi)$ and $a_j(S)$ by $a_j(S,\xi)=g_{S,\xi}(x_j)(h_0(x_j))$; since $\xi$ is independent of the sample, $(S,\xi)$ has the same conditional law given $E_i$ in scenarios A and $B_i$, and the argument is unchanged. In (c), the two targets of an edge with unsampled direction receive the same sample and the same seed, so the charging in Lemma~\ref{lem:H-charge} is carried out for each pair $(S,\xi)$ and the expectation is taken jointly. Consequently Proposition~\ref{prop:highprob}(b),(c) also hold when $\eps_n$ is the infimum over learners with a seed.

\begin{proof}[Proof of Proposition~\ref{prop:highprob}(b)]
Write $F:=F_\lambda(\Hc,n)$, $D:=\Delta_n(\Hc)=s_n$ and $L:=\ln(1/\delta)$; the hypothesis $s\ge2$ means $\Delta_{n+1}(\Hc)\ge2$.
\emph{Upper bound.} The learner that equals $g^\infty$ if $e\ln3\,D+c\ln(2/\delta)\le e\ln3\,F+cK\ln(2/\delta)$, with $c:=2e/(e-1)$, and $\bar g$ otherwise depends only on $(\Hc,\lambda,n,\delta)$, and by Theorems~\ref{thm:H-prefix} and~\ref{thm:H-dis} its $\delta$-quantile is at most $2/n$ times the smaller expression. As $\ln(2/\delta)\le2L$ and $e\ln3<4e/(e-1)$, $\eps_n\le(2/n)(4e/(e-1))\min\{F+KL,D+L\}$, and $\min\{F+KL,D+L\}\le F+\min\{KL,D\}+L$.

\emph{Lower bound.} As $\delta\le1/64$, $\ln(1/(4\delta))\ge L/2$ and $\ln(1/(8\delta))\ge L/2$, and as $L\le n$, the minima with $1$ in Theorem~\ref{thm:H-lower}(a),(b) are at least $L/(2n)$. Part (a) gives $\eps_n\ge L/(8n)$. Part (b) with $m=\Delta_{n+1}(\Hc)\ge2$ gives $\eps_n\ge\min\{5(m-1)/(64n),KL/(8n)\}\ge(5/128)\min\{D,KL\}/n$, as $m-1\ge m/2\ge D/2$. Part (c) gives $\eps_n\ge(1/e-1/32)F_\lambda(\Hc,n+1)/(8(n+1))\ge F/(36n)$, by Lemma~\ref{lem:basic}, $n+1\le3n/2$ and $(1/e-1/32)/12>1/36$. The largest of three numbers is at least their average, and $\min\{1/8,5/128,1/36\}=1/36$.
\end{proof}

\begin{proof}[Proof of Proposition~\ref{prop:highprob}(c)]
Let $K$ be an integer with $K>76C_0$, $\lambda:=K-1\ge1$, $m:=\lceil472\,C_0K\rceil+1$, let $x_1,\dots,x_m\in\X$ be distinct, $h_0\equiv0$, $h_i:=\one_{\{x_i\}}$ and $\Hc:=\{h_0,\dots,h_m\}$. Then $x_1,\dots,x_m$ and $h_0,\dots,h_m$ form a star set of size $m$, $\Hc$ is nontrivial, and its star number is $m$, since every hypothesis vanishes outside $\{x_1,\dots,x_m\}$. Let $\delta:=e^{-m}/8\le1/64$ and $n_0:=m+3$, and let $n\ge n_0$; then $2\le m\le n+1$ and $\ln(1/\delta)=m+\ln8\le n$.

\emph{$F_\lambda(\Hc,n)\le K$.} For finite $U$, $G(\Hc|_U)$ is a star with $k:=|U\cap\{x_1,\dots,x_m\}|$ leaves, as two leaves differ in two coordinates, and has no edge if $k=0$. For $k\ge1$, the action with mass $\beta:=K/(k+\lambda)\in(0,1]$ on $\abst$ and $1-\beta$ on $0$ on every edge costs $k\beta$ at the center and $K-\lambda\beta$ at each leaf, both $kK/(k+\lambda)\le K$, so $\tau_\lambda(\Hc|_U)\le K$ by Lemma~\ref{lem:lp}.

\emph{Upper and lower bounds.} By Theorem~\ref{thm:H-prefix} and $F_\lambda(\Hc,n)\le K$, $\eps_n(\Hc,\lambda,1/64)\le(2/n)\big(e\ln3+(2e/(e-1))\ln128\big)K<36.8\,K/n$. By Theorem~\ref{thm:H-dis} with $\Delta_n(\Hc)=\min\{m,n\}=m$, $\eps_n(\Hc,\lambda,\delta)\le(6.33/n)(m+\ln(2/\delta))=(6.33/n)(2m+\ln16)<19m/n$, as $\ln16<3\le m$. Conversely, Theorem~\ref{thm:H-lower}(a) with $K\ge2$, $\ln16\le n$ and $\ln(1/(4\delta))=m+\ln2\le n$ gives $\eps_n(\Hc,\lambda,1/64)\ge\ln16/(2n)$ and $\eps_n(\Hc,\lambda,\delta)\ge(m+\ln2)/(2n)$. As $K$ and $m$ do not depend on $n$, both quantiles are of order $1/n$.

\emph{The two confidence levels.} Let $g$ be a learner at sample size $n$ with $\sup_{P,h}Q_{1/64}(g;P,h)\le C_0\eps_n(\Hc,\lambda,1/64)$. As $Q_{1/32}\le Q_{1/64}$, $\sup_{P,h}Q_{1/32}(g;P,h)<36.8\,C_0K/n\le5(m-1)/(64n)$, as $m-1\ge472\,C_0K>471.04\,C_0K=(64\cdot36.8/5)C_0K$. Theorem~\ref{thm:H-lower}(b), with $\ln(1/(8\delta))=m\le n$, gives $\sup_{P,h}Q_\delta(g;P,h)\ge Km/(4n)>19C_0m/n>C_0\,\eps_n(\Hc,\lambda,\delta)$, as $K>76C_0$.
\end{proof}

\paragraph{The factor $K$ in the confidence term.} For $\lambda\le1$ ($K\le2$), Theorem~\ref{thm:H-prefix} gives the learner $\bar g$, which does not depend on $\delta$, risk at most $12.66(F_\lambda(\Hc,n)+\ln(2/\delta))/n$ with probability $1-\delta$, and at $\lambda=\infty$ Theorem~\ref{thm:H-dis} gives $g^\infty$ risk at most $6.33(F_\infty(\Hc,n)+\ln(2/\delta))/n$; both have the form of the PAC bound of \citet{ACSZ23}. Under the conditions of Proposition~\ref{prop:highprob}(b), this bound for $\bar g$ is within an absolute constant of $\eps_n$ at every $\delta$. For $n\ge2$, $K:=\lceil\sqrt n\rceil$, $\lambda:=K-1$, $\delta:=e^{-K}/8\le1/32$ and the star class $\{h_0,\dots,h_m\}$ of the proof of~(c) with $m=n+1$, Theorem~\ref{thm:H-lower}(b) gives $\eps_n(\Hc,\lambda,\delta)\ge\min\{5/64,K^2/(4n)\}=5/64$, while $(F_\lambda(\Hc,n)+\ln(2/\delta))/n\le(2K+\ln16)/n\to0$. Thus the factor $K$ on $\ln(2/\delta)$ in Theorem~\ref{thm:H-prefix} cannot be replaced by an absolute constant uniformly in $\lambda$.

\end{document}